\documentclass[10pt,twocolumn,letterpaper]{article}

\usepackage[pagenumbers]{cvpr} 

\usepackage[dvipsnames]{xcolor}
\usepackage{mathtools}
\usepackage{tikz}
\usepackage{pgfplots}

\newcommand{\RR}{\mathbb{R}}

\definecolor{cvprblue}{rgb}{0.21,0.49,0.74}
\usepackage[pagebackref,breaklinks,colorlinks,citecolor=cvprblue]{hyperref}

\def\paperID{387} 
\def\confName{3DV\xspace}
\def\confYear{2026\xspace}

\title{Minimal Solvers for Full DoF Motion Estimation from Asynchronous Tracks}

\author{Petr Hruby\\
ETH Z\"urich\\
R\"amistrasse 101, 8006 Z\"urich\\
{\tt\small petr.hruby@inf.ethz.ch}
\and
Marc Pollefeys\\
ETH Z\"urich / Microsoft Spatial AI Lab\\
R\"amistrasse 101, 8006 Z\"urich\\
{\tt\small marc.pollefeys@inf.ethz.ch}
}

\begin{document}
\maketitle
\begin{abstract}
We address the problem of estimating both translational and angular velocity of a camera from asynchronous point tracks, a formulation relevant to rolling shutter and event cameras. Since the original problem is non-polynomial, we propose a polynomial approximation, classify the resulting minimal problems, and determine their algebraic degrees. Furthermore, we develop minimal solvers for several problems with low degrees and evaluate them on synthetic and real datasets. The code will be made publicly available.
\end{abstract}    
\section{Introduction}
\label{sec:intro}

Relative pose estimation is a fundamental task in computer vision, with applications in 3D reconstruction~\cite{schoenberger2016sfm, pan2024glomap}, Simultaneous Localization and Mapping (SLAM)~\cite{DBLP:journals/corr/DeToneMR17}, camera self-calibration~\cite{DBLP:conf/cvpr/CinDMP24}, multi-view stereo~\cite{DBLP:conf/cvpr/FurukawaCSS10}, and visual odometry~\cite{DBLP:conf/cvpr/NisterNB04}. Most minimal solvers for relative pose estimation assume a pinhole~\cite{DBLP:journals/pami/Nister04} or radially distorted~\cite{DBLP:conf/cvpr/HrubyKDOPPL23} camera model, where all image rows are captured simultaneously. However, this assumption does not hold for two important sensor types: \emph{rolling shutter cameras} and \emph{event cameras}. Rolling shutter cameras capture images row by row, while event cameras output a continuous stream of intensity changes. When these cameras move, the resulting image differs from that of the pinhole model.

Among these two sensor modalities, rolling shutter cameras have received more attention. Numerous methods have been developed for absolute pose estimation~\cite{DBLP:conf/eccv/Ait-AiderALM06,DBLP:conf/iros/SaurerPL15,DBLP:conf/eccv/MagerandBAP12,DBLP:journals/pami/AlblKLP20,DBLP:conf/cvpr/AlblKP16,DBLP:conf/cvpr/BaiSB22,DBLP:conf/eccv/LaoAB18,DBLP:conf/eccv/KukelovaASSP20,DBLP:conf/accv/KukelovaASP18,DBLP:conf/cvpr/Ait-AiderBA07,DBLP:conf/cvpr/AlblKP15,DBLP:journals/corr/abs-cs-0503076}, image rectification~\cite{DBLP:conf/cvpr/RengarajanRA16,DBLP:conf/iccv/PurkaitZL17,DBLP:conf/wacv/PurkaitZ18,DBLP:journals/prl/LaoAA18,DBLP:conf/iccv/YanTZL23,DBLP:conf/cvpr/ZhuangTJCC19,DBLP:conf/cvpr/AlblKLPPS20,DBLP:journals/ivc/FanDZW22,DBLP:conf/iccv/ShangRFWLZ23,DBLP:conf/iccv/Ait-AiderB09,DBLP:conf/eccv/ZhuangT20,DBLP:conf/mmsp/JiaE12,DBLP:journals/ijcv/RingabyF12,DBLP:conf/cvpr/ForssenR10,DBLP:conf/iccv/QuLWWZ023,DBLP:journals/pami/QuLZAL23,DBLP:conf/cvpr/VasuRR18,DBLP:journals/cviu/FanDW21,DBLP:journals/spl/SunLS16,DBLP:journals/pami/FanDL23,DBLP:journals/corr/abs-2204-13886}, structure-from-motion~\cite{DBLP:conf/iccvw/HedborgRFF11,DBLP:conf/iccv/SaurerKBP13,DBLP:conf/ismar/KleinM09}, and bundle adjustment~\cite{DBLP:conf/cvpr/HedborgFFR12,DBLP:conf/bmvc/NguyenL16,DBLP:conf/cvpr/LiaoQXZL23,DBLP:conf/cvpr/SaurerPL16,DBLP:conf/iclr/LiWZLL24,DBLP:conf/eccv/ZhangLXLLL24,DBLP:conf/eccv/XuLGLL24,DBLP:conf/eccv/SchubertDUSC18}. 

Relative pose estimation for rolling shutter cameras, however, remains a particularly challenging problem. Existing methods typically treat the cameras as moving independently, requiring the estimation of 17 parameters in the bifocal case (5 for the relative pose and 6 for the motion of each camera). The full-DoF problem is still unsolved, and many approaches rely on simplifying assumptions: zero angular velocity~\cite{DBLP:conf/cvpr/DaiLK16, DBLP:conf/iccv/ZhuangCL17}, known angular velocity~\cite{DBLP:journals/corr/abs-1712-00184,DBLP:journals/corr/abs-1904-06770}, or known vertical direction~\cite{ours_no_model}. Other methods refine pinhole-based pose estimates with rolling shutter models~\cite{DBLP:journals/corr/LeeY17a}, or target special cases such as planar scenes~\cite{9020067} or camera rigs~\cite{DBLP:conf/icip/WangFD20}. Recent works have surveyed minimal solvers for rolling shutter relative pose estimation using points and lines under special order-one models~\cite{DBLP:journals/corr/abs-2403-11295}, or purely from lines without a camera model~\cite{ours_no_model}. For video-based rolling shutter imagery, \cite{linear_pure_trans} proposed a method assuming constant translational velocity over neighboring frames, with known angular velocity.

Event camera pose estimation has been studied less extensively. Some works addressed motion estimation from collinear events, both under known~\cite{DBLP:conf/iccv/GaoSGC0K23,DBLP:conf/cvpr/GaoGS0K24} and unknown~\cite{DBLP:conf/cvpr/0001GLK25} angular velocity. A recent method~\cite{linear_pure_trans} proposed estimating translational motion from asynchronous point tracks assuming known angular velocity.

In this paper, we extend the problem of~\cite{linear_pure_trans} to the full-DoF setting, where both translational and angular velocities are unknown. We approximate the rotation with a polynomial model of degree $K$, classify minimal problems for $K<6$, and construct minimal solvers for $K=1$ and $K=2$. Furthermore, we provide a geometric explanation for why, in the pure translational case, a single essential matrix fits all data, which allows the classical five-point solver~\cite{DBLP:journals/pami/Nister04} to be applied. Throughout the paper, we assume a fully calibrated camera without radial distortion.

\section{Problem Formulation}
\label{sec:formulation}

In this section, we formulate the problem of estimating the camera velocity and rotation velocity from asynchronous tracks, introduce polynomial approximations of this problem, and enumerate minimal cases of this problem.

We consider a calibrated perspective camera whose center moves along a straight line with a constant translational velocity $\mathbf{V} \in \mathbb{R}^3$, while simultaneously rotating around a fixed axis $\mathbf{a} \in \mathbb{R}^3$ (with $\lVert \mathbf{a} \rVert = 1$) at a constant angular velocity $\omega \in \mathbb{R}$.

Due to gauge freedom, we may, without loss of generality, assume that at the initial time $t=0$ the camera center is located at the origin,  
\[
    \mathbf{C}(0) = \mathbf{0},
\]
and its orientation is given by the identity rotation matrix,  
\[
    \mathbf{R}(0) = \mathbf{I}.
\]
For a general time $t \in \mathbb{R}$, the camera center and orientation are given by
\begin{equation}
    \mathbf{C}(t) = t \cdot \mathbf{V},
\end{equation}
\begin{equation}
    \mathbf{R}(t) = \mathbf{I} + \sin(t \cdot \omega)\,[\mathbf{a}]_{\times} + \big(1 - \cos(t \cdot \omega)\big) [\mathbf{a}]_{\times}^2, \label{eq:orientation_exact}
\end{equation}
where $[\mathbf{a}]_{\times}$ denotes the skew-symmetric matrix associated with the vector $\mathbf{a}$.

Let $\{\mathbf{X}_i \in \mathbb{R}^3 \mid i = 1,\dots,n\}$ be a set of $n \in \mathbb{Z}_{+}$ 3D points.  
Each point $\mathbf{X}_i$ is observed at $m \in \mathbb{Z}_{+}$ distinct time instants $\{t_{i,j} \in \mathbb{R} \mid j = 1,\dots,m\}$. The perspective projection $\mathbf{p}_{i,j} \in \mathbb{P}^2$ of point $\mathbf{X}_i$ at time $t_{i,j}$ is given by
\begin{equation}
    \mathbf{p}_{i,j} \sim \mathbf{R}(t_{i,j}) \cdot \big( \mathbf{X}_i - \mathbf{C}(t_{i,j}) \big),
    \label{eq:projection_exact}
\end{equation}
where $\sim$ denotes equality up to a nonzero scale.

In the considered problem formulation, the projections $\mathbf{p}_{i,j}$ and the corresponding time instants $t_{i,j}$ are assumed to be known. The task is to jointly estimate the motion parameters $\mathbf{V}$, $\mathbf{a}$, and $\omega$, as well as the 3D coordinates of the points $\mathbf{X}_i$, $i = 1,\dots,n$, such that the projection model~\eqref{eq:projection_exact} holds for all observations.


\subsection{Capture Times for Rolling Shutter Cameras}\label{sec:capture_times}

Event cameras and rolling shutter cameras differ in the way they provide the capture times $t_{i,j}$ corresponding to the projections $\mathbf{p}_{i,j}$.  
For event cameras, these times are measured explicitly along with the projections, so both $\mathbf{p}_{i,j}$ and $t_{i,j}$ are directly available.  
In contrast, for rolling shutter cameras, the capture times must be computed from the projection coordinates 
\[
    \mathbf{p}_{i,j} = [u_{i,j} \ v_{i,j} \ 1]^\top
\]
and the known camera parameters.

Let $f \in \mathbb{R}$ denote the focal length of the camera in pixels, $h \in \mathbb{R}$ the image height in pixels, $t_r$ the frame readout time, and $t_d$ the inter-frame delay.
Since a rolling shutter camera acquires the image row by row, the time required to read a single row is $\frac{t_r}{h}$.  
The row index of the projection $\mathbf{p}_{i,j}$ can be determined from the $y$-coordinate of its uncalibrated projection $\mathbf{K} \mathbf{p}_{i,j}$, which is given by
\[
    f \cdot v_{i,j} + \frac{h}{2}.
\]


We choose $t=0$ to correspond to the capture time of the middle row ($\frac{h}{2}$-th) of the first frame.  
Under this convention, the capture time $t_{i,1}$ of projection $\mathbf{p}_{i,1}$ in the first frame is
\begin{equation}
    t_{i,1} = \frac{f \cdot v_{i,1} \cdot t_r}{h}.
\end{equation}
Furthermore, the middle row of the second frame is captured at $t = t_r + t_d$, and, in general, the middle row of the $j$-th frame is captured at
\[
    t = (j-1) \cdot (t_r + t_d).
\]
Thus, the capture time $t_{i,j}$ of projection $\mathbf{p}_{i,j}$ is
\begin{equation}
    t_{i,j} = (j-1) \cdot (t_r + t_d) + \frac{f \cdot v_{i,j} \cdot t_r}{h}. \label{eq:capture_time_rs}
\end{equation}

The parameters $t_r$ and $t_d$ can be obtained from the camera specifications or through calibration, as in~\cite{DBLP:conf/cvpr/ForssenR10}.  
However, this information may not always be available.  
Since the delay $t_d$ is often negligible, an alternative approach is to assume $t_d = 0$ and set
\[
    t_{i,j} = (j-1) + \frac{f \cdot v_{i,j}}{h},
\]
while redefining the motion parameters as
\[
    \mathbf{V} \coloneq t_r \cdot \mathbf{V}, \quad \omega \coloneq t_r \cdot \omega.
\]
In this case, the recovered velocities are expressed in scaled units rather than physical ones; nonetheless, the camera poses at the middle row of the $j$-th frame, $\mathbf{C}(j-1)$ and $\mathbf{R}(j-1)$, can still be determined without knowledge of the readout time.

\subsection{Approximating Constant Rotation} \label{sec:appx}

Here, we introduce polynomial approximations of the camera motion model~\eqref{eq:projection_exact} and discuss their implications for minimal problem formulations.

Equation~\eqref{eq:projection_exact} imposes constraints on the motion parameters $\mathbf{V}, \mathbf{a}, \omega$ and on the 3D structure of the scene.  
However, the exact rotation model $\mathbf{R}(t)$ given in~\eqref{eq:orientation_exact}, which assumes a constant angular velocity $\omega \in \mathbb{R}$, cannot be expressed exactly as a polynomial function.  
The reason is that, for any fixed $\omega \neq 0$, there exist infinitely many isolated values of $t$ such that $\mathbf{R}(t) = \mathbf{I}$; conversely, for any fixed $t \neq 0$, there exist infinitely many isolated values of $\omega$ for which $\mathbf{R}(t) = \mathbf{I}$.  
This contradicts a fundamental property of polynomial systems, which either have a finite number of solutions or their solution sets have dimension at least one.

Therefore, to solve the problem using standard algebraic methods such as Gr\"obner basis techniques~\cite{DBLP:conf/cvpr/LarssonAO17} or homotopy continuation~\cite{DBLP:conf/cvpr/FabbriDFRPTWHGK20}, the Rodrigues formula~\eqref{eq:orientation_exact} must be approximated by a polynomial expression.  
For this purpose, we use its Taylor expansion~\cite{pro}:
\begin{equation}
    \mathbf{R}(t) = \sum_{k=0}^{\infty} \frac{t^k \cdot \omega^k}{k!} [\mathbf{a}]_{\times}^k 
    = \sum_{k=0}^{\infty} \frac{t^k}{k!} [\omega \cdot \mathbf{a}]_{\times}^k,
    \label{eq:taylor_full}
\end{equation}
where $[\mathbf{a}]_{\times}$ denotes the skew-symmetric matrix of $\mathbf{a}$.

Truncating~\eqref{eq:taylor_full} to terms of degree at most $K$ yields a polynomial approximation $\mathbf{R}_K(t)$ of the rotation matrix.  
Defining the Euler vector $\mathbf{v} = \omega \cdot \mathbf{a} \in \RR^3$, we can write
\begin{equation}
    \mathbf{R}_K(t) = \sum_{k=0}^{K} \frac{t}{k!} [\mathbf{v}]_{\times}^k.
    \label{eq:taylor_finite}
\end{equation}
In the special case $K=1$, we obtain the linearized model $\mathbf{R}(t) \approx \mathbf{I} + t \,[\mathbf{v}]_{\times}$, which has been employed in various applications~\cite{DBLP:conf/cvpr/AlblKP15, DBLP:conf/cvpr/AlblKP16, DBLP:journals/pami/AlblKLP20}.

Since $\mathbf{R}_K(t)$ is generally not an exact rotation matrix, the identity $\mathbf{R}_K(t)^{-T} = \mathbf{R}_K(t)$ does not hold in general.  
Therefore, there are two natural ways to approximate~\eqref{eq:projection_exact} using $\mathbf{R}_K(t)$:  

\begin{itemize}
    \item \textbf{Approximation~1}:
    \begin{equation}
        \mathbf{p}_{i,j} \sim \mathbf{R}_K(t_{i,j}) \big( \mathbf{X}_i - \mathbf{C}(t_{i,j}) \big),
        \label{eq:appx1_proj}
    \end{equation}
    which leads to the constraint
    \begin{equation}
        [\mathbf{p}_{i,j}]_{\times} \,\mathbf{R}_K(t_{i,j}) \big( \mathbf{X}_i - \mathbf{C}(t_{i,j}) \big) = 0.
        \label{eq:appx1_cons}
    \end{equation}
    
    \item \textbf{Approximation~2}:
    \begin{equation}
        \mathbf{R}_K(t_{i,j})^T \,\mathbf{p}_{i,j} \sim \big( \mathbf{X}_i - \mathbf{C}(t_{i,j}) \big),
        \label{eq:appx2_proj}
    \end{equation}
    which leads to the constraint
    \begin{equation}
        \big[ \mathbf{R}_K(t_{i,j})^T \,\mathbf{p}_{i,j} \big]_{\times} \big( \mathbf{X}_i - \mathbf{C}(t_{i,j}) \big) = 0.
        \label{eq:appx2_cons}
    \end{equation}
\end{itemize}

In both cases, the constraints~\eqref{eq:appx1_cons} and~\eqref{eq:appx2_cons} are homogeneous in the elements of $\mathbf{V}$ and $\mathbf{X}_i$, implying that velocity and 3D structure can only be recovered up to a common scale.  
Furthermore, as these constraints are also homogeneous in the elements of $\mathbf{R}_K(t_{i,j})$, we use the adjoint of $\mathbf{R}_K(t_{i,j})$ instead of the inverse to avoid dividing by $\det \mathbf{R}_K(t_{i,j})$, and multiply each term in~\eqref{eq:taylor_finite} by $K!$ to avoid division by $k!$.  
As shown in Sec.~\ref{sec:minimal_problems}, the two approximations lead to minimal problems with different numbers of solutions.

\subsection{Multifocal constraints} \label{sec:multifocal}

Here, we are going to discuss the bi- and trifocal constraints obtained by eliminating 3D points $\mathbf{X}_i$ from constraints \eqref{eq:appx1_proj}, \eqref{eq:appx2_proj}.

For each observation $(i,j)$, we define the projection matrix
\begin{equation}
    \mathbf{P}_{i,j} =
    \begin{bmatrix}
        \mathbf{I} \ \big\vert\  -t_{i,j} \mathbf{V}
    \end{bmatrix}.
\end{equation}
Furthermore, we set
\[
    \mathbf{A}_{i,j} =
    \begin{cases}
        \mathbf{R}_K(t_{i,j})^{-1}, & \text{for Approximation~1}, \\
        \mathbf{R}_K(t_{i,j})^{T}, & \text{for Approximation~2}.
    \end{cases}
\]
With these definitions, the projection equation can be written as
\begin{equation}
    \mathbf{A}_{i,j} \,\mathbf{p}_{i,j} \ \sim\  \mathbf{P}_{i,j}
    \begin{bmatrix}
        \mathbf{X}_i \\ 1
    \end{bmatrix}.
    \label{eq:projection_with_A}
\end{equation}

Fixing a point index $i \in \{1,\dots,n\}$ and two distinct observation indices $j, j' \in \{1,\dots,m\}$, the essential matrix between the corresponding projection matrices $\mathbf{P}_{i,j}$ and $\mathbf{P}_{i,j'}$ is simply $[\mathbf{V}]_{\times}$.  
Therefore, the transformed image vectors $\mathbf{A}_{i,j} \mathbf{p}_{i,j}$ and $\mathbf{A}_{i,j'} \mathbf{p}_{i,j'}$ satisfy the epipolar constraint
\begin{equation}
    \mathbf{p}_{i,j'}^{\top} \,\mathbf{A}_{i,j'}^{\top} \,[\mathbf{V}]_{\times} \,\mathbf{A}_{i,j} \,\mathbf{p}_{i,j} = 0. \label{eq:bifocal_constraint}
\end{equation}
For Approximation~1, this becomes
\begin{equation}
    \mathbf{p}_{i,j'}^{\top} \,\mathbf{R}_K(t_{i,j'})^{-T} \,[\mathbf{V}]_{\times} \,\mathbf{R}_K(t_{i,j})^{-1} \,\mathbf{p}_{i,j} = 0,
\end{equation}
and for Approximation~2,
\begin{equation}
    \mathbf{p}_{i,j'}^{\top} \,\mathbf{R}_K(t_{i,j'}) \,[\mathbf{V}]_{\times} \,\mathbf{R}_K(t_{i,j})^{T} \,\mathbf{p}_{i,j} = 0. \label{eq:bifocal_constraint_A2}
\end{equation}

In the special case of pure translation, the constraint reduces to
\begin{equation}
    \mathbf{p}_{i,j'}^{\top} \,[\mathbf{V}]_{\times} \,\mathbf{p}_{i,j} = 0,
\end{equation}
which reveals that a single essential matrix $[\mathbf{V}]_{\times}$ exists for all correspondences, regardless of their capture times.

We can also derive a trifocal constraint.  
Fix $i \in \{1,\dots,n\}$ and three distinct observation indices $j, j', j'' \in \{1,\dots,m\}$.  
The trifocal tensor relating $\mathbf{P}_{i,j}$, $\mathbf{P}_{i,j'}$, and $\mathbf{P}_{i,j''}$ takes the form~\cite{HZ}:
\begin{equation}
\begin{split}
    \mathbf{T}_1^{j,j',j''} &= (t_{i,j} - t_{i,j''})\,\mathbf{e}_1 \mathbf{V}^{\top} - (t_{i,j} - t_{i,j'})\,\mathbf{V} \mathbf{e}_1^{\top}, \\
    \mathbf{T}_2^{j,j',j''} &= (t_{i,j} - t_{i,j''})\,\mathbf{e}_2 \mathbf{V}^{\top} - (t_{i,j} - t_{i,j'})\,\mathbf{V} \mathbf{e}_2^{\top}, \\
    \mathbf{T}_3^{j,j',j''} &= (t_{i,j} - t_{i,j''})\,\mathbf{e}_3 \mathbf{V}^{\top} - (t_{i,j} - t_{i,j'})\,\mathbf{V} \mathbf{e}_3^{\top},
\end{split}
\end{equation}
where $\mathbf{e}_1, \mathbf{e}_2, \mathbf{e}_3 \in \mathbb{R}^3$ are the standard basis vectors.

From~\eqref{eq:projection_with_A}, the transformed points $\mathbf{A}_{i,j} \mathbf{p}_{i,j}$, $\mathbf{A}_{i,j'} \mathbf{p}_{i,j'}$, and $\mathbf{A}_{i,j''} \mathbf{p}_{i,j''}$ satisfy the trifocal constraint
\begin{equation}
\begin{split}
    [\mathbf{A}_{i,j'} \mathbf{p}_{i,j'}]_{\times}
    \big( (\mathbf{A}_{i,j} \mathbf{p}_{i,j})_{(1)} \,\mathbf{T}_1^{j,j',j''}
       + (\mathbf{A}_{i,j} \mathbf{p}_{i,j})_{(2)} \,\mathbf{T}_2^{j,j',j''} \\
       + (\mathbf{A}_{i,j} \mathbf{p}_{i,j})_{(3)} \,\mathbf{T}_3^{j,j',j''} \big)
    [\mathbf{A}_{i,j''} \mathbf{p}_{i,j''}]_{\times} = \mathbf{0}.
    \label{eq:trifocal_constraint}
\end{split}
\end{equation}
For Approximation~1, we substitute
$\mathbf{A}_{i,j} = \mathbf{R}_K(t_{i,j})^{-1}$,
$\mathbf{A}_{i,j'} = \mathbf{R}_K(t_{i,j'})^{-1}$,
$\mathbf{A}_{i,j''} = \mathbf{R}_K(t_{i,j''})^{-1}$,
and for Approximation~2,
$\mathbf{A}_{i,j} = \mathbf{R}_K(t_{i,j})^{T}$, 
$\mathbf{A}_{i,j'} = \mathbf{R}_K(t_{i,j'})^{T}$, 
$\mathbf{A}_{i,j''} = \mathbf{R}_K(t_{i,j''})^{T}$.

\subsection{Minimal Problems} \label{sec:minimal_problems}

We now enumerate the minimal problems for the proposed formulations and report their numbers of solutions.

A \textit{minimal problem} is a parametric polynomial system $f(\theta,\mathbf{x})=0$ that has a finite, nonzero number of solutions $\mathbf{x}^*$ for generic parameters $\theta$ \cite{DBLP:journals/pami/DuffKLP24}.
Compared to overconstrained systems, minimal problems offer two key advantages: (i) they typically involve the smallest sufficient number of constraints, reducing RANSAC runtime; and (ii) they can be solved via standard algebraic techniques, such as Gr\"obner bases\cite{DBLP:conf/cvpr/LarssonAO17} or homotopy continuation~\cite{DBLP:conf/cvpr/FabbriDFRPTWHGK20}.

A necessary condition for minimality is that the problem is \textit{balanced}, \textit{i.e.}, the number of independent constraints equals the number of degrees of freedom (DoF) of the space of unknowns.

For $K=0$, the model reduces to pure translation, which is a special case of the known angular velocity problem treated in~\cite{linear_pure_trans}.
We now analyze the balanced condition for problems~\eqref{eq:appx1_cons} and~\eqref{eq:appx2_cons}, assuming $K \geq 1$.

The Euler vector $\mathbf{v} \in \mathbb{R}^3$ has $3$ DoF. The translational velocity $\mathbf{V} \in \mathbb{R}^3$, being recoverable only up to scale, has $2$ DoF. Each 3D point contributes $3$ DoF. Hence, the unknown space has $3 + 2 + 3\cdot n$ DoF. Each constraint of the form \eqref{eq:appx1_cons} or \eqref{eq:appx2_cons} fixes $2$ DoF. Therefore, the condition for balanced problems is
\begin{equation}
        2 \cdot m \cdot n = 5 + 3 \cdot n. \label{eq:balanced_condition}
\end{equation}
Balanced problems correspond to all $m,n \in \mathbb{N}$ satisfying \eqref{eq:balanced_condition}, regardless of the approximation type or degree $K$.
For fixed $m$, the number of points $n$ for a balanced problem is
\begin{equation}
    n(m) = \frac{5}{2 \cdot m - 3}.
\end{equation}
The only pairs $(m,n)$ that are balanced are:
\begin{itemize}
\item $m=2$, $n=5$,
\item $m=4$, $n=1$.
\end{itemize}

For $m=3$, $n(3) = 1.66$, so there is no balanced problem using all constraints. However, we can set
\begin{itemize}
\item $m=3$, $n=2$,
\end{itemize}
and obtain a balanced problem by omitting one equation.

For $m=1$, the system is underconstrained for all $n$. For $m \geq 5$, the problems are overconstrained for any $n$, since $n(4)=1$ and $n(m)$ increases strictly for $m \geq 2$.

To verify the minimality of the problems and to determine their numbers of solutions, we encoded the case $m=2, n=5$ using the bifocal constraints~\eqref{eq:bifocal_constraint}, and the cases $m=3, n=2$ and $m=4, n=1$ using the original constraints~\eqref{eq:appx1_cons} or~\eqref{eq:appx2_cons}.
Using Gr"obner basis computations, we verified that all balanced problems with $N \leq 6$ are minimal, and we computed their degrees. The degrees are presented in Table~\ref{tab:degrees_appx1} for both Approximation1 and Approximation2.

The results in the table indicate that, in certain cases, the chosen type of approximation has a substantial impact on the number of solutions. For example, the problem with $m=2$, $n=5$, and $K=1$ has $120$ solutions under Approximation~1, but only $20$ solutions under Approximation~2. Conversely, the problem with $m=4$, $n=1$, and $K=1$ has only $2$ solutions for Approximation~1, compared to $8$ solutions for Approximation~2. In general, problems with $m=2$, $n=5$ and $m=3$, $n=2$ tend to have fewer solutions under Approximation~2, whereas problems with $m=4$, $n=1$ have identical numbers of solutions for both approximations, except in the case $K=1$.

These findings demonstrate that the choice of approximation can significantly influence the runtime of the solver, since the computational cost often scales with the number of solutions. Moreover, the table shows that for $2 \leq K \leq 6$, the number of solutions for each problem grows linearly with $K$. While it is impossible to verify this numerically for all $K \geq 2$ up to infinity, polynomial dependencies of the number of solutions on problem parameters have been proven for other geometric problems—such as optimal multi-view triangulation~\cite{DBLP:conf/iccv/RydellST23}—using tools from intersection theory~\cite{3264}.


\begin{table}
    \centering
    \renewcommand{\arraystretch}{0.9}
    \begin{tabular}{c|c c c|c c c}
          & \multicolumn{3}{c|}{Approximation 1} & \multicolumn{3}{c}{Approximation 2} \\
        K & m=2 & m=3 & m=4 & m=2 & m=3 & m=4 \\
        \hline
        1 & 120 & 22 & 2 & 20 & 20 & 8 \\
        2 & 426 & 120 & 36 & 122 & 102 & 36 \\
        3 & 732 & 210 & 64 & 276 & 196 & 64 \\
        4 & 1038 & 300 & 92 & 430 & 290 & 92 \\
        5 & 1344 & 390 & 120 & 584 & 384 & 120 \\
        6 & 1650 & 480 & 148 & 738 & 478 & 148 \\
    \end{tabular}
    \vspace{-0.2cm}
    \caption{Numbers of solutions of the minimal problems for estimating translational and angular velocity from asynchronous tracks. Here, $m$ denotes the track length, and $K$ the degree of the rotation approximation. See Section~\ref{sec:minimal_problems} for details.
}
    \label{tab:degrees_appx1}
\end{table}

\section{Minimal Solvers} \label{sec:solvers}

In this section, we propose minimal solvers for selected problems enumerated in Sec.~\ref{sec:minimal_problems}. Our focus is on problems with a small number of solutions, which are typically more efficient and numerically stable. 

We propose solvers based on two approaches: Gr\"obner basis and homotopy continuation. All Gr\"obner basis solvers are generated using the solver generator~\cite{DBLP:conf/cvpr/LarssonAO17}, while the solvers based on homotopy continuation are implemented within the MINUS framework~\cite{DBLP:conf/cvpr/FabbriDFRPTWHGK20}. The solvers are written in C++, and all runtimes were benchmarked on an AMD Ryzen 9 CPU with 3.9~GHz.

For brevity, we denote Approximation~1 by A1 and Approximation~2 by A2 throughout  this section.


\subsection{Problems with m=2, n=5} \label{sec:m2n5}

We now present minimal solvers for the problems with $n=5$ points and $m=2$ observations per point. We begin with the case $K=1$ and A2, since this problem has a relatively low number of $20$ solutions. We model it using five constraints of the form~\eqref{eq:bifocal_constraint_A2}, with $i \in \{1,\ldots,5\}$, $j=1$, and $j'=2$. Because the translational velocity $\mathbf{V}$ can only be recovered up to scale, its last element was fixed to $1$ before feeding the system into the automatic generator~\cite{DBLP:conf/cvpr/LarssonAO17}. The resulting solver has an elimination template with $190$ rows and a runtime of about $470\,\mu s$.  

Since the automatic generator often produces more efficient solvers when the number of variables is reduced, we further eliminated $\mathbf{V}$ using the hidden variable approach: The constraints~\eqref{eq:bifocal_constraint_A2} are linear and homogeneous in $\mathbf{V}$, and can therefore be rewritten as
\begin{equation}
    c_{i,1}(\mathbf{v}) \cdot \mathbf{V}_{(1)} + c_{i,2}(\mathbf{v}) \cdot \mathbf{V}_{(2)} + c_{i,3}(\mathbf{v}) \cdot \mathbf{V}_{(3)} = 0,
\end{equation}
where $i \in \{1,\ldots,5\}$ is the point index, and $c_{i,1}, c_{i,2}, c_{i,3}$ are polynomials in $\mathbf{v}$. By selecting three points $i,i',i''$, we construct the matrix
\begin{equation}
    \mathbf{M}_{i,i',i''}(\mathbf{v}) = 
    \begin{bmatrix}
        c_{i,1}(\mathbf{v}) & c_{i,2}(\mathbf{v}) & c_{i,3}(\mathbf{v}) \\
        c_{i',1}(\mathbf{v}) & c_{i',2}(\mathbf{v}) & c_{i',3}(\mathbf{v}) \\
        c_{i'',1}(\mathbf{v}) & c_{i'',2}(\mathbf{v}) & c_{i'',3}(\mathbf{v})
    \end{bmatrix},
\end{equation}
which satisfies
\begin{equation}
    \mathbf{M}_{i,i',i''}(\mathbf{v}) \cdot \mathbf{V} = \mathbf{0},
\end{equation}
\begin{equation}
    \det \mathbf{M}_{i,i',i''}(\mathbf{v}) = 0. \label{eq:hidden_variable}
\end{equation}
Collecting all $10$ equations of the form~\eqref{eq:hidden_variable}, we fed them into the automatic generator. The resulting solver has an elimination template with only $36$ rows and a runtime of approximately $120\,\mu s$.  

In addition, we developed solvers for the problems with $K=1$+A1, and with $K=2$+A2. Due to the high algebraic degrees of these problems ($120$ and $122$), these solvers are based on homotopy continuation. Both are formulated using five constraints of the form~\eqref{eq:bifocal_constraint}. The runtime of the solver for $K=1$+A1 is about $180\,\text{ms}$, while that of the solver for $K=2$+A2 is about $210\,\text{ms}$.  

For comparison, we also constructed a homotopy continuation solver for the problem $K=1$, A2, using the same formulation as the previous two solvers, achieving a runtime of about $5.2\,\text{ms}$.

\subsection{Problems with m=3, n=2}

We now present minimal solvers for the problems with $n=2$ points and $m=3$ observations per point. We begin with the case $K=1$, where we developed Gr\"obner basis solvers for both A1 and A2. These problems are formulated using the bifocal constraints~\eqref{eq:bifocal_constraint} and the trifocal constraints~\eqref{eq:trifocal_constraint}. As before, the last element of $\mathbf{V}$ is fixed due to scale ambiguity.  

Since the problems are overconstrained, we omit the bifocal constraints involving the observation $\mathbf{p}_{2,3}$, and retain only the first column from the second trifocal constraint. Geometrically, this enforces the existence of a point $\mathbf{X}_2 \in \RR^3$ that projects onto $\mathbf{p}_{2,1}$ at time $t_{2,1}$, onto $\mathbf{p}_{2,2}$ at time $t_{2,2}$, and onto the horizontal line passing through $\mathbf{p}_{2,3}$ at time $t_{2,3}$. With this relaxation, the system becomes minimal and can be solved using an automatic solver generator.  

The minimal solver for A1 has an elimination template with $363$ rows and achieves a runtime of $2.1\,\text{ms}$. The solver for A2 has an elimination template with $387$ rows and a runtime of $2.3\,\text{ms}$. We also experimented with the hidden variable approach described in Sec.~\ref{sec:m2n5}, but the coefficients used for building the elimination template grew prohibitively large, making the resulting solvers too unstable for practical use.  

In addition, we developed homotopy continuation minimal solvers for both $K=1$ and $K=2$.
These solvers use the original formulations~\eqref{eq:appx1_cons} and~\eqref{eq:appx2_cons} with variables $\mathbf{v}$, $\mathbf{V}$, $\mathbf{X}_1$, and $\mathbf{X}_2$. To enforce minimality, only the first element of the constraint involving $\mathbf{p}_{2,3}$ is retained. The runtimes of these solvers are presented in Table~\ref{tab:solvers}.

\subsection{Problems with m=4, n=1}

We now present minimal solvers for the problems with $n=1$ point and $m=4$ observations of this point. We first developed Gr\"obner basis solvers for the problems $K=1$+A1, $K=1$+A2, and $K=2$+A2. These solvers are based on the formulation using both the bifocal constraints~\eqref{eq:bifocal_constraint} and the trifocal constraints~\eqref{eq:trifocal_constraint}, with the last element of $\mathbf{V}$ fixed to resolve scale ambiguity. The solver for $K=1$+A1 has an elimination template with $73$ rows and a runtime of $81\,\mu s$, while the solver for $K=1$+A2 has $103$ rows and runs in $100\,\mu s$. The solver for $K=2$+A2 is substantially larger, with an elimination template of $1073$ rows and a runtime of $31\,\text{ms}$.  

We also attempted to design a Gr\"obner basis solver for the $K=2$+A1 problem, which also has a relatively low algebraic degree. However, the resulting solver failed to compile, likely due to the large size of the coefficients. Similarly, we explored the hidden variable approach in order to simplify the solvers, but in this case the generator was unable to find a suitable elimination template within a reasonable time.  

In addition, we developed homotopy continuation-based minimal solvers for the problems $K=1$+A1, $K=1$+A2, $K=2$+A1, and $K=2$+A2. These solvers employ the original formulations~\eqref{eq:appx1_cons} and~\eqref{eq:appx2_cons}, with variables $\mathbf{v}$, $\mathbf{V}$, and $\mathbf{X}_1$. The runtimes of these homotopy continuation solvers are reported in Table~\ref{tab:solvers}.


\begin{table}[]
    \centering
    \renewcommand{\arraystretch}{0.8}
    \begin{tabular}{c|c|c|c||c|c||c}
        m,n & K & appx & type & \#vars& \#sols & time ($\mu s$)  \\
        \hline
        2,5 & 1 & A1 & HC & 5 & 120 & 1.8E5 \\
        2,5 & 1 & A2 & HC & 5 & 20 & 5.2E3 \\
        2,5 & 1 & A2 & GB & 5 & 20 & 4.7E2 \\
        2,5 & 1 & A2 & GB & 3 & 20 & \textbf{1.2E2} \\
        \hline
        2,5 & 2 & A2 & HC & 5 & 122 & 2.1E5 \\
        \hline
        \hline
        3,2 & 1 & A1 & HC & 11 & 22 & 1.6E4 \\
        3,2 & 1 & A1 & GB & 5 & 22 & \textbf{2.1E3} \\
        3,2 & 1 & A2 & HC & 11 & 20 & 1.2E4 \\
        3,2 & 1 & A2 & GB & 5 & 20 & 2.3E3 \\
        \hline
        3,2 & 2 & A1 & HC & 11 & 120 & 3.4E5 \\
        3,2 & 2 & A2 & HC & 11 & 120 & 2.1E5 \\
        \hline
        \hline
        4,1 & 1 & A1 & HC & 5 & 2 & 6.7E2 \\
        4,1 & 1 & A1 & GB & 5 & 8 & \textbf{8.1E1} \\
        4,1 & 1 & A2 & HC & 8 & 8 & 3.2E3 \\
        4,1 & 1 & A2 & GB & 8 & 8 &  1.0E2  \\
        \hline
        4,2 & 2 & A1 & HC & 8 & 36 & 6.6E4 \\
        4,2 & 2 & A2 & HC & 8 & 36 & 2.8E4 \\
        4,2 & 2 & A2 & GB & 5 & 36 & 3.1E4 \\
        
    \end{tabular}
    \caption{Overview of the minimal solvers presented in Section~\ref{sec:solvers}. Here, $m$ and $n$ denote the track length and number of tracks; $K$ is the maximal degree of the rotation approximation; \textit{appx} indicates the type of approximation (Sec.~\ref{sec:appx}); and \textit{type} specifies the solver method, either HC (Homotopy Continuation) or GB (Gr\"obner Basis). \#vars denotes the number of estimated variables (\textit{i.e.}, not eliminated beforehand). Runtimes are reported in microseconds, with the lowest time for each $(m,n)$ highlighted in bold.
    }
    \label{tab:solvers}
\end{table}

\section{Local Optimization} \label{sec:lo}

We now introduce the local optimization procedure used to refine the output of the minimal solvers presented in Section~\ref{sec:solvers}. This refinement has two main benefits. First, it reduces the effect of noise by incorporating a larger number of correspondences. Second, unlike the minimal solvers, which rely on the approximate rotation model~\eqref{eq:taylor_finite}, the local optimization employs the exact rotation model~\eqref{eq:orientation_exact}, thereby mitigating the errors caused by the approximation. This local optimization is based on the Ceres framework~\cite{ceres}, although efficiency can be further improved by employing the Poselib framework~\cite{PoseLib}.  

The optimization is initialized with estimates $\mathbf{v}_0$ of the Euler vector and $\mathbf{V}_0$ of the translational velocity obtained with a minimal solver. From these, the initial rotation axis $\mathbf{a}_0$ and angular velocity $\omega_0$ are obtained as $\mathbf{a}_0 = \frac{1}{\lVert \mathbf{v}_0 \rVert} \cdot \mathbf{v}_0$ and $\omega_0 = \lVert \mathbf{v}_0 \rVert$. For every pair of correspondences $\mathbf{p}_{i,j}$ and $\mathbf{p}_{i,j'}$, captured at times $t_{i,j}$ and $t_{i,j'}$, respectively, we construct an essential matrix  
\begin{equation}
    \mathbf{E}_{i,j,j'} = \mathbf{R}(t_{i,j'}) \,[\mathbf{V}]_{\times} \,\mathbf{R}(t_{i,j'})^T.
\end{equation}
The local optimization minimizes the sum of squared Sampson errors
\begin{equation}
    \frac{\mathbf{p}_{i,j'}^T \mathbf{E}_{i,j,j'} \mathbf{p}_{i,j} }{\sqrt{ \lVert (\mathbf{E}_{i,j'} \mathbf{p}_{i,j})_{(1:2)} \rVert^2 + \lVert (\mathbf{E}_{i,j'}^T \mathbf{p}_{i,j'})_{(1:2)} \rVert^2 }},
\end{equation}
over all correspondences that are inliers to the initial motion estimate.

\section{Experiments} \label{sec:experiments}

In this section, we evaluate the minimal solvers introduced in Section~\ref{sec:solvers} and summarized in Table~\ref{tab:solvers}. We first present synthetic experiments to assess numerical stability (Sec.~\ref{sec:stability}) and robustness to noise as well as to high angular velocities (Sec.~\ref{sec:robustness}). Finally, we validate the solvers in real-world experiments (Sec.~\ref{sec:real}) using the rolling shutter datasets Fastec and Carla~\cite{Liu2020CVPR}.


\subsection{Numerical Stability} \label{sec:stability}
To evaluate the numerical stability of the proposed solvers, we synthetically generated instances of the corresponding minimal problems, using the approximate rotation $\mathbf{R}_K(\mathbf{v})$ instead of the exact rotation $\mathbf{R}(\mathbf{v})$. For each problem instance, we sampled an Euler vector $\mathbf{v}_{GT}$ from the normal distribution with $\mu = \mathbf{0}$ and $\sigma = 0.1$, and a translation vector $\mathbf{V}_{GT}$ from a normal distribution with $\mu = \mathbf{0}$ and $\sigma = 1$. We further sampled $n$ points $\mathbf{X}_i$ from a normal distribution with $\mu = [0 \ 0 \ 2]^T$ and $\sigma = 1$, and, for each point, we generated $m$ capture times $t_{i,j}$ from a normal distribution with $\mu = 0$ and $\sigma = 1$.  

The image points $\mathbf{p}_{i,j}$ were obtained by projecting $\mathbf{X}_i$ according to~\eqref{eq:appx1_proj} for A1, and according to~\eqref{eq:appx2_proj} for A2. Each solver was applied to recover an estimate of the Euler vector $\mathbf{v}_{est}$ and the translation vector $\mathbf{V}_{est}$. To assess accuracy, we constructed the rotation matrices $\mathbf{R}(\mathbf{v}_{GT})$ and $\mathbf{R}(\mathbf{v}_{est})$ at time $t=1$, and measured the rotation error as the angle of the relative rotation $\mathbf{R}(\mathbf{v}_{est})^T \mathbf{R}(\mathbf{v}_{GT})$. The translation error was computed as the angle between $\mathbf{V}_{GT}$ and $\mathbf{V}_{est}$.  

\begin{figure}[!]
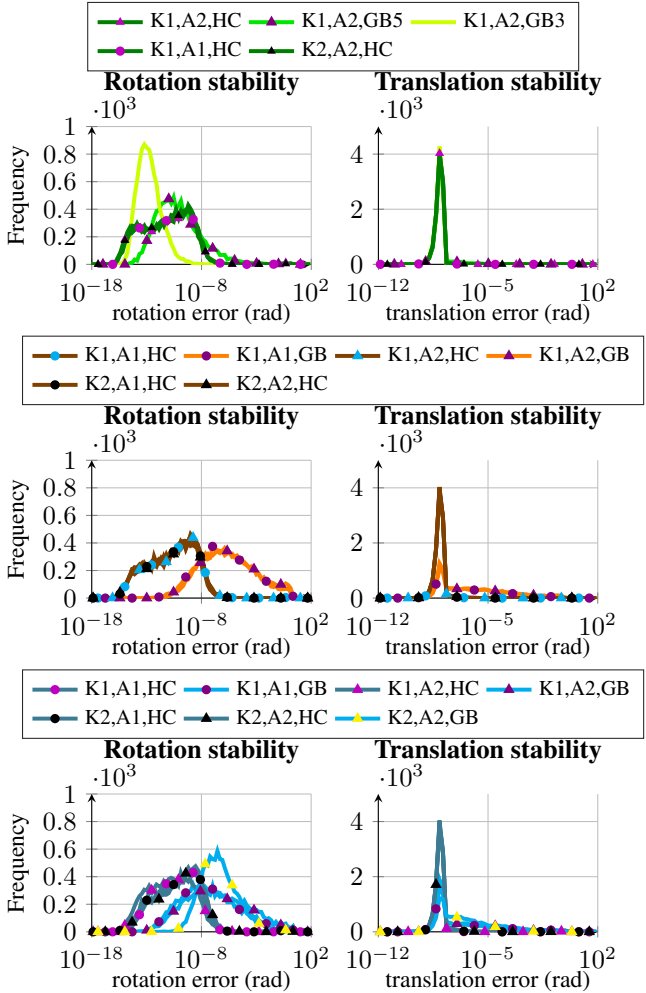

    \centering
    \begin{tikzpicture}

\input{plot/rotation_stability}
\input{plot/translation_stability}

\end{tikzpicture}
    \begin{tikzpicture}

\input{plot/rotation_stability_3v}
\input{plot/translation_stability_3v}

\end{tikzpicture}
    \begin{tikzpicture}

\input{plot/rotation_stability_4v}
\input{plot/translation_stability_4v}

\end{tikzpicture}
    \vspace{-0.7cm}
    \caption{\textbf{Stability test.} Histograms of rotation (\textit{left}) and translation (\textit{right}) errors, computed over $10^4$ noiseless samples. Results are shown for solvers with \textit{top:} $m=2,n=5$, \textit{middle:} $m=3,n=2$, and \textit{bottom:} $m=4,n=1$. K1 and K2 denote $K=1$ and $K=2$, respectively, while A1 and A2 indicate the type of approximation (Sec.~\ref{sec:appx}).}
    \label{fig:stability}
\end{figure}

The results, depicted in Figure~\ref{fig:stability}, show that all solvers exhibit sufficient numerical stability for use in standard pose estimation pipelines. However, the Gr\"obner basis solvers for the cases with $m=3$ and $m=4$ display somewhat reduced stability, which may negatively affect performance in some cases. 

\subsection{Robustness to Angular Velocity and Noise} \label{sec:robustness}
We now evaluate the solvers in a more realistic scenario, where the image projections are generated using the exact rotation $\mathbf{R}(\mathbf{v})$. Since the solvers instead rely on the approximation $\mathbf{R}_K(\mathbf{v})$ with $K=1$ or $K=2$, the recovered motion parameters inevitably exhibit nonzero error, even in the absence of projection noise. This experiment therefore assesses the accuracy of the approximations underlying the solvers.  

The experimental setup follows the same procedure as in the previous section, except that the angular velocity $\omega$ is treated as a controllable parameter rather than sampled randomly. We consider two different cases: the \textit{event camera} and the \textit{rolling shutter camera}. In the event camera case, the capture times are sampled randomly, as in Section~\ref{sec:stability}, and then ordered increasingly. In the rolling shutter case, the capture times are determined by jointly estimating the projection $\mathbf{p}_{i,j}$ and the capture time $t_{i,j}$ such that~\eqref{eq:capture_time_rs} holds. Since this is not a polynomial system, we locate the correct $t_{i,j}$ by bisection.  

To increase realism, we assume a focal length of $f=700 \, \text{px}$ and add Gaussian noise with $\sigma = (1 \, \text{px})/f$ to each projection $\mathbf{p}_{i,j}$. In the rolling shutter case, the capture times $t_{i,j}$ are recovered using the noisy projections. We evaluate the rotation error at the time corresponding to the center of the second frame and the translation error as in the previous section. The resulting errors, plotted as functions of angular velocity, are shown in Figure~\ref{fig:appx_tests} for the rolling shutter case and in Figure~\ref{fig:appx_tests_event} for the event camera case. The noise-free variant of the experiment is included in the Appendix.

In this evaluation, we compare our solvers against the five-point solver~\cite{DBLP:journals/pami/Nister04}. When $\omega = 0$, this solver is able to recover the translational velocity $\mathbf{V}$. 
The results confirm that the errors of all solvers increase with angular velocity, as expected, since the approximation becomes less accurate for higher $\omega$. In the rolling shutter scenario, the five-point solver performs well for low angular velocities, but in the event camera setting its error grows rapidly as $\omega$ increases. Overall, most of the proposed solvers demonstrate greater robustness than the five-point solver. In particular, solvers based on quadratic formulations with $m=2$ or $m=3$ exhibit the highest robustness, while those assuming $m=4$ tend to be less stable and appear to be degenerate in the special case of $\omega=0$.

\begin{figure}
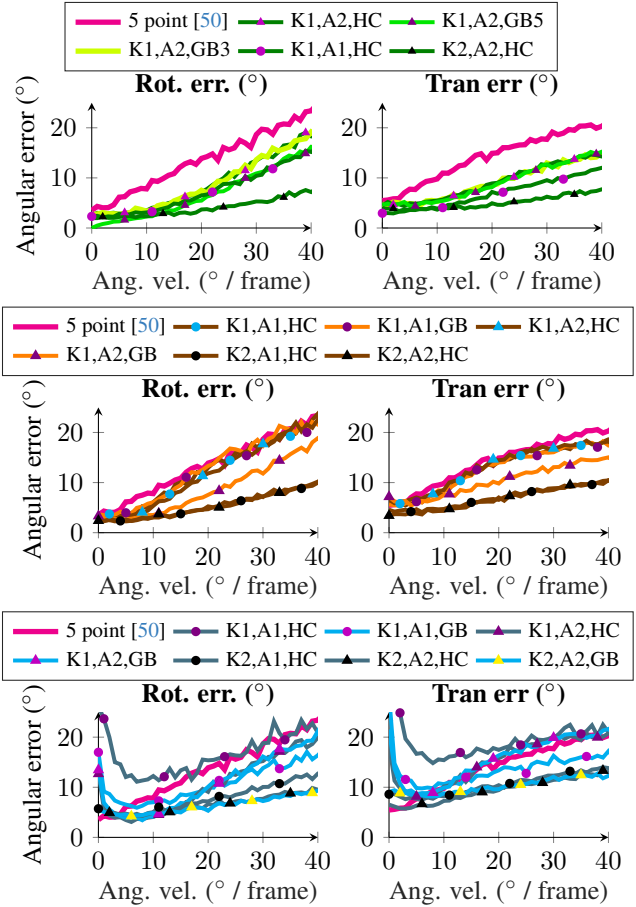

    \centering
    \begin{tikzpicture}

\begin{axis}[%
width=0.35\columnwidth,
height=0.20\columnwidth,
at={(0.879in,0.389in)},
scale only axis,
xmin=0,
xmax=40,
xlabel style={font=\color{white!15!black}, yshift=0.03in},
xlabel={Ang. vel. ($^{\circ}$ / frame)},
title={Rot. err. ($^{\circ}$)},
ymin=0,
ymax=25,
ymode=normal,
yminorticks=true,
axis lines = left,
axis background/.style={fill=white},
title style={font=\bfseries, yshift=-0.1in},
ylabel style={yshift=-0.15in},
ylabel={Angular error ($^{\circ}$)},
legend style={at={(1,1.25)}, anchor=south, legend cell align=left, align=left, draw=white!15!black, font=\footnotesize,nodes={scale=1.0, transform shape}},
legend columns=3
]


\addplot [color=magenta,line width=2pt, mark options={solid, red}]
  table[row sep=crcr]{%
0.0 3.52444 \\
1.0 4.31173 \\
2.0 4.0343 \\
3.0 4.0706 \\
4.0 4.60859 \\
5.0 6.15181 \\
6.0 6.24793 \\
7.0 7.21825 \\
8.0 7.49083 \\
9.0 7.90129 \\
10.0 7.79831 \\
11.0 9.47014 \\
12.0 9.29346 \\
13.0 9.86081 \\
14.0 10.5163 \\
15.0 10.8548 \\
16.0 12.0643 \\
17.0 12.8066 \\
18.0 13.2233 \\
19.0 12.7013 \\
20.0 13.8905 \\
21.0 14.2373 \\
22.0 14.777 \\
23.0 13.5278 \\
24.0 16.0616 \\
25.0 15.6041 \\
26.0 15.9871 \\
27.0 17.4976 \\
28.0 16.9479 \\
29.0 16.0874 \\
30.0 18.643 \\
31.0 19.7081 \\
32.0 19.579 \\
33.0 18.7268 \\
34.0 20.4989 \\
35.0 20.766 \\
36.0 20.6674 \\
37.0 22.2542 \\
38.0 21.4987 \\
39.0 23.1913 \\
40.0 23.3862 \\
41.0 24.5823 \\
42.0 23.3445 \\
43.0 23.3269 \\
44.0 23.8749 \\
45.0 24.4596 \\
};
\addlegendentry{5 point \cite{DBLP:journals/pami/Nister04}}


\addplot [color=green!50!black,line width=1.5pt, mark options={solid, violet!50!magenta}, mark=triangle*, mark repeat=11, mark phase=7, mark size=0.5pt]
  table[row sep=crcr]{%
0.0 2.53941 \\
1.0 3.02195 \\
2.0 2.91189 \\
3.0 2.95136 \\
4.0 3.05318 \\
5.0 2.80095 \\
6.0 3.06832 \\
7.0 3.06618 \\
8.0 3.31624 \\
9.0 3.58435 \\
10.0 3.50469 \\
11.0 4.03292 \\
12.0 4.25024 \\
13.0 4.58291 \\
14.0 4.98721 \\
15.0 5.197 \\
16.0 5.53548 \\
17.0 6.23538 \\
18.0 6.47331 \\
19.0 6.73645 \\
20.0 7.21683 \\
21.0 7.78545 \\
22.0 8.58447 \\
23.0 9.29294 \\
24.0 9.94923 \\
25.0 11.0341 \\
26.0 10.1935 \\
27.0 11.3808 \\
28.0 11.4616 \\
29.0 12.9944 \\
30.0 13.4363 \\
31.0 14.1422 \\
32.0 14.4227 \\
33.0 14.1264 \\
34.0 15.1016 \\
35.0 16.4705 \\
36.0 16.6455 \\
37.0 16.1139 \\
38.0 18.5285 \\
39.0 19.0303 \\
40.0 18.3024 \\
41.0 18.9957 \\
42.0 19.1734 \\
43.0 19.9511 \\
44.0 21.2413 \\
45.0 20.7601 \\
};
\addlegendentry{K1,A2,HC}

\addplot [color=green!90!black,line width=1.5pt, mark options={solid, violet}, mark=triangle*, mark repeat=11, mark phase=7, mark size=0.5pt]
  table[row sep=crcr]{%
0.0 0.0 \\
1.0 0.48075 \\
2.0 0.795428 \\
3.0 1.00154 \\
4.0 1.27107 \\
5.0 1.322 \\
6.0 1.60099 \\
7.0 1.75909 \\
8.0 1.9471 \\
9.0 2.06729 \\
10.0 2.32379 \\
11.0 2.34309 \\
12.0 2.86185 \\
13.0 2.91434 \\
14.0 3.50492 \\
15.0 3.50733 \\
16.0 3.99682 \\
17.0 4.54144 \\
18.0 4.9887 \\
19.0 5.38321 \\
20.0 5.46098 \\
21.0 6.27 \\
22.0 6.57939 \\
23.0 7.24977 \\
24.0 7.33172 \\
25.0 7.8931 \\
26.0 9.12663 \\
27.0 9.33789 \\
28.0 9.92615 \\
29.0 10.441 \\
30.0 10.8888 \\
31.0 11.1756 \\
32.0 11.556 \\
33.0 12.3119 \\
34.0 12.7436 \\
35.0 13.8205 \\
36.0 14.4592 \\
37.0 14.2243 \\
38.0 15.2686 \\
39.0 14.8528 \\
40.0 16.2105 \\
41.0 16.1379 \\
42.0 17.599 \\
43.0 18.6405 \\
44.0 20.1729 \\
45.0 19.7089 \\
   }; 
\addlegendentry{K1,A2,GB5}

\addplot [color=green!20!yellow,line width=2pt, mark options={solid, red}]
  table[row sep=crcr]{%
0.0 2.62845 \\
1.0 2.78948 \\
2.0 3.07247 \\
3.0 2.85311 \\
4.0 2.85363 \\
5.0 2.74231 \\
6.0 2.84488 \\
7.0 3.03259 \\
8.0 3.94758 \\
9.0 3.68957 \\
10.0 3.36321 \\
11.0 3.60831 \\
12.0 4.43152 \\
13.0 4.74057 \\
14.0 4.60547 \\
15.0 5.41716 \\
16.0 5.8349 \\
17.0 6.36101 \\
18.0 6.74639 \\
19.0 6.75112 \\
20.0 7.52668 \\
21.0 6.98607 \\
22.0 8.19771 \\
23.0 9.32405 \\
24.0 9.28919 \\
25.0 10.0191 \\
26.0 11.1923 \\
27.0 11.2469 \\
28.0 12.5181 \\
29.0 11.8499 \\
30.0 13.6191 \\
31.0 14.4781 \\
32.0 14.1825 \\
33.0 15.9289 \\
34.0 15.3892 \\
35.0 15.7572 \\
36.0 16.1461 \\
37.0 17.3783 \\
38.0 17.6312 \\
39.0 17.8302 \\
40.0 19.1644 \\
41.0 18.4464 \\
42.0 20.3534 \\
43.0 18.8358 \\
44.0 20.0788 \\
45.0 21.094 \\
};
\addlegendentry{K1,A2,GB3}


\addplot [color=green!50!black,line width=1.5pt, mark options={solid, violet!50!magenta}, mark=*, mark repeat=11, mark size=1pt]
  table[row sep=crcr]{%
0.0 2.32337 \\
1.0 2.42596 \\
2.0 2.1163 \\
3.0 2.32164 \\
4.0 2.24159 \\
5.0 2.84551 \\
6.0 2.62851 \\
7.0 2.76028 \\
8.0 3.12355 \\
9.0 3.00263 \\
10.0 3.11055 \\
11.0 3.26469 \\
12.0 3.58837 \\
13.0 3.87867 \\
14.0 4.17472 \\
15.0 4.86991 \\
16.0 4.78156 \\
17.0 5.4723 \\
18.0 5.48739 \\
19.0 6.01713 \\
20.0 6.50139 \\
21.0 6.68475 \\
22.0 7.17544 \\
23.0 7.99422 \\
24.0 8.77223 \\
25.0 8.83445 \\
26.0 9.02782 \\
27.0 8.73973 \\
28.0 10.2879 \\
29.0 9.60394 \\
30.0 10.4118 \\
31.0 10.6218 \\
32.0 12.0909 \\
33.0 11.8224 \\
34.0 12.7794 \\
35.0 13.2825 \\
36.0 13.5298 \\
37.0 14.1537 \\
38.0 14.1354 \\
39.0 15.4283 \\
40.0 15.1717 \\
41.0 16.033 \\
42.0 16.4529 \\
43.0 16.8688 \\
44.0 17.6712 \\
45.0 17.5988 \\
   }; 
\addlegendentry{K1,A1,HC}

\addplot [color=green!50!black,line width=1.5pt, mark options={solid,violet, draw=black,fill=violet}, mark=triangle*, mark repeat=11, mark size=0.4pt, mark phase=3]
  table[row sep=crcr]{%
0.0 1.97909 \\
1.0 1.97067 \\
2.0 2.23571 \\
3.0 2.02554 \\
4.0 2.02999 \\
5.0 2.2006 \\
6.0 2.35354 \\
7.0 2.39577 \\
8.0 2.12574 \\
9.0 2.43842 \\
10.0 2.3375 \\
11.0 2.49215 \\
12.0 2.7161 \\
13.0 2.87773 \\
14.0 3.06728 \\
15.0 2.58326 \\
16.0 3.02414 \\
17.0 3.09683 \\
18.0 2.85016 \\
19.0 3.08529 \\
20.0 3.69968 \\
21.0 3.76278 \\
22.0 3.49879 \\
23.0 4.10506 \\
24.0 4.17371 \\
25.0 4.35954 \\
26.0 4.36857 \\
27.0 4.71314 \\
28.0 4.81176 \\
29.0 5.04088 \\
30.0 5.29946 \\
31.0 5.2831 \\
32.0 5.17039 \\
33.0 5.47405 \\
34.0 6.33204 \\
35.0 6.11364 \\
36.0 7.10819 \\
37.0 6.41473 \\
38.0 7.13914 \\
39.0 7.57103 \\
40.0 7.17562 \\
41.0 7.84705 \\
42.0 8.11091 \\
43.0 8.58085 \\
44.0 8.3074 \\
45.0 9.44545 \\
   }; 
\addlegendentry{K2,A2,HC}

\end{axis}

\input{plot/noise_test_t}

\end{tikzpicture}%
    \begin{tikzpicture}

\begin{axis}[%
width=0.35\columnwidth,
height=0.20\columnwidth,
at={(0.879in,0.389in)},
scale only axis,
xmin=0,
xmax=40,
xlabel style={font=\color{white!15!black}, yshift=0.03in},
xlabel={Ang. vel. ($^{\circ}$ / frame)},
title={Rot. err. ($^{\circ}$)},
ymin=0,
ymax=25,
ymode=normal,
yminorticks=true,
axis lines = left,
axis background/.style={fill=white},
title style={font=\bfseries, yshift=-0.1in},
ylabel style={yshift=-0.15in},
ylabel={Angular error ($^{\circ}$)},
legend style={at={(1,1.25)}, anchor=south, legend cell align=left, align=left, draw=white!15!black, font=\footnotesize,nodes={scale=1.0, transform shape}},
legend columns=4
]


\addplot [color=magenta,line width=2pt, mark options={solid, red}]
  table[row sep=crcr]{%
0.0 3.52444 \\
1.0 4.31173 \\
2.0 4.0343 \\
3.0 4.0706 \\
4.0 4.60859 \\
5.0 6.15181 \\
6.0 6.24793 \\
7.0 7.21825 \\
8.0 7.49083 \\
9.0 7.90129 \\
10.0 7.79831 \\
11.0 9.47014 \\
12.0 9.29346 \\
13.0 9.86081 \\
14.0 10.5163 \\
15.0 10.8548 \\
16.0 12.0643 \\
17.0 12.8066 \\
18.0 13.2233 \\
19.0 12.7013 \\
20.0 13.8905 \\
21.0 14.2373 \\
22.0 14.777 \\
23.0 13.5278 \\
24.0 16.0616 \\
25.0 15.6041 \\
26.0 15.9871 \\
27.0 17.4976 \\
28.0 16.9479 \\
29.0 16.0874 \\
30.0 18.643 \\
31.0 19.7081 \\
32.0 19.579 \\
33.0 18.7268 \\
34.0 20.4989 \\
35.0 20.766 \\
36.0 20.6674 \\
37.0 22.2542 \\
38.0 21.4987 \\
39.0 23.1913 \\
40.0 23.3862 \\
41.0 24.5823 \\
42.0 23.3445 \\
43.0 23.3269 \\
44.0 23.8749 \\
45.0 24.4596 \\
};
\addlegendentry{5 point \cite{DBLP:journals/pami/Nister04}}



\addplot [color=orange!50!black,line width=1.5pt, mark options={solid, orange}, mark options={solid, cyan}, mark=*, mark repeat=11, mark phase = 3, mark size=1pt]
  table[row sep=crcr]{%
0.0 3.49591 \\
1.0 3.10145 \\
2.0 3.73004 \\
3.0 3.2781 \\
4.0 3.59196 \\
5.0 3.77715 \\
6.0 3.97425 \\
7.0 4.33099 \\
8.0 5.28816 \\
9.0 4.91346 \\
10.0 6.10745 \\
11.0 7.06346 \\
12.0 8.01295 \\
13.0 7.70788 \\
14.0 9.57878 \\
15.0 10.0761 \\
16.0 9.71765 \\
17.0 10.3623 \\
18.0 12.0754 \\
19.0 11.6499 \\
20.0 12.9444 \\
21.0 13.935 \\
22.0 14.5244 \\
23.0 14.1736 \\
24.0 14.4482 \\
25.0 15.9195 \\
26.0 16.1981 \\
27.0 15.9175 \\
28.0 17.7807 \\
29.0 17.0119 \\
30.0 18.2816 \\
31.0 17.4952 \\
32.0 19.0593 \\
33.0 18.4658 \\
34.0 19.9008 \\
35.0 19.234 \\
36.0 19.8035 \\
37.0 21.6561 \\
38.0 20.3569 \\
39.0 22.5565 \\
40.0 23.7467 \\
41.0 22.3342 \\
42.0 23.2366 \\
43.0 22.9145 \\
44.0 23.7844 \\
45.0 24.3418 \\
  };
\addlegendentry{K1,A1,HC}

\addplot [color=orange,line width=1.5pt, mark options={solid, orange}, mark options={solid, violet}, mark=*, mark repeat=11, mark phase = 6, mark size=1pt]
  table[row sep=crcr]{%
0.0 3.64952 \\
1.0 3.51868 \\
2.0 3.37602 \\
3.0 3.57506 \\
4.0 4.18499 \\
5.0 3.97703 \\
6.0 3.79136 \\
7.0 4.96596 \\
8.0 5.22668 \\
9.0 5.78411 \\
10.0 6.07794 \\
11.0 6.70377 \\
12.0 7.28025 \\
13.0 7.85667 \\
14.0 8.8691 \\
15.0 9.48996 \\
16.0 11.0653 \\
17.0 12.0925 \\
18.0 11.9139 \\
19.0 12.5595 \\
20.0 13.3266 \\
21.0 13.7525 \\
22.0 14.9072 \\
23.0 15.769 \\
24.0 16.8001 \\
25.0 16.6449 \\
26.0 16.3124 \\
27.0 15.3652 \\
28.0 17.6265 \\
29.0 18.538 \\
30.0 17.3208 \\
31.0 18.7321 \\
32.0 19.2138 \\
33.0 21.3205 \\
34.0 20.8838 \\
35.0 20.6479 \\
36.0 19.9658 \\
37.0 22.0517 \\
38.0 20.0108 \\
39.0 22.0859 \\
40.0 21.56 \\
41.0 22.1615 \\
42.0 22.1915 \\
43.0 22.7695 \\
44.0 21.7558 \\
45.0 23.2729 \\
  };
\addlegendentry{K1,A1,GB}

\addplot [color=orange!50!black,line width=1.5pt, mark options={solid, orange}, mark options={solid, cyan}, mark=triangle*, mark repeat=11, mark phase = 9, mark size=1pt]
  table[row sep=crcr]{%
0.0 2.72967 \\
1.0 2.72273 \\
2.0 2.91356 \\
3.0 2.76019 \\
4.0 3.04632 \\
5.0 3.65348 \\
6.0 3.42049 \\
7.0 3.78221 \\
8.0 4.06701 \\
9.0 4.39675 \\
10.0 5.34828 \\
11.0 6.2632 \\
12.0 6.36041 \\
13.0 6.76643 \\
14.0 7.95079 \\
15.0 8.25267 \\
16.0 9.14764 \\
17.0 9.96398 \\
18.0 10.198 \\
19.0 11.3358 \\
20.0 11.8258 \\
21.0 12.1783 \\
22.0 12.4018 \\
23.0 13.3113 \\
24.0 14.246 \\
25.0 14.9212 \\
26.0 15.5867 \\
27.0 16.2721 \\
28.0 15.5524 \\
29.0 16.4237 \\
30.0 17.7011 \\
31.0 17.5137 \\
32.0 18.1635 \\
33.0 19.5453 \\
34.0 19.9238 \\
35.0 20.4607 \\
36.0 21.9821 \\
37.0 20.2268 \\
38.0 21.639 \\
39.0 20.9392 \\
40.0 22.6131 \\
41.0 21.9982 \\
42.0 22.4546 \\
43.0 22.9258 \\
44.0 22.667 \\
45.0 24.4495 \\
  };
\addlegendentry{K1,A2,HC}

\addplot [color=orange,line width=1.5pt, mark options={solid, orange}, mark options={solid, violet}, mark=triangle*, mark repeat=11, mark size=1pt]
  table[row sep=crcr]{%
0.0 3.31499 \\
1.0 2.9594 \\
2.0 2.48702 \\
3.0 2.73195 \\
4.0 2.43265 \\
5.0 2.65819 \\
6.0 2.7968 \\
7.0 2.98697 \\
8.0 3.20932 \\
9.0 3.50819 \\
10.0 3.53285 \\
11.0 3.85583 \\
12.0 3.90656 \\
13.0 4.76702 \\
14.0 5.07322 \\
15.0 5.05221 \\
16.0 5.91711 \\
17.0 6.46281 \\
18.0 6.81043 \\
19.0 7.02427 \\
20.0 7.5526 \\
21.0 8.08469 \\
22.0 8.35949 \\
23.0 9.45967 \\
24.0 9.14195 \\
25.0 10.2953 \\
26.0 10.6046 \\
27.0 11.3431 \\
28.0 12.2788 \\
29.0 12.4241 \\
30.0 11.7944 \\
31.0 12.7789 \\
32.0 14.3655 \\
33.0 14.402 \\
34.0 14.852 \\
35.0 15.6648 \\
36.0 16.3256 \\
37.0 16.9606 \\
38.0 16.2229 \\
39.0 18.0254 \\
40.0 18.8649 \\
41.0 18.2472 \\
42.0 18.2649 \\
43.0 19.3343 \\
44.0 19.3342 \\
45.0 20.0472 \\
  };
\addlegendentry{K1,A2,GB}

\addplot [color=orange!50!black,line width=1.5pt, mark options={solid, orange}, mark options={solid, black}, mark=*, mark repeat=11, mark phase = 5, mark size=1pt]
  table[row sep=crcr]{%
0.0 2.51167 \\
1.0 2.28367 \\
2.0 2.58678 \\
3.0 2.50746 \\
4.0 2.34919 \\
5.0 2.64448 \\
6.0 2.33085 \\
7.0 2.49476 \\
8.0 2.79548 \\
9.0 2.90561 \\
10.0 3.03909 \\
11.0 3.02511 \\
12.0 3.38999 \\
13.0 3.27132 \\
14.0 3.57744 \\
15.0 3.77161 \\
16.0 4.02897 \\
17.0 4.34647 \\
18.0 4.50275 \\
19.0 4.60242 \\
20.0 4.68234 \\
21.0 5.10613 \\
22.0 5.37197 \\
23.0 6.09312 \\
24.0 5.81688 \\
25.0 6.32802 \\
26.0 6.37334 \\
27.0 6.57171 \\
28.0 6.50765 \\
29.0 7.60079 \\
30.0 7.10838 \\
31.0 7.83522 \\
32.0 8.58551 \\
33.0 8.04273 \\
34.0 7.72062 \\
35.0 8.64196 \\
36.0 8.96003 \\
37.0 8.81727 \\
38.0 9.33433 \\
39.0 9.73357 \\
40.0 10.1882 \\
41.0 10.8688 \\
42.0 10.3058 \\
43.0 11.2694 \\
44.0 11.3518 \\
45.0 11.4762 \\
  };
\addlegendentry{K2,A1,HC}

\addplot [color=orange!50!black,line width=1.5pt, mark options={solid, orange}, mark options={solid, black}, mark=triangle*, mark repeat=11, mark size=1pt]
  table[row sep=crcr]{%
0.0 2.39574 \\
1.0 2.39452 \\
2.0 2.51304 \\
3.0 2.25832 \\
4.0 2.38266 \\
5.0 2.25464 \\
6.0 2.3666 \\
7.0 2.61006 \\
8.0 2.98923 \\
9.0 2.75171 \\
10.0 3.09079 \\
11.0 3.71353 \\
12.0 3.40863 \\
13.0 3.04894 \\
14.0 3.38426 \\
15.0 3.95019 \\
16.0 4.06436 \\
17.0 4.23021 \\
18.0 4.46897 \\
19.0 4.55985 \\
20.0 5.25767 \\
21.0 5.04951 \\
22.0 5.04342 \\
23.0 5.26783 \\
24.0 5.39711 \\
25.0 5.72549 \\
26.0 6.04889 \\
27.0 6.78561 \\
28.0 6.77401 \\
29.0 6.86903 \\
30.0 7.26406 \\
31.0 7.48617 \\
32.0 7.85627 \\
33.0 7.9787 \\
34.0 8.6146 \\
35.0 8.47507 \\
36.0 8.80078 \\
37.0 9.40795 \\
38.0 9.20071 \\
39.0 9.35159 \\
40.0 9.9709 \\
41.0 9.67443 \\
42.0 10.2813 \\
43.0 11.2911 \\
44.0 10.8002 \\
45.0 10.8089 \\
  };
\addlegendentry{K2,A2,HC}

\end{axis}

\input{plot/noise_test_t_3v}

\end{tikzpicture}%
    \begin{tikzpicture}

\begin{axis}[%
width=0.35\columnwidth,
height=0.20\columnwidth,
at={(0.879in,0.389in)},
scale only axis,
xmin=0,
xmax=40,
xlabel style={font=\color{white!15!black}, yshift=0.03in},
xlabel={Ang. vel. ($^{\circ}$ / frame)},
title={Rot. err. ($^{\circ}$)},
ymin=0,
ymax=25,
ymode=normal,
yminorticks=true,
axis lines = left,
axis background/.style={fill=white},
title style={font=\bfseries, yshift=-0.1in},
ylabel style={yshift=-0.15in},
ylabel={Angular error ($^{\circ}$)},
legend style={at={(1,1.25)}, anchor=south, legend cell align=left, align=left, draw=white!15!black, font=\footnotesize,nodes={scale=1.0, transform shape}},
legend columns=4
]



\addplot [color=magenta,line width=2pt, mark options={solid, red}]
  table[row sep=crcr]{%
0.0 3.52444 \\
1.0 4.31173 \\
2.0 4.0343 \\
3.0 4.0706 \\
4.0 4.60859 \\
5.0 6.15181 \\
6.0 6.24793 \\
7.0 7.21825 \\
8.0 7.49083 \\
9.0 7.90129 \\
10.0 7.79831 \\
11.0 9.47014 \\
12.0 9.29346 \\
13.0 9.86081 \\
14.0 10.5163 \\
15.0 10.8548 \\
16.0 12.0643 \\
17.0 12.8066 \\
18.0 13.2233 \\
19.0 12.7013 \\
20.0 13.8905 \\
21.0 14.2373 \\
22.0 14.777 \\
23.0 13.5278 \\
24.0 16.0616 \\
25.0 15.6041 \\
26.0 15.9871 \\
27.0 17.4976 \\
28.0 16.9479 \\
29.0 16.0874 \\
30.0 18.643 \\
31.0 19.7081 \\
32.0 19.579 \\
33.0 18.7268 \\
34.0 20.4989 \\
35.0 20.766 \\
36.0 20.6674 \\
37.0 22.2542 \\
38.0 21.4987 \\
39.0 23.1913 \\
40.0 23.3862 \\
41.0 24.5823 \\
42.0 23.3445 \\
43.0 23.3269 \\
44.0 23.8749 \\
45.0 24.4596 \\
};
\addlegendentry{5 point \cite{DBLP:journals/pami/Nister04}}

\addplot [color=cyan!40!black,line width=1.5pt, mark options={solid, orange}, mark options={solid, violet}, mark=*, mark repeat=11, mark size=1pt]
  table[row sep=crcr]{%
0.0 26.1784 \\
1.0 23.6638 \\
2.0 20.0015 \\
3.0 14.9992 \\
4.0 14.5923 \\
5.0 12.0428 \\
6.0 12.4133 \\
7.0 11.1059 \\
8.0 11.3483 \\
9.0 11.5604 \\
10.0 12.2472 \\
11.0 11.3962 \\
12.0 12.1086 \\
13.0 13.7184 \\
14.0 11.9885 \\
15.0 13.3767 \\
16.0 13.8266 \\
17.0 11.5303 \\
18.0 12.8923 \\
19.0 15.524 \\
20.0 15.378 \\
21.0 15.4622 \\
22.0 15.7917 \\
23.0 16.1075 \\
24.0 16.5753 \\
25.0 16.3589 \\
26.0 16.2832 \\
27.0 17.0077 \\
28.0 19.2515 \\
29.0 17.2873 \\
30.0 18.7898 \\
31.0 18.8772 \\
32.0 23.1548 \\
33.0 20.8004 \\
34.0 19.4697 \\
35.0 20.5404 \\
36.0 20.6927 \\
37.0 23.4036 \\
38.0 21.6073 \\
39.0 22.7479 \\
40.0 20.8896 \\
41.0 21.7608 \\
42.0 25.7171 \\
43.0 23.2978 \\
44.0 24.4622 \\
45.0 26.3875 \\
  };
\addlegendentry{K1,A1,HC}

\addplot [color=cyan,line width=1.5pt, mark options={solid, orange}, mark options={solid, violet!50!magenta}, mark=*, mark repeat=11, mark size=1pt]
  table[row sep=crcr]{%
0.0 16.9815 \\
1.0 9.15824 \\
2.0 8.66351 \\
3.0 7.34077 \\
4.0 6.87415 \\
5.0 6.54345 \\
6.0 6.26388 \\
7.0 5.91157 \\
8.0 6.31164 \\
9.0 6.90024 \\
10.0 7.01063 \\
11.0 7.28832 \\
12.0 7.38065 \\
13.0 6.847 \\
14.0 7.44754 \\
15.0 8.57033 \\
16.0 8.41865 \\
17.0 8.66181 \\
18.0 8.974 \\
19.0 9.22819 \\
20.0 9.9405 \\
21.0 9.91899 \\
22.0 11.3701 \\
23.0 10.3383 \\
24.0 10.9389 \\
25.0 11.0311 \\
26.0 12.063 \\
27.0 13.0686 \\
28.0 13.5617 \\
29.0 12.6103 \\
30.0 13.8206 \\
31.0 14.4087 \\
32.0 15.4185 \\
33.0 13.7115 \\
34.0 13.8692 \\
35.0 14.4083 \\
36.0 16.0902 \\
37.0 15.3115 \\
38.0 17.05 \\
39.0 15.5806 \\
40.0 16.3821 \\
41.0 17.5388 \\
42.0 17.0392 \\
43.0 17.5556 \\
44.0 17.237 \\
45.0 16.5097 \\
};
\addlegendentry{K1,A1,GB}

\addplot [color=cyan!40!black,line width=1.5pt, mark options={solid, orange}, mark options={solid, violet}, mark=triangle*, mark repeat=11, mark size=1pt]
  table[row sep=crcr]{%
0.0 12.6758 \\
1.0 5.76475 \\
2.0 4.39105 \\
3.0 3.93762 \\
4.0 4.01157 \\
5.0 3.45809 \\
6.0 3.07275 \\
7.0 3.6737 \\
8.0 3.6248 \\
9.0 3.9246 \\
10.0 4.06926 \\
11.0 4.53007 \\
12.0 4.60642 \\
13.0 5.5166 \\
14.0 4.93638 \\
15.0 5.82471 \\
16.0 7.06876 \\
17.0 7.39533 \\
18.0 8.60699 \\
19.0 9.66304 \\
20.0 9.08474 \\
21.0 11.109 \\
22.0 10.8276 \\
23.0 10.7607 \\
24.0 12.6719 \\
25.0 12.4765 \\
26.0 13.1378 \\
27.0 13.3815 \\
28.0 14.7759 \\
29.0 15.0103 \\
30.0 15.2237 \\
31.0 16.0609 \\
32.0 16.0933 \\
33.0 17.1265 \\
34.0 16.8617 \\
35.0 17.7612 \\
36.0 19.1176 \\
37.0 19.4067 \\
38.0 17.9611 \\
39.0 18.627 \\
40.0 19.9197 \\
41.0 20.6514 \\
42.0 20.1355 \\
43.0 20.4486 \\
44.0 21.5211 \\
45.0 20.5702 \\
  };
\addlegendentry{K1,A2,HC}

\addplot [color=cyan,line width=1.5pt, mark options={solid, orange}, mark options={solid, violet!50!magenta}, mark=triangle*, mark repeat=11, mark size=1pt]
  table[row sep=crcr]{%
0.0 13.455 \\
1.0 6.91458 \\
2.0 5.5678 \\
3.0 4.75468 \\
4.0 4.23333 \\
5.0 3.91006 \\
6.0 4.02306 \\
7.0 3.95655 \\
8.0 3.41694 \\
9.0 4.16525 \\
10.0 4.13346 \\
11.0 4.72307 \\
12.0 5.24632 \\
13.0 5.85784 \\
14.0 5.90232 \\
15.0 6.86551 \\
16.0 6.76656 \\
17.0 8.05367 \\
18.0 8.01929 \\
19.0 9.13096 \\
20.0 10.1086 \\
21.0 11.0874 \\
22.0 10.7798 \\
23.0 11.6563 \\
24.0 12.0902 \\
25.0 13.3712 \\
26.0 12.3329 \\
27.0 13.4169 \\
28.0 14.2534 \\
29.0 14.7846 \\
30.0 16.1338 \\
31.0 15.5463 \\
32.0 16.931 \\
33.0 17.6706 \\
34.0 17.5126 \\
35.0 17.4806 \\
36.0 18.2129 \\
37.0 18.3719 \\
38.0 19.2577 \\
39.0 19.3071 \\
40.0 21.2176 \\
41.0 18.7628 \\
42.0 20.3947 \\
43.0 21.093 \\
44.0 20.5898 \\
45.0 20.3174 \\
  };
\addlegendentry{K1,A2,GB}

\addplot [color=cyan!40!black,line width=1.5pt, mark options={solid, orange}, mark options={solid, black}, mark=*, mark repeat=11, mark size=1pt]
  table[row sep=crcr]{%
0.0 5.71929 \\
1.0 5.43127 \\
2.0 4.96114 \\
3.0 4.76879 \\
4.0 4.64166 \\
5.0 4.73148 \\
6.0 4.38923 \\
7.0 4.69517 \\
8.0 4.31112 \\
9.0 4.84488 \\
10.0 5.03811 \\
11.0 6.03586 \\
12.0 5.50859 \\
13.0 4.9894 \\
14.0 5.66163 \\
15.0 5.94529 \\
16.0 6.66786 \\
17.0 6.74625 \\
18.0 6.79943 \\
19.0 7.23812 \\
20.0 7.16495 \\
21.0 7.6678 \\
22.0 8.17283 \\
23.0 8.23125 \\
24.0 7.96325 \\
25.0 8.48835 \\
26.0 8.7243 \\
27.0 9.05193 \\
28.0 10.0262 \\
29.0 9.39778 \\
30.0 10.0625 \\
31.0 10.0769 \\
32.0 10.8879 \\
33.0 10.7228 \\
34.0 10.7273 \\
35.0 11.4563 \\
36.0 12.2226 \\
37.0 11.9873 \\
38.0 12.5395 \\
39.0 11.6242 \\
40.0 12.7259 \\
41.0 13.2896 \\
42.0 13.9886 \\
43.0 13.0009 \\
44.0 13.6671 \\
45.0 14.7425 \\
  };
\addlegendentry{K2,A1,HC}

\addplot [color=cyan!40!black,line width=1.5pt, mark options={solid, orange}, mark options={solid, black}, mark=triangle*, mark repeat=11, mark phase=3, mark size=1pt]
  table[row sep=crcr]{%
0.0 5.71636 \\
1.0 5.48751 \\
2.0 4.93266 \\
3.0 4.83095 \\
4.0 4.44253 \\
5.0 4.64628 \\
6.0 4.45775 \\
7.0 4.36037 \\
8.0 4.33442 \\
9.0 4.60177 \\
10.0 4.91006 \\
11.0 4.68134 \\
12.0 5.45816 \\
13.0 5.05988 \\
14.0 5.37209 \\
15.0 5.14933 \\
16.0 6.05915 \\
17.0 5.99307 \\
18.0 6.08659 \\
19.0 5.8077 \\
20.0 6.66285 \\
21.0 6.44115 \\
22.0 7.40917 \\
23.0 6.52019 \\
24.0 6.80749 \\
25.0 7.18016 \\
26.0 7.34402 \\
27.0 7.04437 \\
28.0 7.26944 \\
29.0 7.66437 \\
30.0 7.68748 \\
31.0 8.15808 \\
32.0 8.4206 \\
33.0 8.52932 \\
34.0 9.0733 \\
35.0 8.80481 \\
36.0 8.88293 \\
37.0 8.97345 \\
38.0 8.75375 \\
39.0 9.86279 \\
40.0 9.51048 \\
41.0 9.62497 \\
42.0 9.47384 \\
43.0 10.1947 \\
44.0 10.3967 \\
45.0 10.2816 \\
  };
\addlegendentry{K2,A2,HC}

\addplot [color=cyan,line width=1.5pt, mark options={solid, orange}, mark options={solid, yellow}, mark=triangle*, mark repeat=11, mark phase=7, mark size=1pt]
  table[row sep=crcr]{%
0.0 5.55547 \\
1.0 5.07815 \\
2.0 4.85731 \\
3.0 4.63796 \\
4.0 4.41919 \\
5.0 4.57249 \\
6.0 4.24033 \\
7.0 4.4067 \\
8.0 4.43867 \\
9.0 4.73179 \\
10.0 4.88915 \\
11.0 4.49681 \\
12.0 4.84934 \\
13.0 5.37761 \\
14.0 5.12664 \\
15.0 4.97697 \\
16.0 5.6462 \\
17.0 6.03732 \\
18.0 5.92603 \\
19.0 5.85291 \\
20.0 5.31454 \\
21.0 6.22081 \\
22.0 6.81647 \\
23.0 6.45571 \\
24.0 6.97051 \\
25.0 6.59487 \\
26.0 7.28361 \\
27.0 7.05781 \\
28.0 7.29616 \\
29.0 7.59271 \\
30.0 7.2393 \\
31.0 7.49995 \\
32.0 8.13811 \\
33.0 7.96066 \\
34.0 8.40543 \\
35.0 8.26216 \\
36.0 8.86379 \\
37.0 8.61098 \\
38.0 8.90025 \\
39.0 8.88142 \\
40.0 9.33679 \\
41.0 9.34308 \\
42.0 9.28591 \\
43.0 10.3524 \\
44.0 9.95521 \\
45.0 10.7613 \\
  };
\addlegendentry{K2,A2,GB}

\end{axis}

\input{plot/noise_test_t_4v}

\end{tikzpicture}%
    \caption{
    \textbf{Noise test, rolling shutter cameras. } Rotation (\textit{left}) and translation (\textit{right}) errors of the solvers with \textit{top:} $m=2,n=5$, \textit{middle:} $m=3,n=2$, and \textit{bottom:} $m=4,n=1$, as the function of the angular velocity $\omega$, Averaged over $1000$ synthetic rolling shutter samples with additional noise of magnitude $1px$.  The result of the five point solver is shown in all graphs.
    }
    \label{fig:noise_tests}
\end{figure}

\begin{figure}
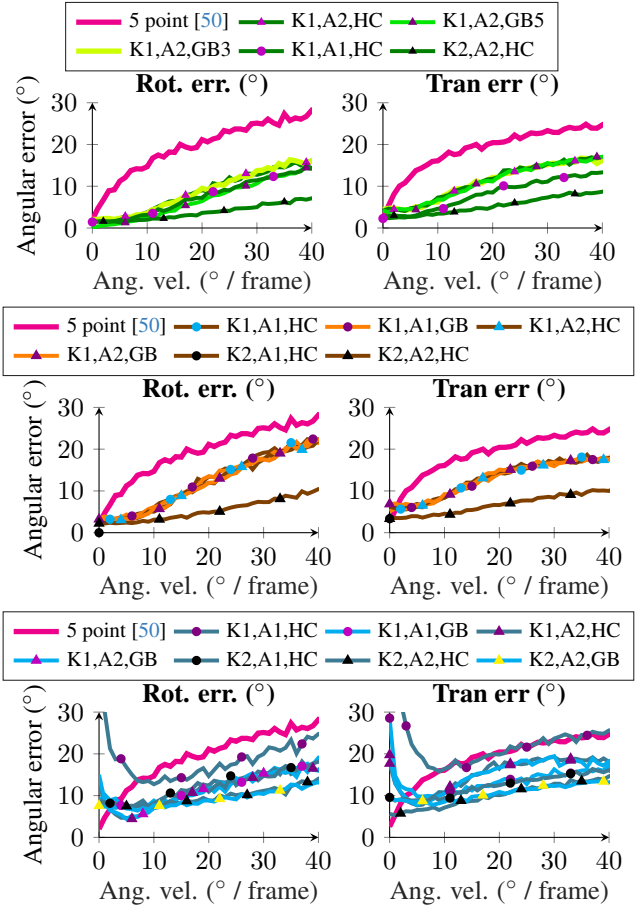

    \centering
    \begin{tikzpicture}

\begin{axis}[%
width=0.35\columnwidth,
height=0.2\columnwidth,
at={(0.879in,0.389in)},
scale only axis,
xmin=0,
xmax=40,
xlabel style={font=\color{white!15!black}, yshift=0.03in},
xlabel={Ang. vel. ($^{\circ}$ / frame)},
title={Rot. err. ($^{\circ}$)},
ymin=0,
ymax=30,
ymode=normal,
yminorticks=true,
axis lines = left,
axis background/.style={fill=white},
title style={font=\bfseries, yshift=-0.1in},
ylabel style={yshift=-0.15in},
ylabel={Angular error ($^{\circ}$)},
legend style={at={(1,1.25)}, anchor=south, legend cell align=left, align=left, draw=white!15!black, font=\footnotesize,nodes={scale=1.0, transform shape}},
legend columns=3
]


\addplot [color=magenta,line width=2pt, mark options={solid, red}]
  table[row sep=crcr]{%
0.0 1.99402 \\
1.0 4.68332 \\
2.0 6.71471 \\
3.0 8.9243 \\
4.0 9.14814 \\
5.0 11.2215 \\
6.0 12.0914 \\
7.0 13.755 \\
8.0 14.1749 \\
9.0 14.1179 \\
10.0 14.6752 \\
11.0 16.7565 \\
12.0 17.4719 \\
13.0 17.0091 \\
14.0 18.2004 \\
15.0 18.1031 \\
16.0 20.0305 \\
17.0 19.4151 \\
18.0 19.0276 \\
19.0 19.2755 \\
20.0 21.1534 \\
21.0 20.6281 \\
22.0 21.3156 \\
23.0 21.6687 \\
24.0 23.333 \\
25.0 23.0191 \\
26.0 23.7146 \\
27.0 23.7087 \\
28.0 23.4044 \\
29.0 24.9635 \\
30.0 25.1161 \\
31.0 24.9048 \\
32.0 26.0714 \\
33.0 25.1783 \\
34.0 24.6176 \\
35.0 27.3313 \\
36.0 25.7256 \\
37.0 26.3565 \\
38.0 26.0542 \\
39.0 26.713 \\
40.0 28.2 \\
41.0 27.6481 \\
42.0 28.0926 \\
43.0 27.6151 \\
44.0 29.1614 \\
45.0 28.9057 \\
};
\addlegendentry{5 point \cite{DBLP:journals/pami/Nister04}}


\addplot [color=green!50!black,line width=1.5pt, mark options={solid, violet!50!magenta}, mark=triangle*, mark repeat=11, mark phase=7, mark size=0.5pt]
  table[row sep=crcr]{%
0.0 1.83981 \\
1.0 2.1954 \\
2.0 2.12964 \\
3.0 1.91915 \\
4.0 2.0131 \\
5.0 2.34145 \\
6.0 2.81887 \\
7.0 2.62662 \\
8.0 2.98546 \\
9.0 3.32578 \\
10.0 3.78208 \\
11.0 4.5307 \\
12.0 5.14252 \\
13.0 5.75637 \\
14.0 6.3536 \\
15.0 6.45553 \\
16.0 6.57953 \\
17.0 7.86195 \\
18.0 8.12873 \\
19.0 8.67295 \\
20.0 9.04159 \\
21.0 8.73305 \\
22.0 10.7293 \\
23.0 10.4634 \\
24.0 10.8276 \\
25.0 12.1565 \\
26.0 11.3015 \\
27.0 11.6749 \\
28.0 13.0629 \\
29.0 12.9892 \\
30.0 12.7242 \\
31.0 13.0838 \\
32.0 13.5345 \\
33.0 14.5414 \\
34.0 14.5499 \\
35.0 14.5036 \\
36.0 15.0154 \\
37.0 15.7529 \\
38.0 14.9821 \\
39.0 15.7246 \\
40.0 16.1979 \\
41.0 16.1433 \\
42.0 17.1439 \\
43.0 17.015 \\
44.0 17.308 \\
45.0 18.0733 \\
};
\addlegendentry{K1,A2,HC}

\addplot [color=green!90!black,line width=1.5pt, mark options={solid, violet}, mark=triangle*, mark repeat=11, mark phase=7, mark size=0.5pt]
  table[row sep=crcr]{%
0.0 0.0 \\
1.0 0.389403 \\
2.0 0.612092 \\
3.0 0.786225 \\
4.0 0.897454 \\
5.0 1.10731 \\
6.0 1.27341 \\
7.0 1.61809 \\
8.0 1.80362 \\
9.0 2.13327 \\
10.0 2.49842 \\
11.0 2.86779 \\
12.0 3.13059 \\
13.0 3.7194 \\
14.0 4.03298 \\
15.0 4.15165 \\
16.0 5.15639 \\
17.0 5.38173 \\
18.0 5.7649 \\
19.0 6.04115 \\
20.0 6.40865 \\
21.0 7.19724 \\
22.0 7.81333 \\
23.0 7.90347 \\
24.0 8.34302 \\
25.0 8.95332 \\
26.0 9.67022 \\
27.0 9.73184 \\
28.0 10.0571 \\
29.0 10.0987 \\
30.0 10.6514 \\
31.0 11.555 \\
32.0 12.3096 \\
33.0 12.0906 \\
34.0 12.4062 \\
35.0 12.5524 \\
36.0 13.0234 \\
37.0 13.0981 \\
38.0 13.8029 \\
39.0 14.4172 \\
40.0 14.1317 \\
41.0 14.8239 \\
42.0 15.7045 \\
43.0 15.444 \\
44.0 16.1365 \\
45.0 17.1366 \\
   }; 
\addlegendentry{K1,A2,GB5}

\addplot [color=green!20!yellow,line width=2pt, mark options={solid, red}]
  table[row sep=crcr]{%
0.0 2.06542 \\
1.0 2.04508 \\
2.0 2.26059 \\
3.0 2.29282 \\
4.0 2.06321 \\
5.0 2.3576 \\
6.0 2.28813 \\
7.0 2.55263 \\
8.0 3.14786 \\
9.0 3.24205 \\
10.0 3.49791 \\
11.0 4.44715 \\
12.0 4.67589 \\
13.0 5.65506 \\
14.0 5.87056 \\
15.0 6.23202 \\
16.0 6.87204 \\
17.0 7.82462 \\
18.0 7.49287 \\
19.0 8.49674 \\
20.0 9.52526 \\
21.0 9.15141 \\
22.0 9.89299 \\
23.0 10.4168 \\
24.0 10.2367 \\
25.0 11.3502 \\
26.0 11.6269 \\
27.0 12.3684 \\
28.0 12.2494 \\
29.0 12.595 \\
30.0 13.7608 \\
31.0 13.8302 \\
32.0 14.4662 \\
33.0 14.0242 \\
34.0 15.4393 \\
35.0 14.8506 \\
36.0 16.3393 \\
37.0 16.1186 \\
38.0 15.7063 \\
39.0 15.3616 \\
40.0 16.1429 \\
41.0 15.5823 \\
42.0 17.9971 \\
43.0 16.1512 \\
44.0 18.2745 \\
45.0 17.0322 \\
};
\addlegendentry{K1,A2,GB3}


\addplot [color=green!50!black,line width=1.5pt, mark options={solid, violet!50!magenta}, mark=*, mark repeat=11, mark size=1pt]
  table[row sep=crcr]{%
0.0 1.44851 \\
1.0 1.54166 \\
2.0 1.63454 \\
3.0 1.57137 \\
4.0 1.63273 \\
5.0 1.95668 \\
6.0 2.23374 \\
7.0 2.21849 \\
8.0 2.67106 \\
9.0 3.05592 \\
10.0 3.14578 \\
11.0 3.51328 \\
12.0 4.49651 \\
13.0 4.08088 \\
14.0 4.54707 \\
15.0 5.20892 \\
16.0 5.92349 \\
17.0 6.81888 \\
18.0 6.82104 \\
19.0 7.20298 \\
20.0 7.22489 \\
21.0 8.32737 \\
22.0 8.69479 \\
23.0 8.55644 \\
24.0 9.13521 \\
25.0 9.01534 \\
26.0 9.76413 \\
27.0 10.1786 \\
28.0 11.0202 \\
29.0 11.3168 \\
30.0 11.432 \\
31.0 11.7329 \\
32.0 12.0014 \\
33.0 12.3315 \\
34.0 12.6447 \\
35.0 12.8582 \\
36.0 13.2993 \\
37.0 13.2541 \\
38.0 13.5687 \\
39.0 14.9023 \\
40.0 14.3386 \\
41.0 15.0322 \\
42.0 14.8109 \\
43.0 16.2979 \\
44.0 16.0653 \\
45.0 15.9678 \\
   }; 
\addlegendentry{K1,A1,HC}

\addplot [color=green!50!black,line width=1.5pt, mark options={solid,violet, draw=black,fill=violet}, mark=triangle*, mark repeat=11, mark size=0.4pt, mark phase=3]
  table[row sep=crcr]{%
0.0 1.64468 \\
1.0 1.5461 \\
2.0 1.60289 \\
3.0 1.4766 \\
4.0 1.60302 \\
5.0 1.32898 \\
6.0 1.43156 \\
7.0 1.88173 \\
8.0 1.87625 \\
9.0 1.92071 \\
10.0 1.97553 \\
11.0 2.2597 \\
12.0 2.23871 \\
13.0 2.29278 \\
14.0 2.58842 \\
15.0 2.51652 \\
16.0 2.86027 \\
17.0 3.0436 \\
18.0 3.12693 \\
19.0 3.00586 \\
20.0 3.62312 \\
21.0 3.73677 \\
22.0 3.8093 \\
23.0 3.91656 \\
24.0 4.16768 \\
25.0 4.10055 \\
26.0 4.66324 \\
27.0 4.70554 \\
28.0 4.75189 \\
29.0 4.83874 \\
30.0 4.99675 \\
31.0 5.90918 \\
32.0 5.75032 \\
33.0 5.83803 \\
34.0 5.95558 \\
35.0 6.26334 \\
36.0 5.91765 \\
37.0 6.88876 \\
38.0 6.97012 \\
39.0 6.79276 \\
40.0 7.11059 \\
41.0 7.02653 \\
42.0 8.11872 \\
43.0 7.3551 \\
44.0 8.14729 \\
45.0 7.71118 \\
   }; 
\addlegendentry{K2,A2,HC}

\end{axis}

\input{plot/noise_test_event_t}

\end{tikzpicture}%
    \begin{tikzpicture}

\begin{axis}[%
width=0.35\columnwidth,
height=0.2\columnwidth,
at={(0.879in,0.389in)},
scale only axis,
xmin=0,
xmax=40,
xlabel style={font=\color{white!15!black}, yshift=0.03in},
xlabel={Ang. vel. ($^{\circ}$ / frame)},
title={Rot. err. ($^{\circ}$)},
ymin=0,
ymax=30,
ymode=normal,
yminorticks=true,
axis lines = left,
axis background/.style={fill=white},
title style={font=\bfseries, yshift=-0.1in},
ylabel style={yshift=-0.15in},
ylabel={Angular error ($^{\circ}$)},
legend style={at={(1,1.25)}, anchor=south, legend cell align=left, align=left, draw=white!15!black, font=\footnotesize,nodes={scale=1.0, transform shape}},
legend columns=4
]


\addplot [color=magenta,line width=2pt, mark options={solid, red}]
  table[row sep=crcr]{%
0.0 1.99402 \\
1.0 4.68332 \\
2.0 6.71471 \\
3.0 8.9243 \\
4.0 9.14814 \\
5.0 11.2215 \\
6.0 12.0914 \\
7.0 13.755 \\
8.0 14.1749 \\
9.0 14.1179 \\
10.0 14.6752 \\
11.0 16.7565 \\
12.0 17.4719 \\
13.0 17.0091 \\
14.0 18.2004 \\
15.0 18.1031 \\
16.0 20.0305 \\
17.0 19.4151 \\
18.0 19.0276 \\
19.0 19.2755 \\
20.0 21.1534 \\
21.0 20.6281 \\
22.0 21.3156 \\
23.0 21.6687 \\
24.0 23.333 \\
25.0 23.0191 \\
26.0 23.7146 \\
27.0 23.7087 \\
28.0 23.4044 \\
29.0 24.9635 \\
30.0 25.1161 \\
31.0 24.9048 \\
32.0 26.0714 \\
33.0 25.1783 \\
34.0 24.6176 \\
35.0 27.3313 \\
36.0 25.7256 \\
37.0 26.3565 \\
38.0 26.0542 \\
39.0 26.713 \\
40.0 28.2 \\
41.0 27.6481 \\
42.0 28.0926 \\
43.0 27.6151 \\
44.0 29.1614 \\
45.0 28.9057 \\
};
\addlegendentry{5 point \cite{DBLP:journals/pami/Nister04}}



\addplot [color=orange!50!black,line width=1.5pt, mark options={solid, orange}, mark options={solid, cyan}, mark=*, mark repeat=11, mark phase=3, mark size=1pt]
  table[row sep=crcr]{%
0.0 3.51084 \\
1.0 3.3343 \\
2.0 3.22727 \\
3.0 3.32321 \\
4.0 3.7161 \\
5.0 3.75062 \\
6.0 3.86794 \\
7.0 4.51904 \\
8.0 4.73359 \\
9.0 5.36478 \\
10.0 6.06122 \\
11.0 7.07144 \\
12.0 7.52431 \\
13.0 7.97013 \\
14.0 8.82902 \\
15.0 8.97356 \\
16.0 9.76326 \\
17.0 10.5685 \\
18.0 12.908 \\
19.0 12.2956 \\
20.0 13.4356 \\
21.0 13.3312 \\
22.0 14.6455 \\
23.0 15.3581 \\
24.0 15.1415 \\
25.0 15.0744 \\
26.0 16.0977 \\
27.0 17.0381 \\
28.0 18.0816 \\
29.0 18.6963 \\
30.0 19.1501 \\
31.0 19.3236 \\
32.0 20.2271 \\
33.0 19.1319 \\
34.0 20.0047 \\
35.0 21.5571 \\
36.0 20.9807 \\
37.0 21.0065 \\
38.0 22.386 \\
39.0 22.0982 \\
40.0 22.3503 \\
41.0 22.7333 \\
42.0 21.8835 \\
43.0 21.3555 \\
44.0 22.8935 \\
45.0 24.8493 \\
  };
\addlegendentry{K1,A1,HC}

\addplot [color=orange,line width=1.5pt, mark options={solid, orange}, mark options={solid, violet}, mark=*, mark repeat=11, mark phase = 7, mark size=1pt]
  table[row sep=crcr]{%
0.0 3.41015 \\
1.0 3.87366 \\
2.0 3.50754 \\
3.0 3.33398 \\
4.0 3.61515 \\
5.0 4.34229 \\
6.0 4.00091 \\
7.0 4.18313 \\
8.0 4.76918 \\
9.0 5.6698 \\
10.0 6.13188 \\
11.0 6.4205 \\
12.0 7.33784 \\
13.0 7.67956 \\
14.0 8.69359 \\
15.0 9.40988 \\
16.0 9.91743 \\
17.0 10.9273 \\
18.0 11.2216 \\
19.0 12.8948 \\
20.0 13.4887 \\
21.0 13.2394 \\
22.0 14.879 \\
23.0 14.8273 \\
24.0 15.3793 \\
25.0 15.6598 \\
26.0 16.116 \\
27.0 16.0995 \\
28.0 17.8586 \\
29.0 17.6924 \\
30.0 17.7676 \\
31.0 18.3538 \\
32.0 19.3681 \\
33.0 18.9332 \\
34.0 20.2212 \\
35.0 20.0156 \\
36.0 21.7257 \\
37.0 20.3578 \\
38.0 21.9841 \\
39.0 22.4275 \\
40.0 21.4626 \\
41.0 21.8523 \\
42.0 23.0904 \\
43.0 22.0931 \\
44.0 24.4392 \\
45.0 24.868 \\
  };
\addlegendentry{K1,A1,GB}

\addplot [color=orange!50!black,line width=1.5pt, mark options={solid, orange}, mark options={solid, cyan}, mark=triangle*, mark repeat=11, mark phase = 5, mark size=1pt]
  table[row sep=crcr]{%
0.0 2.40574 \\
1.0 3.03503 \\
2.0 3.04624 \\
3.0 2.83539 \\
4.0 3.06369 \\
5.0 3.54634 \\
6.0 3.3446 \\
7.0 3.54869 \\
8.0 4.05682 \\
9.0 5.11076 \\
10.0 5.02542 \\
11.0 5.71647 \\
12.0 6.2285 \\
13.0 6.80273 \\
14.0 8.47611 \\
15.0 8.8292 \\
16.0 9.88117 \\
17.0 9.67104 \\
18.0 10.6374 \\
19.0 11.3838 \\
20.0 11.9366 \\
21.0 11.8132 \\
22.0 13.2456 \\
23.0 14.0531 \\
24.0 14.7298 \\
25.0 14.7219 \\
26.0 15.7743 \\
27.0 16.6969 \\
28.0 15.1494 \\
29.0 16.7824 \\
30.0 17.5458 \\
31.0 18.6741 \\
32.0 17.9647 \\
33.0 19.661 \\
34.0 19.4848 \\
35.0 19.8543 \\
36.0 20.5171 \\
37.0 19.8845 \\
38.0 20.5289 \\
39.0 20.8216 \\
40.0 22.1533 \\
41.0 22.2494 \\
42.0 22.8695 \\
43.0 23.5302 \\
44.0 23.853 \\
45.0 23.9318 \\
  };
\addlegendentry{K1,A2,HC}

\addplot [color=orange,line width=1.5pt, mark options={solid, orange}, mark options={solid, violet}, mark=triangle*, mark repeat=11, mark size=1pt]
  table[row sep=crcr]{%
0.0 3.22047 \\
1.0 3.30557 \\
2.0 3.20883 \\
3.0 3.11762 \\
4.0 3.16866 \\
5.0 3.38355 \\
6.0 3.21189 \\
7.0 3.78143 \\
8.0 4.05991 \\
9.0 4.56989 \\
10.0 5.75794 \\
11.0 5.7142 \\
12.0 5.91298 \\
13.0 6.93075 \\
14.0 8.38217 \\
15.0 8.5475 \\
16.0 9.2955 \\
17.0 9.93497 \\
18.0 10.4508 \\
19.0 11.058 \\
20.0 11.9429 \\
21.0 12.0393 \\
22.0 12.9867 \\
23.0 13.0872 \\
24.0 13.9347 \\
25.0 14.2591 \\
26.0 15.4772 \\
27.0 16.2139 \\
28.0 16.576 \\
29.0 17.4173 \\
30.0 17.5441 \\
31.0 17.2542 \\
32.0 17.8496 \\
33.0 19.0127 \\
34.0 19.6002 \\
35.0 21.3191 \\
36.0 20.0911 \\
37.0 20.6293 \\
38.0 20.6841 \\
39.0 21.9208 \\
40.0 22.1455 \\
41.0 22.9325 \\
42.0 22.8404 \\
43.0 22.62 \\
44.0 23.4231 \\
45.0 23.1294 \\
  };
\addlegendentry{K1,A2,GB}

\addplot [color=orange!50!black,line width=1.5pt, mark options={solid, orange}, mark options={solid, black}, mark=*, mark repeat=11, mark size=1pt]
  table[row sep=crcr]{%
0.0 1.48225e-07 \\
  };
\addlegendentry{K2,A1,HC}

\addplot [color=orange!50!black,line width=1.5pt, mark options={solid, orange}, mark options={solid, black}, mark=triangle*, mark repeat=11, mark size=1pt]
  table[row sep=crcr]{%
0.0 2.22115 \\
1.0 2.20233 \\
2.0 2.3791 \\
3.0 2.25959 \\
4.0 2.33535 \\
5.0 2.41725 \\
6.0 2.29963 \\
7.0 2.4849 \\
8.0 2.83092 \\
9.0 2.62938 \\
10.0 3.11256 \\
11.0 3.17935 \\
12.0 3.19426 \\
13.0 3.55605 \\
14.0 3.75249 \\
15.0 4.00449 \\
16.0 3.57371 \\
17.0 4.27048 \\
18.0 4.74098 \\
19.0 4.88944 \\
20.0 4.79423 \\
21.0 5.04308 \\
22.0 5.06192 \\
23.0 5.60492 \\
24.0 5.9979 \\
25.0 6.29677 \\
26.0 6.4695 \\
27.0 6.57735 \\
28.0 7.05156 \\
29.0 7.39206 \\
30.0 7.31304 \\
31.0 7.33916 \\
32.0 8.15453 \\
33.0 8.09669 \\
34.0 8.85774 \\
35.0 8.61578 \\
36.0 8.36538 \\
37.0 9.91152 \\
38.0 9.30353 \\
39.0 9.83588 \\
40.0 10.4429 \\
41.0 10.4869 \\
42.0 10.2873 \\
43.0 11.5339 \\
44.0 11.2027 \\
45.0 11.2932 \\
  };
\addlegendentry{K2,A2,HC}

\end{axis}

\input{plot/noise_test_event_t_3v}

\end{tikzpicture}%
    \begin{tikzpicture}

\begin{axis}[%
width=0.35\columnwidth,
height=0.2\columnwidth,
at={(0.879in,0.389in)},
scale only axis,
xmin=0,
xmax=40,
xlabel style={font=\color{white!15!black}, yshift=0.03in},
xlabel={Ang. vel. ($^{\circ}$ / frame)},
title={Rot. err. ($^{\circ}$)},
ymin=0,
ymax=30,
ymode=normal,
yminorticks=true,
axis lines = left,
axis background/.style={fill=white},
title style={font=\bfseries, yshift=-0.1in},
ylabel style={yshift=-0.15in},
ylabel={Angular error ($^{\circ}$)},
legend style={at={(1,1.25)}, anchor=south, legend cell align=left, align=left, draw=white!15!black, font=\footnotesize,nodes={scale=1.0, transform shape}},
legend columns=4
]


\addplot [color=magenta,line width=2pt, mark options={solid, red}]
  table[row sep=crcr]{%
0.0 1.99402 \\
1.0 4.68332 \\
2.0 6.71471 \\
3.0 8.9243 \\
4.0 9.14814 \\
5.0 11.2215 \\
6.0 12.0914 \\
7.0 13.755 \\
8.0 14.1749 \\
9.0 14.1179 \\
10.0 14.6752 \\
11.0 16.7565 \\
12.0 17.4719 \\
13.0 17.0091 \\
14.0 18.2004 \\
15.0 18.1031 \\
16.0 20.0305 \\
17.0 19.4151 \\
18.0 19.0276 \\
19.0 19.2755 \\
20.0 21.1534 \\
21.0 20.6281 \\
22.0 21.3156 \\
23.0 21.6687 \\
24.0 23.333 \\
25.0 23.0191 \\
26.0 23.7146 \\
27.0 23.7087 \\
28.0 23.4044 \\
29.0 24.9635 \\
30.0 25.1161 \\
31.0 24.9048 \\
32.0 26.0714 \\
33.0 25.1783 \\
34.0 24.6176 \\
35.0 27.3313 \\
36.0 25.7256 \\
37.0 26.3565 \\
38.0 26.0542 \\
39.0 26.713 \\
40.0 28.2 \\
41.0 27.6481 \\
42.0 28.0926 \\
43.0 27.6151 \\
44.0 29.1614 \\
45.0 28.9057 \\
};
\addlegendentry{5 point \cite{DBLP:journals/pami/Nister04}}



\addplot [color=cyan!50!black,line width=1.5pt, mark options={solid, orange}, mark options={solid, violet}, mark=*, mark repeat=11, mark phase=3, mark size=1pt]
  table[row sep=crcr]{%
0.0 31.145 \\
1.0 30.4203 \\
2.0 21.2639 \\
3.0 18.6878 \\
4.0 18.8218 \\
5.0 16.7923 \\
6.0 15.0176 \\
7.0 13.57 \\
8.0 13.2232 \\
9.0 13.5067 \\
10.0 12.7231 \\
11.0 13.3254 \\
12.0 14.5979 \\
13.0 13.4743 \\
14.0 14.0585 \\
15.0 14.2897 \\
16.0 13.8018 \\
17.0 14.6222 \\
18.0 15.8548 \\
19.0 16.5642 \\
20.0 14.681 \\
21.0 16.1616 \\
22.0 17.6613 \\
23.0 16.3659 \\
24.0 17.3835 \\
25.0 17.6569 \\
26.0 19.2626 \\
27.0 19.4556 \\
28.0 17.5751 \\
29.0 19.3036 \\
30.0 19.9651 \\
31.0 21.5553 \\
32.0 21.2764 \\
33.0 20.7377 \\
34.0 21.4463 \\
35.0 23.3396 \\
36.0 20.0672 \\
37.0 22.3759 \\
38.0 24.1098 \\
39.0 23.6619 \\
40.0 24.7655 \\
41.0 25.0142 \\
42.0 25.1369 \\
43.0 23.5715 \\
44.0 24.8466 \\
45.0 26.6884 \\
  };
\addlegendentry{K1,A1,HC}

\addplot [color=cyan,line width=1.5pt, mark options={solid, orange}, mark options={solid, violet!50!magenta}, mark=*, mark repeat=11, mark phase=5, mark size=1pt]
  table[row sep=crcr]{%
0.0 8.41356 \\
1.0 10.2737 \\
2.0 8.86871 \\
3.0 9.29325 \\
4.0 7.6643 \\
5.0 7.98296 \\
6.0 6.76412 \\
7.0 7.48579 \\
8.0 7.93465 \\
9.0 7.70578 \\
10.0 8.80179 \\
11.0 7.95328 \\
12.0 8.90325 \\
13.0 8.05564 \\
14.0 8.90708 \\
15.0 9.99155 \\
16.0 10.0281 \\
17.0 10.0767 \\
18.0 10.4657 \\
19.0 10.8039 \\
20.0 12.0959 \\
21.0 10.9295 \\
22.0 12.1483 \\
23.0 10.9105 \\
24.0 12.7248 \\
25.0 13.4992 \\
26.0 13.2044 \\
27.0 14.491 \\
28.0 13.3898 \\
29.0 14.3943 \\
30.0 15.0768 \\
31.0 15.5208 \\
32.0 16.6369 \\
33.0 16.1977 \\
34.0 16.9445 \\
35.0 16.8578 \\
36.0 16.6031 \\
37.0 17.0466 \\
38.0 15.651 \\
39.0 17.4872 \\
40.0 19.0808 \\
41.0 18.5575 \\
42.0 18.4172 \\
43.0 18.0108 \\
44.0 19.0232 \\
45.0 17.0956 \\
};
\addlegendentry{K1,A1,GB}

\addplot [color=cyan!50!black,line width=1.5pt, mark options={solid, orange}, mark options={solid, violet}, mark=triangle*, mark repeat=11, mark phase=7, mark size=1pt]
  table[row sep=crcr]{%
0.0 12.8445 \\
1.0 8.47992 \\
2.0 6.74203 \\
3.0 5.52698 \\
4.0 4.69447 \\
5.0 4.67367 \\
6.0 4.43836 \\
7.0 5.07855 \\
8.0 5.58739 \\
9.0 6.30029 \\
10.0 7.01529 \\
11.0 7.27258 \\
12.0 8.25868 \\
13.0 8.20023 \\
14.0 9.83262 \\
15.0 10.1421 \\
16.0 10.6437 \\
17.0 10.7125 \\
18.0 11.3172 \\
19.0 12.6511 \\
20.0 12.3995 \\
21.0 13.561 \\
22.0 13.2717 \\
23.0 13.1447 \\
24.0 13.4574 \\
25.0 13.8927 \\
26.0 14.1492 \\
27.0 14.5725 \\
28.0 14.292 \\
29.0 15.7606 \\
30.0 15.4484 \\
31.0 15.3596 \\
32.0 16.3461 \\
33.0 15.0997 \\
34.0 16.8703 \\
35.0 15.1699 \\
36.0 15.7986 \\
37.0 18.1798 \\
38.0 16.6713 \\
39.0 16.5099 \\
40.0 16.0691 \\
41.0 17.5478 \\
42.0 17.9011 \\
43.0 16.3307 \\
44.0 17.2499 \\
45.0 17.5627 \\
  };
\addlegendentry{K1,A2,HC}

\addplot [color=cyan,line width=1.5pt, mark options={solid, orange}, mark options={solid, violet!50!magenta}, mark=triangle*, mark repeat=11, mark phase=9, mark size=1pt]
  table[row sep=crcr]{%
0.0 14.9935 \\
1.0 8.92945 \\
2.0 6.44066 \\
3.0 5.95754 \\
4.0 5.37694 \\
5.0 4.38407 \\
6.0 5.16598 \\
7.0 4.87023 \\
8.0 5.61045 \\
9.0 5.9314 \\
10.0 7.0383 \\
11.0 8.32873 \\
12.0 7.7412 \\
13.0 8.76056 \\
14.0 9.16078 \\
15.0 9.98828 \\
16.0 9.64505 \\
17.0 10.8631 \\
18.0 10.9222 \\
19.0 11.6313 \\
20.0 11.7316 \\
21.0 11.9821 \\
22.0 13.2395 \\
23.0 12.7816 \\
24.0 13.9775 \\
25.0 14.1901 \\
26.0 13.3906 \\
27.0 14.7258 \\
28.0 15.3528 \\
29.0 14.3497 \\
30.0 15.3497 \\
31.0 14.9932 \\
32.0 15.7163 \\
33.0 15.7844 \\
34.0 15.4279 \\
35.0 15.8125 \\
36.0 14.4775 \\
37.0 17.0922 \\
38.0 17.6517 \\
39.0 16.1673 \\
40.0 15.9965 \\
41.0 16.705 \\
42.0 16.1127 \\
43.0 16.6728 \\
44.0 16.573 \\
45.0 18.5849 \\
  };
\addlegendentry{K1,A2,GB}

\addplot [color=cyan!50!black,line width=1.5pt, mark options={solid, orange}, mark options={solid, black}, mark=*, mark repeat=11, mark phase=3, mark size=1pt]
  table[row sep=crcr]{%
0.0 8.79346 \\
1.0 8.44419 \\
2.0 8.16693 \\
3.0 7.57836 \\
4.0 7.05718 \\
5.0 7.69557 \\
6.0 7.25799 \\
7.0 8.02128 \\
8.0 8.80803 \\
9.0 7.34263 \\
10.0 8.17639 \\
11.0 8.90198 \\
12.0 9.62128 \\
13.0 10.6203 \\
14.0 9.91417 \\
15.0 11.3255 \\
16.0 11.1333 \\
17.0 11.6497 \\
18.0 13.0149 \\
19.0 12.3752 \\
20.0 10.9438 \\
21.0 12.0789 \\
22.0 13.0339 \\
23.0 11.9137 \\
24.0 14.7263 \\
25.0 12.4803 \\
26.0 13.4114 \\
27.0 13.9241 \\
28.0 14.7723 \\
29.0 13.6707 \\
30.0 14.7273 \\
31.0 14.5603 \\
32.0 14.6489 \\
33.0 16.1176 \\
34.0 17.0329 \\
35.0 16.7116 \\
36.0 16.0974 \\
37.0 17.1285 \\
38.0 17.0698 \\
39.0 17.7196 \\
40.0 18.1156 \\
41.0 18.0267 \\
42.0 17.5391 \\
43.0 16.837 \\
44.0 18.9886 \\
45.0 18.2146 \\
  };
\addlegendentry{K2,A1,HC}

\addplot [color=cyan!50!black,line width=1.5pt, mark options={solid, orange}, mark options={solid, black}, mark=triangle*, mark repeat=11, mark phase=6, mark size=1pt]
  table[row sep=crcr]{%
0.0 8.22746 \\
1.0 7.63936 \\
2.0 7.55792 \\
3.0 8.06909 \\
4.0 7.29999 \\
5.0 7.493 \\
6.0 7.57892 \\
7.0 7.44305 \\
8.0 7.57855 \\
9.0 7.07099 \\
10.0 7.36904 \\
11.0 7.27278 \\
12.0 7.98632 \\
13.0 8.65347 \\
14.0 8.20905 \\
15.0 8.27461 \\
16.0 8.7081 \\
17.0 8.52848 \\
18.0 8.77164 \\
19.0 9.44156 \\
20.0 8.8623 \\
21.0 8.80379 \\
22.0 10.038 \\
23.0 9.62715 \\
24.0 10.4793 \\
25.0 10.0814 \\
26.0 10.3687 \\
27.0 10.1651 \\
28.0 10.5459 \\
29.0 10.896 \\
30.0 11.1437 \\
31.0 11.3775 \\
32.0 11.84 \\
33.0 11.2999 \\
34.0 10.2819 \\
35.0 12.3021 \\
36.0 12.6624 \\
37.0 13.4427 \\
38.0 13.2426 \\
39.0 13.1392 \\
40.0 14.0571 \\
41.0 14.4015 \\
42.0 14.234 \\
43.0 13.0199 \\
44.0 15.2773 \\
45.0 15.148 \\
  };
\addlegendentry{K2,A2,HC}

\addplot [color=cyan,line width=1.5pt, mark options={solid, orange}, mark options={solid, yellow}, mark=triangle*, mark repeat=11, mark size=1pt]
  table[row sep=crcr]{%
0.0 7.5737 \\
1.0 7.34687 \\
2.0 7.32504 \\
3.0 7.05726 \\
4.0 7.42042 \\
5.0 6.89933 \\
6.0 6.74312 \\
7.0 6.39827 \\
8.0 7.0294 \\
9.0 6.679 \\
10.0 7.68711 \\
11.0 7.58842 \\
12.0 7.17998 \\
13.0 7.66114 \\
14.0 7.69684 \\
15.0 8.38652 \\
16.0 7.6985 \\
17.0 8.12497 \\
18.0 8.03292 \\
19.0 8.80076 \\
20.0 8.82836 \\
21.0 8.5645 \\
22.0 9.22571 \\
23.0 9.58532 \\
24.0 9.3201 \\
25.0 9.09345 \\
26.0 10.3376 \\
27.0 9.53967 \\
28.0 10.2371 \\
29.0 10.5502 \\
30.0 10.3881 \\
31.0 10.9805 \\
32.0 11.0014 \\
33.0 11.2448 \\
34.0 11.6149 \\
35.0 12.658 \\
36.0 12.5192 \\
37.0 11.6187 \\
38.0 12.354 \\
39.0 12.8679 \\
40.0 13.2449 \\
41.0 12.3977 \\
42.0 14.1407 \\
43.0 13.938 \\
44.0 13.9973 \\
45.0 15.2802 \\
  };
\addlegendentry{K2,A2,GB}

\end{axis}

\input{plot/noise_test_event_t_4v}

\end{tikzpicture}%
    \vspace{-0.7cm}
    \caption{
    \textbf{Noise test, event cameras.} Rotation (\textit{left}) and translation (\textit{right}) errors of the solvers with \textit{top:} $m=2,n=5$, \textit{middle:} $m=3,n=2$, and \textit{bottom:} $m=4,n=1$, as the function of the angular velocity $\omega$, Averaged over $1000$ synthetic event samples with additional noise of magnitude $1px$.  The result of the five point solver is shown in all graphs.
    }
    \label{fig:noise_tests_event}
\end{figure}

\subsection{Real experiments} \label{sec:real}

Here, we evaluate the proposed solvers on two real-world rolling shutter datasets, Fastec~\cite{Liu2020CVPR} and Carla~\cite{Liu2020CVPR}, and compare them to the five-point solver~\cite{DBLP:journals/pami/Nister04}. While the rotational velocities in the Fastec sequences are negligible, the Carla dataset typically exhibits a noticeable inter-frame rotation of $1$ to $4$ degrees. Specifically, we compare the solvers $m=2,n=5,A2,K1,HC$, $m=2,n=5,A2,K1,GB3$, $m=2,n=5,A2,K2,HC$, $m=3,n=2,A1,K1,GB$, $m=3,n=2,A2,K2,GB$, and $m=4,n=1,A2,K1,GB$ against the five-point solver. Since the datasets do not provide ground-truth poses or velocities, we reconstructed pseudo-ground-truth poses using COLMAP~\cite{schoenberger2016sfm}. For each solver, we then selected sequences of $m$ consecutive frames, matched them with LightGlue~\cite{lindenberger2023lightglue}, estimated the capture times $t_{i,j}$ following the procedure in Section~\ref{sec:capture_times}, and computed poses using LO-RANSAC~\cite{chum2003locally} with local refinement as described in Section~\ref{sec:lo}. 

The results were evaluated against the pseudo-ground-truth using the same metrics as in Section~\ref{sec:stability}, with the pose error defined as the maximum of the rotation and translation errors. We report the area under the recall curve (AUC) of the pose errors, thresholded at $1^{\circ}, 5^{\circ}, 10^{\circ}, 20^{\circ}$, in Table~\ref{tab:real_test_carla} for the Carla dataset and Table~\ref{tab:real_test_fastec} for the Fastec dataset. The results show that on the Carla dataset, which exhibits small but non-negligible angular velocities, the proposed solvers outperform the five-point solver. In contrast, on the Fastec dataset, where inter-frame rotations are negligible, the five-point solver achieves superior performance.

\setlength{\tabcolsep}{3pt}
\begin{table}[]
    \centering
    \renewcommand{\arraystretch}{0.9}
    \begin{tabular}{c|c c c c c}
        Solver & AUC1 & AUC5 & AUC10 & AUC20  \\
        \hline
        \hline
         5 point \cite{DBLP:journals/pami/Nister04} &		0.10	&	0.41	&	0.60	&	0.72	\\
         \hline
       m=2,A2,K1,HC & 		0.19	&	0.50	&	\textbf{0.64}	&	\textbf{0.74}	\\
       m=2,A2,K1,GB3 & 		0.18	&	0.49	&	0.63	&	\textbf{0.74}	\\
       m=2,A2,K2,HC & 		0.18	&	0.49	&	0.63	&	0.73	\\
       \hline
       m=3,A1,K1,GB & 		0.22	&	0.50	&	0.61	&	0.69	\\
       m=3,A2,K2,GB & 		\textbf{0.23}	&	\textbf{0.52}	&	0.63	&	0.71	\\
       \hline
       m=4,A2,K1,GB &		0.00	&	0.01	&	0.08	&	0.25	\\

    \end{tabular}
    \vspace{-0.3cm}
    \caption{The AUC score at $1^{\circ}$, $5^{\circ}$, $10^{\circ}$, and $20^{\circ}$ on the Carla dataset \cite{Liu2020CVPR}. }
    \label{tab:real_test_carla}
\end{table}

\setlength{\tabcolsep}{3pt}
\begin{table}[]
    \centering
    \renewcommand{\arraystretch}{0.9}
    \begin{tabular}{c|c c c c c}
        Solver & AUC1 & AUC5 & AUC10 & AUC20  \\
        \hline
        \hline
         5 point \cite{DBLP:journals/pami/Nister04} &		\textbf{0.12}	&	\textbf{0.37}	&	\textbf{0.52}	&	\textbf{0.68}	\\
         \hline
       m=2,A2,K1,HC & 		0.09	&	0.30	&	0.46	&	0.63	\\
       m=2,A2,K1,GB3 & 		0.09	&	0.30	&	0.46	&	0.61	\\
       m=2,A2,K2,HC & 		0.09	&	0.30	&	0.47	&	0.63	\\
       \hline
       m=3,A1,K1,GB & 		0.01	&	0.06	&	0.12	&	0.24	\\
       m=3,A2,K2,GB & 		0.02	&	0.08	&	0.14	&	0.23	\\
       \hline
       m=4,A2,K1,GB &		0.01	&	0.17	&	0.32	&	0.47	\\
    \end{tabular}
    \vspace{-0.3cm}
    \caption{The AUC score at $1^{\circ}$, $5^{\circ}$, $10^{\circ}$, and $20^{\circ}$ on the Fastec dataset \cite{Liu2020CVPR}.}
    \label{tab:real_test_fastec}
\end{table}


\section{Conclusion}

In this paper, we addressed the problem of estimating the translational and angular velocity of a camera from asynchronous point tracks, applicable to motion estimation with rolling shutter and event cameras.

Since the original formulation is non-polynomial, we introduced a polynomial approximation, classified the resulting minimal problems, and determined their algebraic degrees. We also provided a geometric explanation for why the five-point solver~\cite{DBLP:journals/pami/Nister04} exactly solves the pure translation case.

We developed minimal solvers for several problems with relatively low degrees (2–122) and evaluated them on synthetic and real data. The experiments show that even with small nonzero rotation, our solvers outperform the five-point solver. The code will be made publicly available.




{
    \small
    \bibliographystyle{ieeenat_fullname}
    \bibliography{main}
}
\clearpage
\setcounter{page}{1}
\maketitlesupplementary

\section{Robustness to Angular Velocity}

We conducted an experiment to assess the robustness of the solvers to varying levels of angular velocity in the absence of additional noise. The proposed solvers rely on either a linear or a quadratic approximation of the rotation $\mathbf{R}(t_{i,j})$ (Sec.\ref{sec:appx}), which inevitably introduces nonzero errors as angular velocity increases. Similarly, while the five-point solver\cite{DBLP:journals/pami/Nister04} provides an exact solution when the angular velocity is zero, its performance deteriorates otherwise.

\begin{figure}[!]
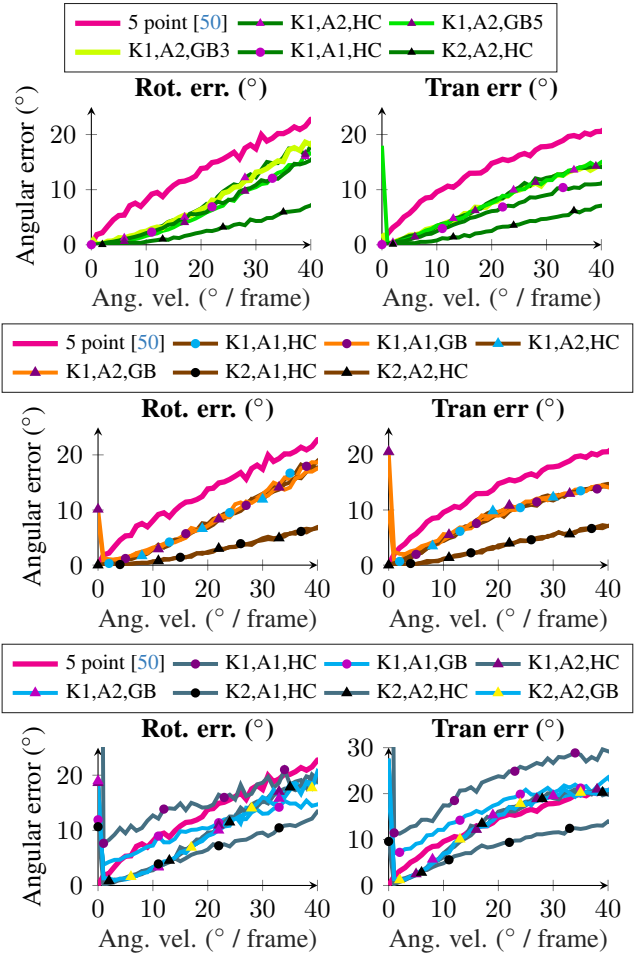

    \centering
    \begin{tikzpicture}

\begin{axis}[%
width=0.35\columnwidth,
height=0.22\columnwidth,
at={(0.879in,0.389in)},
scale only axis,
xmin=0,
xmax=40,
xlabel style={font=\color{white!15!black}, yshift=0.03in},
xlabel={Ang. vel. ($^{\circ}$ / frame)},
title={Rot. err. ($^{\circ}$)},
ymin=0,
ymax=25,
ymode=normal,
yminorticks=true,
axis lines = left,
axis background/.style={fill=white},
title style={font=\bfseries, yshift=-0.1in},
ylabel style={yshift=-0.15in},
ylabel={Angular error ($^{\circ}$)},
legend style={at={(1,1.25)}, anchor=south, legend cell align=left, align=left, draw=white!15!black, font=\footnotesize,nodes={scale=1.0, transform shape}},
legend columns=3
]


\addplot [color=magenta,line width=2pt, mark options={solid, red}]
  table[row sep=crcr]{%
0.0 0.0022219 \\
1.0 1.85416 \\
2.0 2.18186 \\
3.0 3.38412 \\
4.0 4.52827 \\
5.0 5.30342 \\
6.0 5.37784 \\
7.0 7.09994 \\
8.0 7.5512 \\
9.0 6.95374 \\
10.0 8.04667 \\
11.0 9.18663 \\
12.0 9.48936 \\
13.0 8.74055 \\
14.0 9.89586 \\
15.0 10.7786 \\
16.0 11.4276 \\
17.0 11.5423 \\
18.0 11.9007 \\
19.0 12.4542 \\
20.0 13.8287 \\
21.0 13.9512 \\
22.0 14.9124 \\
23.0 15.095 \\
24.0 15.5522 \\
25.0 16.1404 \\
26.0 16.7747 \\
27.0 15.626 \\
28.0 17.5237 \\
29.0 18.137 \\
30.0 17.4101 \\
31.0 19.9508 \\
32.0 18.8117 \\
33.0 19.3262 \\
34.0 20.1772 \\
35.0 20.1603 \\
36.0 20.6474 \\
37.0 21.3843 \\
38.0 20.9099 \\
39.0 21.5072 \\
40.0 22.7475 \\
41.0 22.9492 \\
42.0 25.7099 \\
43.0 24.0498 \\
44.0 24.9044 \\
45.0 23.8587 \\
};
\addlegendentry{5 point \cite{DBLP:journals/pami/Nister04}}


\addplot [color=green!50!black,line width=1.5pt, mark options={solid, violet!50!magenta}, mark=triangle*, mark repeat=11, mark phase=7, mark size=0.5pt]
  table[row sep=crcr]{%
0.0 0.0280612 \\
1.0 0.104975 \\
2.0 0.207327 \\
3.0 0.495947 \\
4.0 0.75764 \\
5.0 1.01602 \\
6.0 1.23102 \\
7.0 1.34757 \\
8.0 1.4802 \\
9.0 2.05025 \\
10.0 2.25241 \\
11.0 2.81004 \\
12.0 3.16344 \\
13.0 3.41425 \\
14.0 4.14701 \\
15.0 4.47361 \\
16.0 4.63083 \\
17.0 5.01558 \\
18.0 5.6154 \\
19.0 6.07063 \\
20.0 6.59551 \\
21.0 7.41492 \\
22.0 7.15212 \\
23.0 8.7022 \\
24.0 9.65405 \\
25.0 10.3296 \\
26.0 10.9646 \\
27.0 10.7271 \\
28.0 12.182 \\
29.0 12.7844 \\
30.0 12.2206 \\
31.0 13.6324 \\
32.0 15.0268 \\
33.0 14.8119 \\
34.0 14.8104 \\
35.0 15.8763 \\
36.0 17.4633 \\
37.0 17.3473 \\
38.0 17.3846 \\
39.0 16.0223 \\
40.0 18.0424 \\
41.0 18.919 \\
42.0 19.5277 \\
43.0 21.2528 \\
44.0 19.9868 \\
45.0 21.5306 \\
};
\addlegendentry{K1,A2,HC}

\addplot [color=green!90!black,line width=1.5pt, mark options={solid, violet}, mark=triangle*, mark repeat=11, mark phase=7, mark size=0.5pt]
  table[row sep=crcr]{%
0.0 0.0 \\
1.0 0.0313134 \\
2.0 0.0903205 \\
3.0 0.180239 \\
4.0 0.283267 \\
5.0 0.479226 \\
6.0 0.623053 \\
7.0 0.840799 \\
8.0 1.05068 \\
9.0 1.28829 \\
10.0 1.62038 \\
11.0 1.89494 \\
12.0 2.07193 \\
13.0 2.52862 \\
14.0 2.7634 \\
15.0 3.17182 \\
16.0 3.71668 \\
17.0 4.04828 \\
18.0 4.34741 \\
19.0 4.75905 \\
20.0 5.28878 \\
21.0 5.87888 \\
22.0 6.28792 \\
23.0 6.73552 \\
24.0 7.08919 \\
25.0 8.19597 \\
26.0 7.96723 \\
27.0 9.19879 \\
28.0 9.72651 \\
29.0 10.1237 \\
30.0 10.4455 \\
31.0 11.0122 \\
32.0 11.1645 \\
33.0 12.3098 \\
34.0 13.0384 \\
35.0 13.3787 \\
36.0 14.4598 \\
37.0 14.4756 \\
38.0 14.7142 \\
39.0 16.4292 \\
40.0 16.5197 \\
41.0 17.2759 \\
42.0 17.262 \\
43.0 19.0089 \\
44.0 19.4109 \\
45.0 19.286 \\
   }; 
\addlegendentry{K1,A2,GB5}

\addplot [color=green!20!yellow,line width=2pt, mark options={solid, red}]
  table[row sep=crcr]{%
0.0 1.28105 \\
1.0 0.102828 \\
2.0 0.174587 \\
3.0 0.640575 \\
4.0 0.696801 \\
5.0 1.28169 \\
6.0 1.21275 \\
7.0 1.57071 \\
8.0 1.76606 \\
9.0 2.27923 \\
10.0 2.45155 \\
11.0 2.93491 \\
12.0 3.20499 \\
13.0 3.02203 \\
14.0 3.96944 \\
15.0 4.3505 \\
16.0 4.9364 \\
17.0 5.37093 \\
18.0 5.24216 \\
19.0 5.87162 \\
20.0 6.41889 \\
21.0 6.8846 \\
22.0 8.10212 \\
23.0 7.98372 \\
24.0 9.36109 \\
25.0 9.64764 \\
26.0 10.4402 \\
27.0 10.5678 \\
28.0 11.6307 \\
29.0 11.6804 \\
30.0 13.1676 \\
31.0 13.3444 \\
32.0 13.8282 \\
33.0 14.575 \\
34.0 15.2121 \\
35.0 15.0936 \\
36.0 17.2031 \\
37.0 17.624 \\
38.0 17.1079 \\
39.0 18.6119 \\
40.0 18.1537 \\
41.0 19.4062 \\
42.0 19.1425 \\
43.0 21.1431 \\
44.0 19.9777 \\
45.0 20.7773 \\
};
\addlegendentry{K1,A2,GB3}


\addplot [color=green!50!black,line width=1.5pt, mark options={solid, violet!50!magenta}, mark=*, mark repeat=11, mark size=1pt]
  table[row sep=crcr]{%
0.0 0.0205019 \\
1.0 0.0607064 \\
2.0 0.235531 \\
3.0 0.441453 \\
4.0 0.571438 \\
5.0 0.762646 \\
6.0 1.05549 \\
7.0 1.21837 \\
8.0 1.45928 \\
9.0 1.74168 \\
10.0 2.1161 \\
11.0 2.29826 \\
12.0 2.70224 \\
13.0 3.02484 \\
14.0 3.60525 \\
15.0 3.78046 \\
16.0 4.47096 \\
17.0 4.61127 \\
18.0 5.61132 \\
19.0 5.51768 \\
20.0 5.67127 \\
21.0 5.92972 \\
22.0 6.88935 \\
23.0 7.81605 \\
24.0 7.53962 \\
25.0 8.10144 \\
26.0 8.24336 \\
27.0 9.66193 \\
28.0 10.0535 \\
29.0 10.2863 \\
30.0 10.8316 \\
31.0 11.7102 \\
32.0 11.4623 \\
33.0 12.0616 \\
34.0 11.9692 \\
35.0 12.7742 \\
36.0 13.5052 \\
37.0 14.6937 \\
38.0 14.5723 \\
39.0 14.8668 \\
40.0 15.3419 \\
41.0 16.4612 \\
42.0 15.5182 \\
43.0 16.6357 \\
44.0 16.7985 \\
45.0 17.2622 \\
   }; 
\addlegendentry{K1,A1,HC}

\addplot [color=green!50!black,line width=1.5pt, mark options={solid,violet, draw=black,fill=violet}, mark=triangle*, mark repeat=11, mark size=0.4pt, mark phase=3]
  table[row sep=crcr]{%
0.0 0.0389094 \\
1.0 0.047351 \\
2.0 0.0371309 \\
3.0 0.0788808 \\
4.0 0.0928927 \\
5.0 0.20049 \\
6.0 0.150226 \\
7.0 0.378643 \\
8.0 0.377737 \\
9.0 0.350526 \\
10.0 0.552642 \\
11.0 0.711282 \\
12.0 0.843427 \\
13.0 1.03239 \\
14.0 0.888571 \\
15.0 1.32275 \\
16.0 1.21555 \\
17.0 1.59607 \\
18.0 1.794 \\
19.0 2.06525 \\
20.0 2.25667 \\
21.0 2.53823 \\
22.0 2.4914 \\
23.0 2.75693 \\
24.0 3.21071 \\
25.0 3.11469 \\
26.0 3.60744 \\
27.0 3.79879 \\
28.0 3.54547 \\
29.0 4.58266 \\
30.0 4.31168 \\
31.0 4.32694 \\
32.0 4.838 \\
33.0 4.88114 \\
34.0 5.42685 \\
35.0 5.94304 \\
36.0 5.99118 \\
37.0 6.11183 \\
38.0 6.24059 \\
39.0 6.79377 \\
40.0 7.13037 \\
41.0 7.85596 \\
42.0 7.89638 \\
43.0 7.68023 \\
44.0 8.02037 \\
45.0 8.82702 \\
   }; 
\addlegendentry{K2,A2,HC}

\end{axis}

\input{plot/appx_test_t}

\end{tikzpicture}%
    \begin{tikzpicture}

\begin{axis}[%
width=0.35\columnwidth,
height=0.22\columnwidth,
at={(0.879in,0.389in)},
scale only axis,
xmin=0,
xmax=40,
xlabel style={font=\color{white!15!black}, yshift=0.03in},
xlabel={Ang. vel. ($^{\circ}$ / frame)},
title={Rot. err. ($^{\circ}$)},
ymin=0,
ymax=25,
ymode=normal,
yminorticks=true,
axis lines = left,
axis background/.style={fill=white},
title style={font=\bfseries, yshift=-0.1in},
ylabel style={yshift=-0.15in},
ylabel={Angular error ($^{\circ}$)},
legend style={at={(1,1.25)}, anchor=south, legend cell align=left, align=left, draw=white!15!black, font=\footnotesize,nodes={scale=1.0, transform shape}},
legend columns=4
]


\addplot [color=magenta,line width=2pt, mark options={solid, red}]
  table[row sep=crcr]{%
0.0 0.0022219 \\
1.0 1.85416 \\
2.0 2.18186 \\
3.0 3.38412 \\
4.0 4.52827 \\
5.0 5.30342 \\
6.0 5.37784 \\
7.0 7.09994 \\
8.0 7.5512 \\
9.0 6.95374 \\
10.0 8.04667 \\
11.0 9.18663 \\
12.0 9.48936 \\
13.0 8.74055 \\
14.0 9.89586 \\
15.0 10.7786 \\
16.0 11.4276 \\
17.0 11.5423 \\
18.0 11.9007 \\
19.0 12.4542 \\
20.0 13.8287 \\
21.0 13.9512 \\
22.0 14.9124 \\
23.0 15.095 \\
24.0 15.5522 \\
25.0 16.1404 \\
26.0 16.7747 \\
27.0 15.626 \\
28.0 17.5237 \\
29.0 18.137 \\
30.0 17.4101 \\
31.0 19.9508 \\
32.0 18.8117 \\
33.0 19.3262 \\
34.0 20.1772 \\
35.0 20.1603 \\
36.0 20.6474 \\
37.0 21.3843 \\
38.0 20.9099 \\
39.0 21.5072 \\
40.0 22.7475 \\
41.0 22.9492 \\
42.0 25.7099 \\
43.0 24.0498 \\
44.0 24.9044 \\
45.0 23.8587 \\
};
\addlegendentry{5 point \cite{DBLP:journals/pami/Nister04}}



\addplot [color=orange!50!black,line width=1.5pt, mark options={solid, orange}, mark options={solid, cyan}, mark=*, mark repeat=11, mark phase = 3, mark size=1pt]
  table[row sep=crcr]{%
0.0 1.19787e-07 \\
1.0 0.16174 \\
2.0 0.310842 \\
3.0 0.459957 \\
4.0 0.789045 \\
5.0 1.15443 \\
6.0 1.27965 \\
7.0 2.1182 \\
8.0 2.20879 \\
9.0 2.51843 \\
10.0 2.81335 \\
11.0 3.71527 \\
12.0 3.22795 \\
13.0 4.18837 \\
14.0 4.63979 \\
15.0 4.7358 \\
16.0 5.8779 \\
17.0 6.15518 \\
18.0 6.38428 \\
19.0 7.60613 \\
20.0 7.45324 \\
21.0 8.41292 \\
22.0 8.6753 \\
23.0 9.32436 \\
24.0 9.52208 \\
25.0 9.74946 \\
26.0 10.14 \\
27.0 11.1785 \\
28.0 11.932 \\
29.0 12.4254 \\
30.0 12.9406 \\
31.0 13.4126 \\
32.0 14.1075 \\
33.0 14.3403 \\
34.0 15.4788 \\
35.0 16.6901 \\
36.0 16.7156 \\
37.0 18.2374 \\
38.0 17.7305 \\
39.0 17.8478 \\
40.0 18.9151 \\
41.0 19.7607 \\
42.0 18.8568 \\
43.0 20.6301 \\
44.0 21.2908 \\
45.0 21.8116 \\
  };
\addlegendentry{K1,A1,HC}

\addplot [color=orange,line width=1.5pt, mark options={solid, orange}, mark options={solid, violet}, mark=*, mark repeat=11, mark phase = 6, mark size=1pt]
  table[row sep=crcr]{%
0.0 8.45523 \\
1.0 0.612677 \\
2.0 0.474265 \\
3.0 1.01809 \\
4.0 1.15377 \\
5.0 1.19837 \\
6.0 1.66832 \\
7.0 2.18092 \\
8.0 2.463 \\
9.0 2.76799 \\
10.0 3.18905 \\
11.0 3.439 \\
12.0 3.8679 \\
13.0 4.31147 \\
14.0 4.484 \\
15.0 4.99427 \\
16.0 5.70202 \\
17.0 6.74142 \\
18.0 6.46402 \\
19.0 6.71863 \\
20.0 7.06935 \\
21.0 8.07131 \\
22.0 8.91214 \\
23.0 8.95911 \\
24.0 9.55931 \\
25.0 11.05 \\
26.0 11.3728 \\
27.0 10.8398 \\
28.0 11.6948 \\
29.0 12.189 \\
30.0 12.9152 \\
31.0 13.7309 \\
32.0 14.9665 \\
33.0 14.5385 \\
34.0 15.8727 \\
35.0 15.5345 \\
36.0 15.6396 \\
37.0 17.5734 \\
38.0 17.93 \\
39.0 18.6695 \\
40.0 18.6089 \\
41.0 19.1659 \\
42.0 20.1754 \\
43.0 20.8013 \\
44.0 22.1486 \\
45.0 21.5213 \\
  };
\addlegendentry{K1,A1,GB}

\addplot [color=orange!50!black,line width=1.5pt, mark options={solid, orange}, mark options={solid, cyan}, mark=triangle*, mark repeat=11, mark phase = 9, mark size=1pt]
  table[row sep=crcr]{%
0.0 0.000250323 \\
1.0 0.1884 \\
2.0 0.257847 \\
3.0 0.524669 \\
4.0 0.705136 \\
5.0 1.06024 \\
6.0 1.05519 \\
7.0 1.43548 \\
8.0 1.67687 \\
9.0 2.07775 \\
10.0 2.82026 \\
11.0 3.24023 \\
12.0 3.18874 \\
13.0 3.5802 \\
14.0 4.53952 \\
15.0 4.61387 \\
16.0 5.14519 \\
17.0 5.83737 \\
18.0 6.35361 \\
19.0 6.6102 \\
20.0 6.74836 \\
21.0 7.31597 \\
22.0 8.1955 \\
23.0 8.35135 \\
24.0 9.27804 \\
25.0 9.41364 \\
26.0 9.82069 \\
27.0 10.6485 \\
28.0 12.0287 \\
29.0 11.8493 \\
30.0 11.9324 \\
31.0 12.7951 \\
32.0 13.351 \\
33.0 13.7056 \\
34.0 14.6647 \\
35.0 14.9894 \\
36.0 15.9193 \\
37.0 15.7603 \\
38.0 16.9081 \\
39.0 17.7483 \\
40.0 17.6878 \\
41.0 18.1133 \\
42.0 18.6258 \\
43.0 18.6292 \\
44.0 20.5152 \\
45.0 19.9058 \\
  };
\addlegendentry{K1,A2,HC}

\addplot [color=orange,line width=1.5pt, mark options={solid, orange}, mark options={solid, violet}, mark=triangle*, mark repeat=11, mark size=1pt]
  table[row sep=crcr]{%
0.0 10.1274 \\
1.0 1.25689 \\
2.0 1.07344 \\
3.0 0.988494 \\
4.0 1.20586 \\
5.0 1.36033 \\
6.0 1.68155 \\
7.0 1.84637 \\
8.0 2.20846 \\
9.0 2.2011 \\
10.0 3.05997 \\
11.0 2.92504 \\
12.0 3.63102 \\
13.0 4.01082 \\
14.0 4.13335 \\
15.0 4.55313 \\
16.0 4.9568 \\
17.0 6.00841 \\
18.0 6.27225 \\
19.0 6.47151 \\
20.0 7.11921 \\
21.0 7.72873 \\
22.0 8.34031 \\
23.0 8.69841 \\
24.0 8.69001 \\
25.0 9.80771 \\
26.0 10.3687 \\
27.0 11.537 \\
28.0 11.2806 \\
29.0 11.9091 \\
30.0 12.3927 \\
31.0 12.6213 \\
32.0 13.5842 \\
33.0 14.1203 \\
34.0 13.6808 \\
35.0 14.0808 \\
36.0 15.7519 \\
37.0 16.067 \\
38.0 16.9644 \\
39.0 16.9601 \\
40.0 17.5351 \\
41.0 17.5005 \\
42.0 18.1037 \\
43.0 19.0885 \\
44.0 19.8114 \\
45.0 19.7595 \\
  };
\addlegendentry{K1,A2,GB}

\addplot [color=orange!50!black,line width=1.5pt, mark options={solid, orange}, mark options={solid, black}, mark=*, mark repeat=11, mark phase = 5, mark size=1pt]
  table[row sep=crcr]{%
0.0 1.48225e-07 \\
1.0 0.000963692 \\
2.0 0.0327956 \\
3.0 0.0470869 \\
4.0 0.10763 \\
5.0 0.127642 \\
6.0 0.21432 \\
7.0 0.400253 \\
8.0 0.416121 \\
9.0 0.411497 \\
10.0 0.582182 \\
11.0 0.716109 \\
12.0 1.12784 \\
13.0 1.06545 \\
14.0 1.30543 \\
15.0 1.4181 \\
16.0 1.39173 \\
17.0 1.77642 \\
18.0 2.17219 \\
19.0 2.04062 \\
20.0 2.39722 \\
21.0 2.66068 \\
22.0 2.81032 \\
23.0 2.94696 \\
24.0 3.2905 \\
25.0 3.35145 \\
26.0 3.891 \\
27.0 3.79349 \\
28.0 4.0318 \\
29.0 4.3888 \\
30.0 4.86936 \\
31.0 5.1495 \\
32.0 4.88257 \\
33.0 5.03419 \\
34.0 5.59629 \\
35.0 5.67775 \\
36.0 6.05843 \\
37.0 6.1022 \\
38.0 5.93517 \\
39.0 6.49407 \\
40.0 6.91738 \\
41.0 7.16535 \\
42.0 7.10344 \\
43.0 6.84596 \\
44.0 7.64007 \\
45.0 7.61635 \\
  };
\addlegendentry{K2,A1,HC}

\addplot [color=orange!50!black,line width=1.5pt, mark options={solid, orange}, mark options={solid, black}, mark=triangle*, mark repeat=11, mark size=1pt]
  table[row sep=crcr]{%
0.0 9.66626e-08 \\
1.0 0.00286411 \\
2.0 0.0167794 \\
3.0 0.0863148 \\
4.0 0.126786 \\
5.0 0.165288 \\
6.0 0.261394 \\
7.0 0.343647 \\
8.0 0.425619 \\
9.0 0.414019 \\
10.0 0.701757 \\
11.0 0.771791 \\
12.0 0.959481 \\
13.0 1.0144 \\
14.0 1.24232 \\
15.0 1.36377 \\
16.0 1.49703 \\
17.0 1.50401 \\
18.0 2.10496 \\
19.0 2.25071 \\
20.0 2.19968 \\
21.0 2.42945 \\
22.0 3.00939 \\
23.0 2.95578 \\
24.0 3.24032 \\
25.0 3.28454 \\
26.0 3.54539 \\
27.0 3.75225 \\
28.0 3.66453 \\
29.0 4.3389 \\
30.0 4.50601 \\
31.0 4.71668 \\
32.0 4.63874 \\
33.0 4.91182 \\
34.0 5.1952 \\
35.0 5.783 \\
36.0 5.68875 \\
37.0 6.40612 \\
38.0 6.60438 \\
39.0 6.46877 \\
40.0 6.6828 \\
41.0 6.79996 \\
42.0 7.07486 \\
43.0 8.22962 \\
44.0 7.56266 \\
45.0 8.62097 \\
  };
\addlegendentry{K2,A2,HC}

\end{axis}

\input{plot/appx_test_3v_t}

\end{tikzpicture}%
    \begin{tikzpicture}

\begin{axis}[%
width=0.35\columnwidth,
height=0.22\columnwidth,
at={(0.879in,0.389in)},
scale only axis,
xmin=0,
xmax=40,
xlabel style={font=\color{white!15!black}, yshift=0.03in},
xlabel={Ang. vel. ($^{\circ}$ / frame)},
title={Rot. err. ($^{\circ}$)},
ymin=0,
ymax=25,
ymode=normal,
yminorticks=true,
axis lines = left,
axis background/.style={fill=white},
title style={font=\bfseries, yshift=-0.1in},
ylabel style={yshift=-0.15in},
ylabel={Angular error ($^{\circ}$)},
legend style={at={(1,1.25)}, anchor=south, legend cell align=left, align=left, draw=white!15!black, font=\footnotesize,nodes={scale=1.0, transform shape}},
legend columns=4
]



\addplot [color=magenta,line width=2pt, mark options={solid, red}]
  table[row sep=crcr]{%
0.0 0.0022219 \\
1.0 1.85416 \\
2.0 2.18186 \\
3.0 3.38412 \\
4.0 4.52827 \\
5.0 5.30342 \\
6.0 5.37784 \\
7.0 7.09994 \\
8.0 7.5512 \\
9.0 6.95374 \\
10.0 8.04667 \\
11.0 9.18663 \\
12.0 9.48936 \\
13.0 8.74055 \\
14.0 9.89586 \\
15.0 10.7786 \\
16.0 11.4276 \\
17.0 11.5423 \\
18.0 11.9007 \\
19.0 12.4542 \\
20.0 13.8287 \\
21.0 13.9512 \\
22.0 14.9124 \\
23.0 15.095 \\
24.0 15.5522 \\
25.0 16.1404 \\
26.0 16.7747 \\
27.0 15.626 \\
28.0 17.5237 \\
29.0 18.137 \\
30.0 17.4101 \\
31.0 19.9508 \\
32.0 18.8117 \\
33.0 19.3262 \\
34.0 20.1772 \\
35.0 20.1603 \\
36.0 20.6474 \\
37.0 21.3843 \\
38.0 20.9099 \\
39.0 21.5072 \\
40.0 22.7475 \\
41.0 22.9492 \\
42.0 25.7099 \\
43.0 24.0498 \\
44.0 24.9044 \\
45.0 23.8587 \\
};
\addlegendentry{5 point \cite{DBLP:journals/pami/Nister04}}

\addplot [color=cyan!40!black,line width=1.5pt, mark options={solid, orange}, mark options={solid, violet}, mark=*, mark repeat=11, mark size=1pt]
  table[row sep=crcr]{%
0.0 180.0 \\
1.0 7.64645 \\
2.0 8.2735 \\
3.0 9.22081 \\
4.0 10.4218 \\
5.0 9.4763 \\
6.0 10.838 \\
7.0 9.92623 \\
8.0 10.4268 \\
9.0 11.7429 \\
10.0 12.7033 \\
11.0 12.1263 \\
12.0 13.8553 \\
13.0 13.9999 \\
14.0 13.899 \\
15.0 14.8833 \\
16.0 13.6678 \\
17.0 14.9905 \\
18.0 14.6248 \\
19.0 14.1232 \\
20.0 14.209 \\
21.0 14.6629 \\
22.0 16.4598 \\
23.0 15.9708 \\
24.0 16.4631 \\
25.0 15.5987 \\
26.0 18.2846 \\
27.0 16.2017 \\
28.0 16.6289 \\
29.0 18.3302 \\
30.0 18.634 \\
31.0 19.7047 \\
32.0 17.7257 \\
33.0 18.6189 \\
34.0 21.0089 \\
35.0 18.3118 \\
36.0 18.842 \\
37.0 19.1214 \\
38.0 17.6937 \\
39.0 20.1428 \\
40.0 19.3104 \\
41.0 20.298 \\
42.0 19.9765 \\
43.0 22.4495 \\
44.0 22.457 \\
45.0 22.2663 \\
  };
\addlegendentry{K1,A1,HC}

\addplot [color=cyan,line width=1.5pt, mark options={solid, orange}, mark options={solid, violet!50!magenta}, mark=*, mark repeat=11, mark size=1pt]
  table[row sep=crcr]{%
0.0 11.9161 \\
1.0 3.76435 \\
2.0 4.27902 \\
3.0 4.64621 \\
4.0 4.6675 \\
5.0 5.26401 \\
6.0 5.48094 \\
7.0 5.93616 \\
8.0 6.49019 \\
9.0 7.02501 \\
10.0 7.23715 \\
11.0 8.95982 \\
12.0 7.82705 \\
13.0 7.82316 \\
14.0 8.43776 \\
15.0 9.84188 \\
16.0 9.71436 \\
17.0 10.0867 \\
18.0 10.1381 \\
19.0 10.4274 \\
20.0 10.316 \\
21.0 10.7574 \\
22.0 11.3863 \\
23.0 11.9729 \\
24.0 12.1831 \\
25.0 12.3813 \\
26.0 11.9714 \\
27.0 12.0384 \\
28.0 13.2584 \\
29.0 13.7688 \\
30.0 13.0042 \\
31.0 13.0692 \\
32.0 13.9057 \\
33.0 14.1997 \\
34.0 15.3527 \\
35.0 14.8509 \\
36.0 14.8768 \\
37.0 14.15 \\
38.0 15.3881 \\
39.0 14.4396 \\
40.0 14.7062 \\
41.0 15.0017 \\
42.0 15.3395 \\
43.0 16.1378 \\
44.0 17.0441 \\
45.0 16.7294 \\
};
\addlegendentry{K1,A1,GB}

\addplot [color=cyan!40!black,line width=1.5pt, mark options={solid, orange}, mark options={solid, violet}, mark=triangle*, mark repeat=11, mark size=1pt]
  table[row sep=crcr]{%
0.0 18.638 \\
1.0 0.849254 \\
2.0 1.03828 \\
3.0 1.01564 \\
4.0 1.02594 \\
5.0 1.21412 \\
6.0 1.48895 \\
7.0 1.74635 \\
8.0 2.21261 \\
9.0 2.76109 \\
10.0 2.88918 \\
11.0 3.44749 \\
12.0 3.89533 \\
13.0 4.53909 \\
14.0 4.9308 \\
15.0 5.66189 \\
16.0 6.32107 \\
17.0 7.31706 \\
18.0 8.33332 \\
19.0 8.35637 \\
20.0 9.59202 \\
21.0 9.55176 \\
22.0 10.9077 \\
23.0 11.4477 \\
24.0 12.5118 \\
25.0 12.4702 \\
26.0 13.4979 \\
27.0 13.5835 \\
28.0 13.9009 \\
29.0 14.3047 \\
30.0 15.3435 \\
31.0 14.8794 \\
32.0 16.3106 \\
33.0 17.05 \\
34.0 17.5609 \\
35.0 17.6713 \\
36.0 17.2841 \\
37.0 17.8613 \\
38.0 18.4741 \\
39.0 18.6916 \\
40.0 19.9267 \\
41.0 20.0965 \\
42.0 20.8104 \\
43.0 20.3213 \\
44.0 20.6317 \\
45.0 21.1985 \\
  };
\addlegendentry{K1,A2,HC}

\addplot [color=cyan,line width=1.5pt, mark options={solid, orange}, mark options={solid, violet!50!magenta}, mark=triangle*, mark repeat=11, mark size=1pt]
  table[row sep=crcr]{%
0.0 18.9708 \\
1.0 1.25563 \\
2.0 1.07499 \\
3.0 0.967775 \\
4.0 0.955136 \\
5.0 1.09811 \\
6.0 1.34728 \\
7.0 1.77913 \\
8.0 2.60222 \\
9.0 2.46119 \\
10.0 3.04279 \\
11.0 3.31753 \\
12.0 3.81256 \\
13.0 4.50667 \\
14.0 5.01166 \\
15.0 6.22918 \\
16.0 5.85502 \\
17.0 7.04601 \\
18.0 7.62938 \\
19.0 8.04283 \\
20.0 8.99583 \\
21.0 9.3498 \\
22.0 10.0172 \\
23.0 10.6977 \\
24.0 10.8372 \\
25.0 12.5797 \\
26.0 11.2483 \\
27.0 14.1116 \\
28.0 14.0496 \\
29.0 15.1009 \\
30.0 15.7282 \\
31.0 15.3846 \\
32.0 14.7271 \\
33.0 15.7852 \\
34.0 16.8379 \\
35.0 18.3928 \\
36.0 18.578 \\
37.0 17.8567 \\
38.0 18.8315 \\
39.0 18.5702 \\
40.0 18.7524 \\
41.0 18.3487 \\
42.0 20.4479 \\
43.0 19.7234 \\
44.0 19.5453 \\
45.0 19.6042 \\
  };
\addlegendentry{K1,A2,GB}

\addplot [color=cyan!40!black,line width=1.5pt, mark options={solid, orange}, mark options={solid, black}, mark=*, mark repeat=11, mark size=1pt]
  table[row sep=crcr]{%
0.0 10.662 \\
1.0 0.0803117 \\
2.0 0.391437 \\
3.0 0.603784 \\
4.0 1.05247 \\
5.0 1.38815 \\
6.0 1.83738 \\
7.0 2.0762 \\
8.0 2.59508 \\
9.0 2.85297 \\
10.0 3.31575 \\
11.0 3.90705 \\
12.0 3.80767 \\
13.0 4.21047 \\
14.0 4.91509 \\
15.0 4.39615 \\
16.0 4.93806 \\
17.0 5.55127 \\
18.0 5.9179 \\
19.0 6.08318 \\
20.0 6.5142 \\
21.0 7.48001 \\
22.0 7.20555 \\
23.0 7.46666 \\
24.0 7.08244 \\
25.0 8.05914 \\
26.0 8.21962 \\
27.0 8.31021 \\
28.0 8.79801 \\
29.0 9.435 \\
30.0 9.58784 \\
31.0 8.97203 \\
32.0 9.47977 \\
33.0 10.4556 \\
34.0 10.8528 \\
35.0 10.9721 \\
36.0 10.77 \\
37.0 11.2797 \\
38.0 11.7239 \\
39.0 12.12 \\
40.0 13.4335 \\
41.0 13.2647 \\
42.0 13.1051 \\
43.0 12.8358 \\
44.0 13.7308 \\
45.0 14.1913 \\
  };
\addlegendentry{K2,A1,HC}

\addplot [color=cyan!40!black,line width=1.5pt, mark options={solid, orange}, mark options={solid, black}, mark=triangle*, mark repeat=11, mark phase=3, mark size=1pt]
  table[row sep=crcr]{%
0.0 18.5924 \\
1.0 1.03934 \\
2.0 0.838023 \\
3.0 1.0599 \\
4.0 0.968991 \\
5.0 1.2987 \\
6.0 1.58059 \\
7.0 1.79989 \\
8.0 2.05127 \\
9.0 2.57974 \\
10.0 2.75429 \\
11.0 3.4089 \\
12.0 3.93228 \\
13.0 4.53574 \\
14.0 4.91114 \\
15.0 6.93405 \\
16.0 6.49138 \\
17.0 7.11155 \\
18.0 7.59669 \\
19.0 8.7697 \\
20.0 9.51949 \\
21.0 9.64883 \\
22.0 10.531 \\
23.0 10.8247 \\
24.0 11.4826 \\
25.0 12.5074 \\
26.0 14.1165 \\
27.0 13.2898 \\
28.0 14.705 \\
29.0 15.2314 \\
30.0 15.1199 \\
31.0 15.9576 \\
32.0 17.0986 \\
33.0 16.8265 \\
34.0 16.2903 \\
35.0 17.8372 \\
36.0 17.3922 \\
37.0 18.1861 \\
38.0 19.7991 \\
39.0 19.4862 \\
40.0 18.8917 \\
41.0 18.238 \\
42.0 20.3886 \\
43.0 20.0105 \\
44.0 20.852 \\
45.0 20.829 \\
  };
\addlegendentry{K2,A2,HC}

\addplot [color=cyan,line width=1.5pt, mark options={solid, orange}, mark options={solid, yellow}, mark=triangle*, mark repeat=11, mark phase=7, mark size=1pt]
  table[row sep=crcr]{%
0.0 18.2707 \\
1.0 1.29341 \\
2.0 1.30182 \\
3.0 0.899834 \\
4.0 1.25027 \\
5.0 1.27192 \\
6.0 1.59028 \\
7.0 1.73451 \\
8.0 2.24857 \\
9.0 2.39441 \\
10.0 3.33825 \\
11.0 3.44268 \\
12.0 4.35456 \\
13.0 4.52659 \\
14.0 5.00281 \\
15.0 6.06711 \\
16.0 6.29506 \\
17.0 6.90811 \\
18.0 7.64757 \\
19.0 8.26176 \\
20.0 9.04446 \\
21.0 9.47981 \\
22.0 11.009 \\
23.0 10.9749 \\
24.0 11.8979 \\
25.0 12.0705 \\
26.0 12.6556 \\
27.0 12.9983 \\
28.0 14.0261 \\
29.0 14.8718 \\
30.0 16.0252 \\
31.0 15.5792 \\
32.0 17.3571 \\
33.0 16.2678 \\
34.0 15.9618 \\
35.0 17.3281 \\
36.0 18.1456 \\
37.0 16.5899 \\
38.0 18.7402 \\
39.0 17.6942 \\
40.0 20.7052 \\
41.0 19.8176 \\
42.0 19.2991 \\
43.0 20.7499 \\
44.0 19.959 \\
45.0 20.5605 \\
  };
\addlegendentry{K2,A2,GB}

\end{axis}

\input{plot/appx_test_4v_t}

\end{tikzpicture}%
    \caption{
    \textbf{Approximation test, rolling shutter cameras. } Rotation (\textit{left}) and translation (\textit{right}) errors of the solvers with \textit{top:} $m=2,n=5$, \textit{middle:} $m=3,n=2$, and \textit{bottom:} $m=4,n=1$, as the function of the angular velocity $\omega$, Averaged over $1000$ synthetic rolling shutter samples without additional noise. The result of the five point solver is shown in all graphs.
    }
    \label{fig:appx_tests}
\end{figure}

To evaluate this behaviour, we generated data following the procedure in Section~\ref{sec:robustness}, but without adding noise to the point projections $\mathbf{p}_{i,j}$, and analyzed the solvers in the same way. The resulting rotation and translation errors for angular velocities ranging from $0$ to $40$ degrees per frame are shown in Figure~\ref{fig:appx_tests} for the rolling shutter camera and in Figure~\ref{fig:appx_tests_event} for the event camera.

The outcomes are consistent with the noisy case (Section~\ref{sec:robustness}): solver errors generally increase with angular velocity, with most proposed solvers being more robust to high angular velocities than the five-point solver. Nevertheless, the five-point solver remains a good approximation for rolling shutter cameras at zero or very low angular velocity, though in the event camera case its errors rise more sharply. Additionally, the experiment revealed degeneracies at $\omega = 0$ for certain solvers—specifically $m=2,n=5,K1,A2,GB5$, $m=3,n=2,K1,A2,GB5$, and most solvers with $m=4,n=1$. Interestingly, in the case of solvers $m=2,n=5,K1,A2,GB5$, $m=3,n=2,K1,A2,GB5$, these degeneracies were not observed in the noisy experiments, likely because the measurement noise prevents the problem from collapsing into the pure translational case.

%
%
%

\begin{figure}[!]
    \centering
    \begin{tikzpicture}

\begin{axis}[%
width=0.35\columnwidth,
height=0.22\columnwidth,
at={(0.879in,0.389in)},
scale only axis,
xmin=0,
xmax=40,
xlabel style={font=\color{white!15!black}, yshift=0.03in},
xlabel={Ang. vel. ($^{\circ}$ / frame)},
title={Rot. err. ($^{\circ}$)},
ymin=0,
ymax=30,
ymode=normal,
yminorticks=true,
axis lines = left,
axis background/.style={fill=white},
title style={font=\bfseries, yshift=-0.1in},
ylabel style={yshift=-0.15in},
ylabel={Angular error ($^{\circ}$)},
legend style={at={(1,1.25)}, anchor=south, legend cell align=left, align=left, draw=white!15!black, font=\footnotesize,nodes={scale=1.0, transform shape}},
legend columns=3
]


\addplot [color=magenta,line width=2pt, mark options={solid, red}]
  table[row sep=crcr]{%
0.0 0.0162566 \\
1.0 4.76197 \\
2.0 6.83872 \\
3.0 8.57609 \\
4.0 10.162 \\
5.0 11.1983 \\
6.0 12.9412 \\
7.0 12.5484 \\
8.0 13.5978 \\
9.0 15.4748 \\
10.0 15.7132 \\
11.0 15.8086 \\
12.0 17.0049 \\
13.0 16.867 \\
14.0 17.4163 \\
15.0 18.0555 \\
16.0 19.0674 \\
17.0 19.6872 \\
18.0 19.4029 \\
19.0 20.2731 \\
20.0 20.3045 \\
21.0 20.5951 \\
22.0 20.3902 \\
23.0 21.5877 \\
24.0 23.6741 \\
25.0 23.0173 \\
26.0 23.3688 \\
27.0 23.6656 \\
28.0 23.6805 \\
29.0 23.0733 \\
30.0 24.3198 \\
31.0 24.7397 \\
32.0 24.5581 \\
33.0 25.8688 \\
34.0 25.5648 \\
35.0 26.581 \\
36.0 27.0537 \\
37.0 26.8892 \\
38.0 27.8486 \\
39.0 26.9442 \\
40.0 27.4182 \\
41.0 27.9059 \\
42.0 28.7945 \\
43.0 27.5029 \\
44.0 29.0051 \\
45.0 28.7457 \\
};
\addlegendentry{5 point \cite{DBLP:journals/pami/Nister04}}


\addplot [color=green!50!black,line width=1.5pt, mark options={solid, violet!50!magenta}, mark=triangle*, mark repeat=11, mark phase=7, mark size=0.5pt]
  table[row sep=crcr]{%
0.0 0.0485197 \\
1.0 0.174933 \\
2.0 0.414387 \\
3.0 0.605753 \\
4.0 0.94171 \\
5.0 1.43143 \\
6.0 1.79438 \\
7.0 2.43903 \\
8.0 2.34911 \\
9.0 3.2611 \\
10.0 3.51406 \\
11.0 4.07127 \\
12.0 4.54945 \\
13.0 4.53103 \\
14.0 6.07147 \\
15.0 6.1146 \\
16.0 6.67708 \\
17.0 7.4272 \\
18.0 7.69436 \\
19.0 8.99518 \\
20.0 8.98013 \\
21.0 9.65017 \\
22.0 9.69531 \\
23.0 10.4737 \\
24.0 10.6301 \\
25.0 11.407 \\
26.0 11.9331 \\
27.0 13.055 \\
28.0 12.3065 \\
29.0 12.6388 \\
30.0 13.5663 \\
31.0 13.8548 \\
32.0 14.5565 \\
33.0 15.0118 \\
34.0 14.889 \\
35.0 15.0675 \\
36.0 15.3268 \\
37.0 15.2543 \\
38.0 16.3022 \\
39.0 16.159 \\
40.0 16.4149 \\
41.0 16.8562 \\
42.0 17.8471 \\
43.0 17.3985 \\
44.0 17.6145 \\
45.0 18.1969 \\
};
\addlegendentry{K1,A2,HC}

\addplot [color=green!90!black,line width=1.5pt, mark options={solid, violet}, mark=triangle*, mark repeat=11, mark phase=7, mark size=0.5pt]
  table[row sep=crcr]{%
0.0 0.0 \\
1.0 0.0373693 \\
2.0 0.115425 \\
3.0 0.249931 \\
4.0 0.473412 \\
5.0 0.592884 \\
6.0 0.857456 \\
7.0 1.09257 \\
8.0 1.45447 \\
9.0 1.79795 \\
10.0 2.0839 \\
11.0 2.74601 \\
12.0 3.13824 \\
13.0 3.40959 \\
14.0 3.90852 \\
15.0 4.58833 \\
16.0 4.96996 \\
17.0 5.31981 \\
18.0 5.56173 \\
19.0 5.69607 \\
20.0 6.70485 \\
21.0 7.17253 \\
22.0 7.49363 \\
23.0 7.92718 \\
24.0 8.75339 \\
25.0 8.8565 \\
26.0 8.80956 \\
27.0 9.35588 \\
28.0 10.456 \\
29.0 10.1699 \\
30.0 11.6656 \\
31.0 11.4611 \\
32.0 11.4619 \\
33.0 11.8111 \\
34.0 12.8514 \\
35.0 13.0601 \\
36.0 13.125 \\
37.0 13.7212 \\
38.0 14.0814 \\
39.0 13.7178 \\
40.0 14.5725 \\
41.0 15.6388 \\
42.0 15.0104 \\
43.0 16.4179 \\
44.0 15.6148 \\
45.0 17.7988 \\
   }; 
\addlegendentry{K1,A2,GB5}

\addplot [color=green!20!yellow,line width=2pt, mark options={solid, red}]
  table[row sep=crcr]{%
0.0 0.337687 \\
1.0 0.12697 \\
2.0 0.42326 \\
3.0 0.657982 \\
4.0 0.715629 \\
5.0 1.02122 \\
6.0 1.50965 \\
7.0 2.02765 \\
8.0 2.55326 \\
9.0 2.71206 \\
10.0 3.30061 \\
11.0 3.93666 \\
12.0 4.34788 \\
13.0 5.0085 \\
14.0 5.27097 \\
15.0 6.71482 \\
16.0 6.45483 \\
17.0 7.32502 \\
18.0 7.32146 \\
19.0 7.79968 \\
20.0 9.73346 \\
21.0 9.83918 \\
22.0 10.1827 \\
23.0 10.1497 \\
24.0 10.9819 \\
25.0 11.1939 \\
26.0 11.1152 \\
27.0 12.3566 \\
28.0 12.5258 \\
29.0 12.1811 \\
30.0 14.1228 \\
31.0 12.6797 \\
32.0 14.2209 \\
33.0 13.4583 \\
34.0 15.1798 \\
35.0 14.2968 \\
36.0 15.527 \\
37.0 15.3003 \\
38.0 14.4284 \\
39.0 16.0288 \\
40.0 16.4681 \\
41.0 16.7005 \\
42.0 16.7092 \\
43.0 16.8076 \\
44.0 18.1809 \\
45.0 18.1612 \\
};
\addlegendentry{K1,A2,GB3}


\addplot [color=green!50!black,line width=1.5pt, mark options={solid, violet!50!magenta}, mark=*, mark repeat=11, mark size=1pt]
  table[row sep=crcr]{%
0.0 6.74733e-05 \\
1.0 0.233327 \\
2.0 0.340914 \\
3.0 0.450682 \\
4.0 0.867503 \\
5.0 1.10026 \\
6.0 1.32159 \\
7.0 1.53645 \\
8.0 2.29416 \\
9.0 2.40595 \\
10.0 3.08748 \\
11.0 3.14899 \\
12.0 3.9226 \\
13.0 4.25928 \\
14.0 4.7167 \\
15.0 5.06805 \\
16.0 5.636 \\
17.0 6.30865 \\
18.0 5.84678 \\
19.0 7.11989 \\
20.0 7.10188 \\
21.0 7.73608 \\
22.0 8.17164 \\
23.0 9.01882 \\
24.0 8.66021 \\
25.0 9.3472 \\
26.0 9.62172 \\
27.0 9.90607 \\
28.0 10.0759 \\
29.0 11.3563 \\
30.0 11.697 \\
31.0 11.9965 \\
32.0 12.058 \\
33.0 12.1155 \\
34.0 12.7833 \\
35.0 12.8948 \\
36.0 13.335 \\
37.0 13.943 \\
38.0 13.8225 \\
39.0 14.5606 \\
40.0 14.3086 \\
41.0 14.2376 \\
42.0 14.774 \\
43.0 15.0125 \\
44.0 15.9475 \\
45.0 17.1619 \\
   }; 
\addlegendentry{K1,A1,HC}

\addplot [color=green!50!black,line width=1.5pt, mark options={solid,violet, draw=black,fill=violet}, mark=triangle*, mark repeat=11, mark size=0.4pt, mark phase=3]
  table[row sep=crcr]{%
0.0 0.038726 \\
1.0 0.0101698 \\
2.0 0.0672355 \\
3.0 0.217749 \\
4.0 0.193722 \\
5.0 0.343667 \\
6.0 0.386782 \\
7.0 0.781684 \\
8.0 0.795769 \\
9.0 0.973824 \\
10.0 1.18198 \\
11.0 1.43635 \\
12.0 1.55649 \\
13.0 1.55039 \\
14.0 1.86869 \\
15.0 2.15675 \\
16.0 2.10618 \\
17.0 2.43834 \\
18.0 2.68475 \\
19.0 2.77889 \\
20.0 3.33492 \\
21.0 3.42825 \\
22.0 3.48947 \\
23.0 3.64833 \\
24.0 3.75431 \\
25.0 4.06312 \\
26.0 4.37763 \\
27.0 4.66262 \\
28.0 4.98804 \\
29.0 4.66172 \\
30.0 5.1046 \\
31.0 5.01443 \\
32.0 5.17218 \\
33.0 5.95955 \\
34.0 6.27084 \\
35.0 6.09999 \\
36.0 5.98005 \\
37.0 6.0675 \\
38.0 7.01875 \\
39.0 6.40173 \\
40.0 6.62677 \\
41.0 7.03476 \\
42.0 7.91452 \\
43.0 7.48194 \\
44.0 8.07069 \\
45.0 8.00197 \\
   }; 
\addlegendentry{K2,A2,HC}

\end{axis}

\input{plot/appx_test_event_t}

\end{tikzpicture}%
    \begin{tikzpicture}

\begin{axis}[%
width=0.35\columnwidth,
height=0.22\columnwidth,
at={(0.879in,0.389in)},
scale only axis,
xmin=0,
xmax=40,
xlabel style={font=\color{white!15!black}, yshift=0.03in},
xlabel={Ang. vel. ($^{\circ}$ / frame)},
title={Rot. err. ($^{\circ}$)},
ymin=0,
ymax=30,
ymode=normal,
yminorticks=true,
axis lines = left,
axis background/.style={fill=white},
title style={font=\bfseries, yshift=-0.1in},
ylabel style={yshift=-0.15in},
ylabel={Angular error ($^{\circ}$)},
legend style={at={(1,1.25)}, anchor=south, legend cell align=left, align=left, draw=white!15!black, font=\footnotesize,nodes={scale=1.0, transform shape}},
legend columns=4
]


\addplot [color=magenta,line width=2pt, mark options={solid, red}]
  table[row sep=crcr]{%
0.0 0.0162566 \\
1.0 4.76197 \\
2.0 6.83872 \\
3.0 8.57609 \\
4.0 10.162 \\
5.0 11.1983 \\
6.0 12.9412 \\
7.0 12.5484 \\
8.0 13.5978 \\
9.0 15.4748 \\
10.0 15.7132 \\
11.0 15.8086 \\
12.0 17.0049 \\
13.0 16.867 \\
14.0 17.4163 \\
15.0 18.0555 \\
16.0 19.0674 \\
17.0 19.6872 \\
18.0 19.4029 \\
19.0 20.2731 \\
20.0 20.3045 \\
21.0 20.5951 \\
22.0 20.3902 \\
23.0 21.5877 \\
24.0 23.6741 \\
25.0 23.0173 \\
26.0 23.3688 \\
27.0 23.6656 \\
28.0 23.6805 \\
29.0 23.0733 \\
30.0 24.3198 \\
31.0 24.7397 \\
32.0 24.5581 \\
33.0 25.8688 \\
34.0 25.5648 \\
35.0 26.581 \\
36.0 27.0537 \\
37.0 26.8892 \\
38.0 27.8486 \\
39.0 26.9442 \\
40.0 27.4182 \\
41.0 27.9059 \\
42.0 28.7945 \\
43.0 27.5029 \\
44.0 29.0051 \\
45.0 28.7457 \\
};
\addlegendentry{5 point \cite{DBLP:journals/pami/Nister04}}



\addplot [color=orange!50!black,line width=1.5pt, mark options={solid, orange}, mark options={solid, cyan}, mark=*, mark repeat=11, mark phase = 3, mark size=1pt]
  table[row sep=crcr]{%
0.0 0.00825856 \\
1.0 0.237914 \\
2.0 0.437964 \\
3.0 0.83876 \\
4.0 0.903215 \\
5.0 1.91269 \\
6.0 2.62505 \\
7.0 3.06258 \\
8.0 3.26355 \\
9.0 4.21991 \\
10.0 5.06463 \\
11.0 6.47727 \\
12.0 6.82595 \\
13.0 7.98909 \\
14.0 8.63781 \\
15.0 8.88056 \\
16.0 9.68639 \\
17.0 11.0628 \\
18.0 11.4889 \\
19.0 11.5277 \\
20.0 11.9821 \\
21.0 13.0324 \\
22.0 13.2806 \\
23.0 15.4921 \\
24.0 14.5784 \\
25.0 16.5873 \\
26.0 16.3624 \\
27.0 17.1994 \\
28.0 18.2476 \\
29.0 17.8069 \\
30.0 19.26 \\
31.0 20.8124 \\
32.0 18.0981 \\
33.0 20.0481 \\
34.0 20.6212 \\
35.0 18.422 \\
36.0 20.183 \\
37.0 20.3374 \\
38.0 22.2692 \\
39.0 22.8976 \\
40.0 21.1895 \\
41.0 21.82 \\
42.0 21.9518 \\
43.0 21.9939 \\
44.0 22.595 \\
45.0 20.8147 \\
  };
\addlegendentry{K1,A1,HC}

\addplot [color=orange,line width=1.5pt, mark options={solid, orange}, mark options={solid, violet}, mark=*, mark repeat=11, mark phase = 6, mark size=1pt]
  table[row sep=crcr]{%
0.0 7.34603 \\
1.0 0.810903 \\
2.0 1.05223 \\
3.0 1.17024 \\
4.0 1.55816 \\
5.0 2.32404 \\
6.0 2.54593 \\
7.0 2.80063 \\
8.0 4.04317 \\
9.0 4.753 \\
10.0 5.53604 \\
11.0 5.84198 \\
12.0 5.96491 \\
13.0 7.44295 \\
14.0 8.33914 \\
15.0 9.23512 \\
16.0 9.64676 \\
17.0 11.2534 \\
18.0 11.1023 \\
19.0 12.1844 \\
20.0 12.974 \\
21.0 13.1096 \\
22.0 14.264 \\
23.0 13.842 \\
24.0 15.9527 \\
25.0 16.7336 \\
26.0 16.3827 \\
27.0 17.245 \\
28.0 18.6865 \\
29.0 17.06 \\
30.0 17.9539 \\
31.0 18.1362 \\
32.0 18.9755 \\
33.0 19.0081 \\
34.0 19.8092 \\
35.0 20.4879 \\
36.0 20.4018 \\
37.0 20.3258 \\
38.0 21.7792 \\
39.0 21.8024 \\
40.0 21.0436 \\
41.0 23.1551 \\
42.0 22.65 \\
43.0 22.9618 \\
44.0 24.9473 \\
45.0 22.8823 \\
  };
\addlegendentry{K1,A1,GB}

\addplot [color=orange!50!black,line width=1.5pt, mark options={solid, orange}, mark options={solid, cyan}, mark=triangle*, mark repeat=11, mark phase = 9, mark size=1pt]
  table[row sep=crcr]{%
0.0 0.00800983 \\
1.0 0.248315 \\
2.0 0.334925 \\
3.0 0.661606 \\
4.0 0.976268 \\
5.0 1.31715 \\
6.0 2.06532 \\
7.0 2.23555 \\
8.0 3.06084 \\
9.0 3.25099 \\
10.0 3.95349 \\
11.0 4.67557 \\
12.0 6.3688 \\
13.0 6.37762 \\
14.0 6.75904 \\
15.0 8.01029 \\
16.0 8.05353 \\
17.0 9.44327 \\
18.0 10.7814 \\
19.0 10.4487 \\
20.0 11.3416 \\
21.0 12.3507 \\
22.0 12.2316 \\
23.0 14.2884 \\
24.0 13.8407 \\
25.0 14.6514 \\
26.0 15.1901 \\
27.0 15.6638 \\
28.0 15.7892 \\
29.0 17.2447 \\
30.0 16.9775 \\
31.0 18.5075 \\
32.0 17.5448 \\
33.0 18.0735 \\
34.0 19.1211 \\
35.0 19.2785 \\
36.0 19.6191 \\
37.0 21.0851 \\
38.0 20.4211 \\
39.0 21.8843 \\
40.0 21.5839 \\
41.0 21.7138 \\
42.0 21.357 \\
43.0 21.4714 \\
44.0 23.4518 \\
45.0 23.2709 \\
  };
\addlegendentry{K1,A2,HC}

\addplot [color=orange,line width=1.5pt, mark options={solid, orange}, mark options={solid, violet}, mark=triangle*, mark repeat=11, mark size=1pt]
  table[row sep=crcr]{%
0.0 8.99563 \\
1.0 0.65017 \\
2.0 0.741201 \\
3.0 1.091 \\
4.0 1.24992 \\
5.0 1.73706 \\
6.0 2.29254 \\
7.0 2.46358 \\
8.0 2.82703 \\
9.0 3.62602 \\
10.0 4.56127 \\
11.0 5.29575 \\
12.0 5.76663 \\
13.0 6.62776 \\
14.0 6.94897 \\
15.0 7.99781 \\
16.0 8.9947 \\
17.0 9.36752 \\
18.0 10.238 \\
19.0 9.98056 \\
20.0 11.9284 \\
21.0 11.8336 \\
22.0 13.722 \\
23.0 12.9511 \\
24.0 13.829 \\
25.0 14.9427 \\
26.0 15.2369 \\
27.0 15.1604 \\
28.0 15.3157 \\
29.0 15.931 \\
30.0 17.6035 \\
31.0 18.2722 \\
32.0 18.752 \\
33.0 18.6882 \\
34.0 19.4451 \\
35.0 19.2287 \\
36.0 19.9652 \\
37.0 20.8179 \\
38.0 21.9675 \\
39.0 21.0216 \\
40.0 21.7439 \\
41.0 21.3606 \\
42.0 24.5734 \\
43.0 23.5824 \\
44.0 23.3988 \\
45.0 23.4238 \\
  };
\addlegendentry{K1,A2,GB}

\addplot [color=orange!50!black,line width=1.5pt, mark options={solid, orange}, mark options={solid, black}, mark=*, mark repeat=11, mark phase = 5, mark size=1pt]
  table[row sep=crcr]{%
0.0 1.48225e-07 \\
  };
\addlegendentry{K2,A1,HC}

\addplot [color=orange!50!black,line width=1.5pt, mark options={solid, orange}, mark options={solid, black}, mark=triangle*, mark repeat=11, mark size=1pt]
  table[row sep=crcr]{%
0.0 1.65175e-07 \\
1.0 0.00462639 \\
2.0 0.180433 \\
3.0 0.112852 \\
4.0 0.318873 \\
5.0 0.542281 \\
6.0 0.628709 \\
7.0 0.91595 \\
8.0 0.878719 \\
9.0 1.30052 \\
10.0 1.83424 \\
11.0 1.67817 \\
12.0 2.17724 \\
13.0 2.38854 \\
14.0 2.62259 \\
15.0 3.12429 \\
16.0 3.26832 \\
17.0 3.38677 \\
18.0 3.7604 \\
19.0 4.49318 \\
20.0 4.32529 \\
21.0 4.37516 \\
22.0 4.57709 \\
23.0 4.69875 \\
24.0 5.84947 \\
25.0 5.54099 \\
26.0 6.04662 \\
27.0 6.40539 \\
28.0 6.48971 \\
29.0 7.03729 \\
30.0 7.03467 \\
31.0 6.90715 \\
32.0 7.47423 \\
33.0 7.87922 \\
34.0 8.44709 \\
35.0 8.61188 \\
36.0 8.24976 \\
37.0 8.9831 \\
38.0 9.15988 \\
39.0 9.43113 \\
40.0 9.73051 \\
41.0 10.3902 \\
42.0 9.85969 \\
43.0 10.9865 \\
44.0 11.2104 \\
45.0 11.8847 \\
  };
\addlegendentry{K2,A2,HC}

\end{axis}

\input{plot/appx_test_event_3v_t}

\end{tikzpicture}%
    \begin{tikzpicture}

\begin{axis}[%
width=0.35\columnwidth,
height=0.22\columnwidth,
at={(0.879in,0.389in)},
scale only axis,
xmin=0,
xmax=40,
xlabel style={font=\color{white!15!black}, yshift=0.03in},
xlabel={Ang. vel. ($^{\circ}$ / frame)},
title={Rot. err. ($^{\circ}$)},
ymin=0,
ymax=30,
ymode=normal,
yminorticks=true,
axis lines = left,
axis background/.style={fill=white},
title style={font=\bfseries, yshift=-0.1in},
ylabel style={yshift=-0.15in},
ylabel={Angular error ($^{\circ}$)},
legend style={at={(1,1.25)}, anchor=south, legend cell align=left, align=left, draw=white!15!black, font=\footnotesize,nodes={scale=1.0, transform shape}},
legend columns=4
]



\addplot [color=magenta,line width=2pt, mark options={solid, red}]
  table[row sep=crcr]{%
0.0 0.0162566 \\
1.0 4.76197 \\
2.0 6.83872 \\
3.0 8.57609 \\
4.0 10.162 \\
5.0 11.1983 \\
6.0 12.9412 \\
7.0 12.5484 \\
8.0 13.5978 \\
9.0 15.4748 \\
10.0 15.7132 \\
11.0 15.8086 \\
12.0 17.0049 \\
13.0 16.867 \\
14.0 17.4163 \\
15.0 18.0555 \\
16.0 19.0674 \\
17.0 19.6872 \\
18.0 19.4029 \\
19.0 20.2731 \\
20.0 20.3045 \\
21.0 20.5951 \\
22.0 20.3902 \\
23.0 21.5877 \\
24.0 23.6741 \\
25.0 23.0173 \\
26.0 23.3688 \\
27.0 23.6656 \\
28.0 23.6805 \\
29.0 23.0733 \\
30.0 24.3198 \\
31.0 24.7397 \\
32.0 24.5581 \\
33.0 25.8688 \\
34.0 25.5648 \\
35.0 26.581 \\
36.0 27.0537 \\
37.0 26.8892 \\
38.0 27.8486 \\
39.0 26.9442 \\
40.0 27.4182 \\
41.0 27.9059 \\
42.0 28.7945 \\
43.0 27.5029 \\
44.0 29.0051 \\
45.0 28.7457 \\
};
\addlegendentry{5 point \cite{DBLP:journals/pami/Nister04}}

\addplot [color=cyan!40!black,line width=1.5pt, mark options={solid, orange}, mark options={solid, violet}, mark=*, mark repeat=11, mark size=1pt]
  table[row sep=crcr]{%
0.0 180.0 \\
1.0 8.8225 \\
2.0 10.798 \\
3.0 9.99697 \\
4.0 10.1633 \\
5.0 11.2848 \\
6.0 12.1308 \\
7.0 11.2973 \\
8.0 10.5692 \\
9.0 11.7356 \\
10.0 11.8542 \\
11.0 10.8948 \\
12.0 12.4905 \\
13.0 14.4074 \\
14.0 14.0975 \\
15.0 15.2838 \\
16.0 16.2249 \\
17.0 15.5114 \\
18.0 15.4219 \\
19.0 16.345 \\
20.0 16.8876 \\
21.0 16.6848 \\
22.0 15.2596 \\
23.0 16.2869 \\
24.0 16.6299 \\
25.0 16.2816 \\
26.0 19.1744 \\
27.0 20.2732 \\
28.0 20.2627 \\
29.0 17.4905 \\
30.0 21.1225 \\
31.0 21.4033 \\
32.0 19.9192 \\
33.0 20.7797 \\
34.0 20.2304 \\
35.0 22.6371 \\
36.0 22.0198 \\
37.0 25.6139 \\
38.0 23.9922 \\
39.0 22.8589 \\
40.0 24.5323 \\
41.0 26.1232 \\
42.0 27.1996 \\
43.0 24.8766 \\
44.0 27.132 \\
45.0 26.3892 \\
  };
\addlegendentry{K1,A1,HC}

\addplot [color=cyan,line width=1.5pt, mark options={solid, orange}, mark options={solid, violet!50!magenta}, mark=*, mark repeat=11, mark size=1pt]
  table[row sep=crcr]{%
0.0 10.4907 \\
1.0 4.35184 \\
2.0 3.96072 \\
3.0 4.30584 \\
4.0 4.82671 \\
5.0 5.34236 \\
6.0 5.93089 \\
7.0 5.71624 \\
8.0 6.30198 \\
9.0 7.5451 \\
10.0 6.72252 \\
11.0 7.74774 \\
12.0 8.15579 \\
13.0 9.12054 \\
14.0 8.15508 \\
15.0 9.20744 \\
16.0 9.25884 \\
17.0 9.02244 \\
18.0 10.6084 \\
19.0 11.3671 \\
20.0 10.6846 \\
21.0 11.0758 \\
22.0 11.4965 \\
23.0 12.0945 \\
24.0 12.7223 \\
25.0 14.4506 \\
26.0 12.8885 \\
27.0 14.6939 \\
28.0 14.2382 \\
29.0 14.9127 \\
30.0 14.7257 \\
31.0 15.4682 \\
32.0 16.183 \\
33.0 14.8936 \\
34.0 16.2736 \\
35.0 15.6762 \\
36.0 16.5723 \\
37.0 17.3484 \\
38.0 15.997 \\
39.0 18.0691 \\
40.0 18.6475 \\
41.0 18.8011 \\
42.0 18.7117 \\
43.0 18.6176 \\
44.0 18.7952 \\
45.0 19.7399 \\
};
\addlegendentry{K1,A1,GB}

\addplot [color=cyan!40!black,line width=1.5pt, mark options={solid, orange}, mark options={solid, violet}, mark=triangle*, mark repeat=11, mark size=1pt]
  table[row sep=crcr]{%
0.0 40.0 \\
1.0 0.413701 \\
2.0 0.68834 \\
3.0 1.00196 \\
4.0 1.10746 \\
5.0 1.92183 \\
6.0 2.35346 \\
7.0 3.61254 \\
8.0 4.73242 \\
9.0 5.60979 \\
10.0 6.11249 \\
11.0 7.2228 \\
12.0 8.18697 \\
13.0 9.06665 \\
14.0 8.78314 \\
15.0 9.49426 \\
16.0 10.0667 \\
17.0 11.2569 \\
18.0 11.7198 \\
19.0 11.8902 \\
20.0 12.1229 \\
21.0 13.0285 \\
22.0 13.0401 \\
23.0 15.4097 \\
24.0 13.5418 \\
25.0 14.4148 \\
26.0 14.8303 \\
27.0 14.9876 \\
28.0 15.7013 \\
29.0 15.2459 \\
30.0 15.8679 \\
31.0 14.3973 \\
32.0 17.2629 \\
33.0 14.2515 \\
34.0 16.5076 \\
35.0 16.5102 \\
36.0 16.6088 \\
37.0 16.6813 \\
38.0 16.5147 \\
39.0 16.6543 \\
40.0 17.6609 \\
41.0 16.223 \\
42.0 18.1459 \\
43.0 17.4422 \\
44.0 17.3736 \\
45.0 17.3593 \\
  };
\addlegendentry{K1,A2,HC}

\addplot [color=cyan,line width=1.5pt, mark options={solid, orange}, mark options={solid, violet!50!magenta}, mark=triangle*, mark repeat=11, mark size=1pt]
  table[row sep=crcr]{%
0.0 17.9257 \\
1.0 0.744438 \\
2.0 0.912123 \\
3.0 1.14848 \\
4.0 1.37648 \\
5.0 2.1218 \\
6.0 2.37247 \\
7.0 3.45133 \\
8.0 4.55301 \\
9.0 5.16264 \\
10.0 5.36572 \\
11.0 6.78059 \\
12.0 7.34497 \\
13.0 8.09422 \\
14.0 8.31717 \\
15.0 9.94553 \\
16.0 10.5622 \\
17.0 10.8511 \\
18.0 11.4875 \\
19.0 11.9088 \\
20.0 12.0328 \\
21.0 12.6483 \\
22.0 12.2988 \\
23.0 13.8649 \\
24.0 12.9975 \\
25.0 13.1502 \\
26.0 13.4184 \\
27.0 14.3257 \\
28.0 13.6184 \\
29.0 14.8719 \\
30.0 15.3979 \\
31.0 16.2493 \\
32.0 15.5027 \\
33.0 14.3084 \\
34.0 14.7917 \\
35.0 13.7768 \\
36.0 15.6084 \\
37.0 16.9828 \\
38.0 15.3716 \\
39.0 16.6533 \\
40.0 15.4617 \\
41.0 17.0911 \\
42.0 16.5787 \\
43.0 17.0824 \\
44.0 16.6956 \\
45.0 16.6411 \\
  };
\addlegendentry{K1,A2,GB}

\addplot [color=cyan!40!black,line width=1.5pt, mark options={solid, orange}, mark options={solid, black}, mark=*, mark repeat=11, mark size=1pt]
  table[row sep=crcr]{%
0.0 18.3722 \\
1.0 0.323483 \\
2.0 1.1077 \\
3.0 2.3371 \\
4.0 2.28915 \\
5.0 3.19471 \\
6.0 3.7016 \\
7.0 5.02691 \\
8.0 5.96666 \\
9.0 6.4119 \\
10.0 7.00129 \\
11.0 7.73588 \\
12.0 7.78785 \\
13.0 8.7054 \\
14.0 9.20885 \\
15.0 9.028 \\
16.0 8.6926 \\
17.0 10.0549 \\
18.0 11.3055 \\
19.0 11.4904 \\
20.0 11.3787 \\
21.0 11.6684 \\
22.0 11.7558 \\
23.0 12.4844 \\
24.0 11.3341 \\
25.0 13.8153 \\
26.0 13.3395 \\
27.0 14.9042 \\
28.0 13.7836 \\
29.0 13.7859 \\
30.0 13.6956 \\
31.0 14.5255 \\
32.0 14.5566 \\
33.0 15.7917 \\
34.0 15.0601 \\
35.0 16.7587 \\
36.0 14.9923 \\
37.0 15.7607 \\
38.0 16.5271 \\
39.0 17.2344 \\
40.0 16.6597 \\
41.0 16.8325 \\
42.0 17.2518 \\
43.0 18.872 \\
44.0 17.8072 \\
45.0 19.2224 \\
  };
\addlegendentry{K2,A1,HC}

\addplot [color=cyan!40!black,line width=1.5pt, mark options={solid, orange}, mark options={solid, black}, mark=triangle*, mark repeat=11, mark phase=3, mark size=1pt]
  table[row sep=crcr]{%
0.0 18.9384 \\
1.0 0.348406 \\
2.0 1.0662 \\
3.0 1.76819 \\
4.0 2.51937 \\
5.0 3.76548 \\
6.0 4.05451 \\
7.0 5.05033 \\
8.0 5.05573 \\
9.0 5.87797 \\
10.0 5.69387 \\
11.0 6.59276 \\
12.0 6.57289 \\
13.0 6.35162 \\
14.0 7.1966 \\
15.0 7.11536 \\
16.0 8.13185 \\
17.0 8.78848 \\
18.0 8.04381 \\
19.0 8.44845 \\
20.0 8.70234 \\
21.0 9.27353 \\
22.0 9.25246 \\
23.0 9.2707 \\
24.0 9.56325 \\
25.0 8.94733 \\
26.0 9.15558 \\
27.0 9.95459 \\
28.0 9.90025 \\
29.0 10.1505 \\
30.0 10.5201 \\
31.0 10.6084 \\
32.0 11.228 \\
33.0 11.3297 \\
34.0 12.1804 \\
35.0 11.317 \\
36.0 12.3639 \\
37.0 12.7221 \\
38.0 13.0646 \\
39.0 13.4872 \\
40.0 13.7195 \\
41.0 13.2614 \\
42.0 14.7321 \\
43.0 14.1931 \\
44.0 14.3414 \\
45.0 14.4032 \\
  };
\addlegendentry{K2,A2,HC}

\addplot [color=cyan,line width=1.5pt, mark options={solid, orange}, mark options={solid, yellow}, mark=triangle*, mark repeat=11, mark phase=7, mark size=1pt]
  table[row sep=crcr]{%
0.0 0.0403004 \\
1.0 0.202721 \\
2.0 1.05413 \\
3.0 1.59704 \\
4.0 1.95686 \\
5.0 3.24288 \\
6.0 3.42004 \\
7.0 3.98169 \\
8.0 4.45519 \\
9.0 5.54716 \\
10.0 5.19104 \\
11.0 5.78315 \\
12.0 5.79647 \\
13.0 6.73615 \\
14.0 6.5047 \\
15.0 7.61177 \\
16.0 7.43221 \\
17.0 7.76877 \\
18.0 7.03596 \\
19.0 8.19169 \\
20.0 8.33753 \\
21.0 8.58311 \\
22.0 8.38918 \\
23.0 8.65827 \\
24.0 8.74554 \\
25.0 9.47184 \\
26.0 9.8211 \\
27.0 9.76355 \\
28.0 9.57512 \\
29.0 9.79714 \\
30.0 10.1344 \\
31.0 10.9549 \\
32.0 10.9015 \\
33.0 11.499 \\
34.0 11.4309 \\
35.0 11.3846 \\
36.0 11.9957 \\
37.0 12.8853 \\
38.0 12.955 \\
39.0 11.9569 \\
40.0 13.2418 \\
41.0 13.0156 \\
42.0 13.8584 \\
43.0 13.4288 \\
44.0 12.9764 \\
45.0 15.2765 \\
  };
\addlegendentry{K2,A2,GB}

\end{axis}

\input{plot/appx_test_event_4v_t}

\end{tikzpicture}%
    \caption{
    \textbf{Approximation test, event cameras.} Rotation (\textit{left}) and translation (\textit{right}) errors of the solvers with \textit{top:} $m=2,n=5$, \textit{middle:} $m=3,n=2$, and \textit{bottom:} $m=4,n=1$, as the function of the angular velocity $\omega$, Averaged over $1000$ synthetic event samples without additional noise. The result of the five point solver is shown in all graphs.
    }
    \label{fig:appx_tests_event}
\end{figure}



\end{document}